\documentclass{fairmeta}

\usepackage{amsmath,amsfonts,bm}

\usepackage{amsthm}
\theoremstyle{definition}
\newtheorem{definition}{Definition}

\usepackage{hyperref}
\usepackage{url}
\usepackage{graphicx}

  \usepackage{xurl}
  \let\metaabstract\abstract
  \RenewDocumentEnvironment{abstract}{+b}{\metaabstract{#1}\maketitle}{}

\usepackage{booktabs}
\usepackage{multirow}
\usepackage{wrapfig}
\usepackage{needspace}
\usepackage{placeins}
\usepackage{tikz}
\usepackage{colortbl}
\newlength{\fdrcolwidth}
\newlength{\fdrinset}
\newcommand{\fdrsetwidth}{%
  \setlength{\tabcolsep}{0.3pt}
  \settowidth{\fdrcolwidth}{siglip2}%
  \settowidth{\fdrinset}{00.00}%
  \setlength{\fdrinset}{-\fdrinset}\addtolength{\fdrinset}{\fdrcolwidth}%
  \setlength{\fdrinset}{0.5\fdrinset}}
\newcolumntype{F}{>{\raggedleft\arraybackslash}p{\fdrcolwidth}<{\hspace*{\fdrinset}}}
\newcolumntype{G}{>{\columncolor{black!10}\raggedleft\arraybackslash}p{\fdrcolwidth}<{\hspace*{\fdrinset}}}
\newcommand{\fdrhead}[1]{\multicolumn{1}{c}{#1}}
\newcommand{\fdrheadg}[1]{\multicolumn{1}{>{\columncolor{black!10}}c}{#1}}
\newcommand{\fdrtablefont}{\fontsize{6.3}{7.6}\selectfont}
\usetikzlibrary{arrows.meta,positioning,fit,backgrounds,calc,patterns}

\AddToHook{para/begin}{\looseness=-1\relax}

\newcommand{\method}{FeaTok}

\title{\method{}: Feature-Loss Tokenization for \\ Image Generation}

  \author[1,2,3]{Zhang Xiaofeng}
  \author[1,2,\dagger]{Samuel Lavoie}
  \author[3,6,\dagger]{Nicolas Beltran-Velez}
  \author[3,7]{Jianhao Yuan}
  \author[3]{Felix Friedrich}
  \author[3]{Jakob Verbeek}
  \author[1,2,5,\ddagger]{Aaron Courville}
  \author[3,\ddagger]{Michal Drozdzal}
  \author[1,3,4,5,\ddagger]{Adriana Romero-Soriano}
  \affiliation[1]{Mila - Québec AI Institute}
  \affiliation[2]{Université de Montréal}
  \affiliation[3]{FAIR at Meta}
  \affiliation[4]{McGill University}
  \affiliation[5]{Canada CIFAR AI chair}
  \affiliation[6]{Columbia University}
  \affiliation[7]{University of Oxford}
  \contribution[\dagger]{Equal contribution}
  \contribution[\ddagger]{Co-senior authors}
  \metadata[Website]{\url{https://featok.github.io/}}
  \correspondence{\email{xiaofeng.zhang@mila.quebec}}

\begin{document}

\begin{abstract}

Discrete image generators learn the distribution of tokens produced by an image tokenizer.
These token representations should be easy for generators to model,
not merely enable faithful image reconstruction.
Yet these tokenizers are commonly trained to reconstruct pixels,
which prioritizes low-level fidelity and may disrupt representation structure
useful for generation.
Motivated by feature-predictive self-supervised learning, 
we introduce feature-loss tokenization (\method{}), which trains discrete tokens by reconstructing pretrained features rather than pixels.
Ablations across SigLIP, SigLIP2, DINOv2, and DINOv3
show better generative performance with \method{}. In class-conditional generation, \method{} advances the quality--compute Pareto frontier, reducing gFDr6 at $256{\times}256$ by up to $29\%$ over the strongest discrete baselines and outperforming the strongest unguided continuous baseline (RAEv2) with $38\times$ less sampling compute.
\method{} also achieves higher human-preference reward-model scores than prior art without reward-based training and surpasses continuous and discrete text-conditional baselines on three benchmarks with only $500$ fine-tuning updates after pre-training.\looseness-1

\end{abstract}

\begin{figure}[h]
    \centering
    \includegraphics[width=\textwidth]{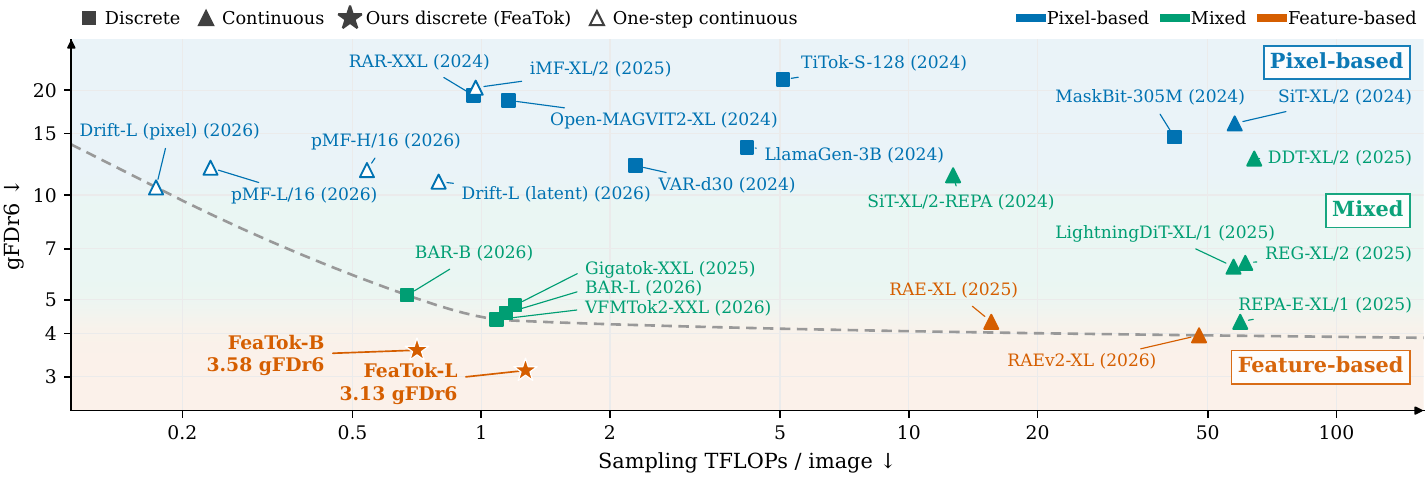}
    \caption{\textbf{Feature-based discrete tokens enable strong generation.}
    \method{} models advance the quality--compute Pareto frontier for image generation.
    We categorize generative image models by how pretrained feature-predictive self-supervised learning representations
    (Def.~\ref{def:features}) shape the generated
    representation or the generator: not at all (pixel-based), together with pixel objectives
    (mixed), or alone (feature-based).
    Generation performance on ImageNet $256\times256$, where all models sample without guidance.
    Both axes are log-scaled.}
    \label{fig:teaser}
\end{figure}

\section{Introduction}
\label{sec:introduction}

Image generation enables machines to create new visual content, supporting applications ranging from creative design to multimodal interaction.
A growing body of work represents images as discrete tokens and models them autoregressively~\citep{llamagen2024,var2024,yu2026bar}.
Discrete tokens share the interface of language models, which makes them a natural choice for unified multimodal models.
They also sample an image with an order of magnitude less compute than continuous diffusion models, as shown in Fig.~\ref{fig:teaser}.\looseness-1

Image generation quality depends on discrete tokens that are both expressive and easy for a generator to model.
Previous discrete tokenizers focus on reconstruction performance, which targets the first property, \emph{expressivity}.
However, generation also depends on the second property, which we call \emph{generative learnability}: how easily a generator fits the token distribution at a given budget.
These two properties do not always align, as observed for both continuous and discrete image tokenizers and generators~\citep{gu2024rethinking,Yao2025,ramanujan2026worse}.

Generative learnability is related to the structure of a representation~\citep{Skorokhodov2025diffusability,chu2026eostok},
which is in turn induced by the training target.
Pixel reconstruction prioritizes high-variance image details over
features useful for perception~\citep{balestriero2024learning}, and
using self-supervised learning (SSL) objectives to predict features instead of pixels yields more semantically meaningful representations~\citep{assran2023self}.
We refer to the representations of encoders trained with SSL objectives in feature space
as \emph{pretrained visual features} (see Def.~\ref{def:features}). Moreover, RAE~\citep{Zheng2025rae} and RAEv2~\citep{Singh2026raev2} show that the benefits of pretrained visual features go beyond representation learning and successfully transfer to image generation. \looseness-1 

\citet{vqkdwang2024vqunderstanding} trained discrete tokens by reconstructing the features of
  image-understanding encoders instead of pixels, and showed that these tokens can outperform
  pixel-trained VQ tokens for image generation.
  Yet the discrete tokenizers that followed still use pixel reconstruction as the central loss,
  and add pretrained visual features with additional losses~\citep{xiong2025gigatok,vqrae2025,yu2026bar,vfmtok2026,chen2026ideal}.
  This raises the question of whether
  \emph{pixel reconstruction loss is essential for modern discrete tokenizer training used for downstream image generation}.
  We argue that it is not.
  Feature objectives should shape both the encoder, pretrained in feature space,
  and the discrete tokens.

We introduce feature-loss tokenization (\textbf{\method{}}), which separates discrete
representation learning from image reconstruction.
Instead of a pixel decoder and a pixel reconstruction loss, a feature decoder that reconstructs
pretrained visual features shapes the discrete tokens.
A separate image decoder is trained with pixel losses on \emph{detached} tokens and only renders generated token sequences, so pixel losses never shape the discrete representation. 
We additionally discretize the encoder's global representation (\textit{e.g.}, CLS token), and generate it before the patch tokens, giving them image-level context. We call this \emph{global-augmented} generative modelling.
Ablations in Sec.~\ref{sec:ablations} show that \method{} has better spatial structure and
generates better than tokenizers trained with pixel or combined losses for all four encoders we
study, and that global-augmented modelling contributes to this improvement.\looseness-1

On class-conditional ImageNet-1k~\citep{imagenet} at \(256\!\times\!256\),
\method{} advances the quality--compute Pareto front for image generation
(Fig.~\ref{fig:teaser}).
We measure quality by gFDr6, the mean Fréchet-distance ratio across six
feature spaces rather than the Inception space alone~\citep{fdloss2026}.
\method{} improves on the strongest discrete baselines by $29\%$ without
guidance and by $25\%$ with it, and outperforms recent continuous
generators at a matched sampling cost in both settings.
The gains grow at \(512\!\times\!512\): relative to the previously best discrete model (BAR~\citep{yu2026bar}), \method{}
reduces gFDr6 by $43\%$ without guidance and $39\%$ with guidance, up from
$32\%$ and $25\%$ at \(256\!\times\!256\), and it also outperforms RAE.
\method{} further achieves higher rewards than discrete and continuous
baselines on three human-preference models~\citep{Kirstain2023Pick,Wu2023HPSv2,hpsv3}
without reward-based training.
Finally, reusing the ImageNet-trained tokenizer, \method{} surpasses the
baselines on three text-to-image benchmarks~\citep{geneval,genaibench,ella}
after $75$K pretraining updates and only $500$ fine-tuning updates.
Discrete tokens shaped by feature losses alone are therefore competitive
with continuous representations while sampling at a fraction of the
cost, which makes them a practical substrate for unified multimodal
models.

\section{Feature-loss tokenization}
\label{sec:method}

In this section, we present \textbf{\method{}}, which learns discrete tokens by reconstructing pretrained visual features.
We use this term to distinguish representations by their pretraining objective.

\begin{definition}
\label{def:features}
We define \emph{pretrained visual features} as representations of encoders learned by self-supervision in feature space,
including language--image contrastive learning~\citep{clip,siglip,siglip2} and self-distillation~\citep{dino2021,dinov2,dinov3,siglip2}.
Encoders trained by pixel reconstruction~\citep{he2022mae}
are excluded.
\end{definition}

We next describe how \method{} uses these features to train the tokenizer with feature losses (Fig.~\ref{fig:method_overview}).

\subsection{Learning discrete tokens with feature loss only}
\label{sec:method_tokenizer}
Conventional discrete tokenizers~\citep{vqvae,dinotok2025,flextok,vfmtok2026,yu2026bar} 
consist of an encoder $E$, a quantizer $Q$, and a pixel decoder $D_{\rm pix}$.
Given an image $x$ partitioned into $N$ patches, the encoder $E$ produces
$N$ feature vectors, one per patch, which the quantizer $Q$ maps to
discrete tokens $z_{1:N}=Q(E(x))$.
The pixel decoder $D_{\rm pix}$ then reconstructs the image as
$\hat{x}=D_{\rm pix}(z_{1:N})$.
Tokenizers are typically trained as an autoencoder with a pixel-level objective
$\mathcal{L}_{\rm pix}$ that combines pixel reconstruction losses
($\ell_1$, $\ell_2$), perceptual losses
(LPIPS~\citep{lpips},   GRAM~\citep{gram}), and adversarial losses~\citep{gan2020goodfellow}. %
Once the tokenizer is trained, an autoregressive model conditioned on class labels or text prompts $c$ learns the distribution $p_\phi(z_{1:N}\mid c)$ of the resulting token sequences using the cross-entropy loss.\looseness-1

\begin{figure*}[t]
\centering
\includegraphics[width=\textwidth]{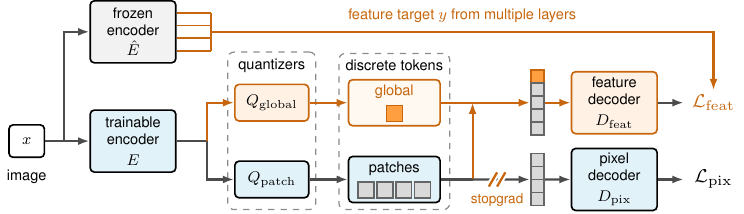}
\caption{\textbf{\method{} tokenizer and generation pipeline.}
Blue denotes the conventional discrete tokenizer and orange our contributions.
Only feature losses shape the discrete tokens: $\mathcal{L}_{\rm feat}$ trains $D_{\rm feat}$ to
reconstruct the multi-layer target $y$ of the frozen encoder $\hat{E}$ from both global and patch
tokens, while $\mathcal{L}_{\rm pix}$ trains only $D_{\rm pix}$ on detached patch tokens.
The generator (not shown) models the global token from $Q_{\rm global}$ together with the patch
tokens from $Q_{\rm patch}$.
}
\label{fig:method_overview}
\end{figure*}

\textbf{Shaping discrete tokens with feature losses.}
\method{} follows the autoencoder-style tokenizer design, but learns the discrete tokens by reconstructing features rather than pixels (see Fig.~\ref{fig:method_overview}).
Concretely, we add a feature decoder $D_{\rm feat}$ that maps the discrete tokens $z_{1:N}$ to a prediction $\hat{y}_{1:N}=D_{\rm feat}(z_{1:N})$ of a feature target $y_{1:N}$.
We train the encoder, quantizer, and feature decoder to match this prediction to the target using a loss $\mathcal{L}_{\rm feat}$ that combines $\ell_1$, $\ell_2$, and cosine distances.

For the feature target, following the insights in \citet{Singh2026raev2}, we use an average of features from multiple layers of the frozen vision encoder $\hat{E}$,
$y_{1:N}(x)=|\mathcal{K}|^{-1}\sum_{k \in \mathcal{K}}\operatorname{LN}_{k}(\hat{E}^{(k)}(x)_{1:N})$,
where $\mathcal{K}$ contains four evenly spaced encoder layers and LayerNorm~\citep{layernorm} normalizes each layer's features before averaging.
This target is intended to combine visual detail from earlier layers with semantic information from later layers,
balancing expressivity and generative learnability.

While we retain the pixel decoder $D_{\rm pix}$ and its pixel-level loss $\mathcal{L}_{\rm{pix}}$ during training, we stop the gradients at the decoder input.
This preserves the ability to decode tokens back into images while preventing the pixel reconstruction losses from shaping the token distribution.

\textbf{Leveraging global tokens.}
In addition to patch features, various vision encoders supply a global representation, \emph{e.g.}, a CLS token in DINO~\citep{dino2021,dinov2,dinov3}
or a global pooling token in SigLIP~\citep{siglip,siglip2}.
Replacing pixel reconstruction with feature reconstruction losses
naturally suggests discretizing this global feature along
with patches.
The benefit of doing so is that it allows us to define a hierarchical generative model
that first generates the global token and then the patch tokens conditioned on it
which simplifies the conditional distributions of the patch tokens by providing a global context.
This global representation is generally ignored by discrete tokenizers
trained using pixel-reconstruction losses,
as the pixel reconstruction over patches does not provide direct supervision on the
global feature discretization.
Prior work follows a similar global-to-local strategy, using learned holistic
tokens~\citep{zheng2025hita}, coarse-to-fine token sequences~\citep{flextok},
or separately generated semantic codes~\citep{DLClavoie}.
In our case, the global token directly discretizes the pretrained encoder's global representation.\looseness-1

Specifically, we use separate quantizers,
$Q_{\rm global}$ and $Q_{\rm patch}$, for global and patch features, respectively.
The feature decoder $D_{\rm feat}$ takes both global and patch tokens and reconstructs
the combined target $y=(y_{\rm global},y_{1:N})$
using the feature loss $\mathcal{L}_{\rm{feat}}$. 
This yields a discrete global token $z_{\rm global}$, which we model jointly with the patch tokens using the following factorization:\looseness-1
\begin{equation}
    p_\phi(z_{\rm global},z_{1:N}\mid c)
    =p_\phi(z_{\rm global}\mid c)\,
     \prod_{i=1}^{N}p_\phi(z_i\mid z_{\rm global},z_{<i},c).
    \label{eq:cls_factorization}
\end{equation}
where $c$ denotes the class label or text prompt.
During inference, the generator first samples the global token
and then samples patch tokens conditioned on both global and external conditions.\looseness-1

\textbf{Combined training objective.}
We train the tokenizer and pixel decoder jointly using the feature and pixel losses described above.
Following prior work on semantic preservation during tokenizer training~\citep{aligntok2025,vqrae2025},
we also include an $\ell_2$ regularizer $\mathcal{L}_{\rm reg}$ between the last-layer features
$h$ and $\hat h$ of the trainable and frozen encoders.
With $z=(z_{\rm global},z_{1:N})$, the complete objective is
\begin{equation}
    \mathcal{L}_{\rm \method{}}
    =\mathbb{E}_{x}\!\left[\underbrace{\mathcal{L}_{\rm feat}\!\left(
        D_{\rm feat}(z),y\right)
        +\lambda_{\rm reg}\mathcal{L}_{\rm reg}(h, \hat h)}_{\text{trains encoder, quantizers, feature decoder}}
    +\underbrace{\lambda_{\rm pix}\mathcal{L}_{\rm pix}\!\left(
        D_{\rm pix}\!\left(\operatorname{\mathrm{sg}}(z_{1:N})\right),x\right)}_{\text{trains only the pixel decoder}}
    \right],
    \label{eq:method_objective}
\end{equation}
where the expectation is over training images $x$.
The weights $\lambda_{\rm reg}$ and $\lambda_{\rm pix}$ balance the loss terms,
and $\mathrm{sg}$ stops gradients at the pixel decoder input.

Once the tokenizer is trained, we keep it fixed and train an autoregressive generator on its tokens using teacher forcing.
We combine the cross-entropy loss for the global token with the average loss over patch tokens, weighted by $\lambda_{\rm global}$ and $\lambda_{\rm patch}$, respectively.
These weights control the emphasis on global and patch prediction independently of the number of patches.\looseness-1

\subsection{Controlled ablations across pretrained vision encoders}
\label{sec:ablations}

\begin{figure*}[t]
\centering
\subfloat[Tokenizer targets.\label{fig:target_patchonly_bars}]{%
  \includegraphics[width=0.33\textwidth]{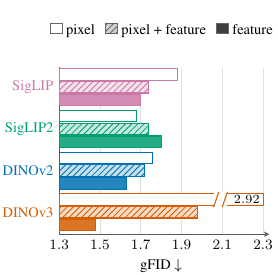}}
\hfill
\subfloat[Spatial dependence.\label{fig:patchonly_target_mi_gfid}]{%
  \includegraphics[width=0.33\textwidth]{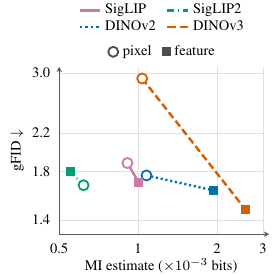}}
\hfill
\subfloat[Global augmentation.\label{fig:global_augmented_bars}]{%
  \includegraphics[width=0.33\textwidth]{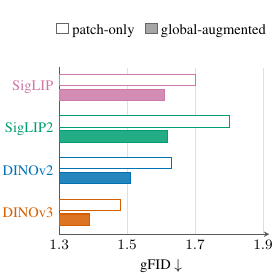}}
\caption{\textbf{Tokenizer-target and generator-type ablations.}
(a) With patch-only generation, feature targets improve gFID for SigLIP, DINOv2, and DINOv3.
The DINOv3 pixel bar has a break to accommodate its gFID of $2.92$.
(b) Spatial dependence between adjacent patch tokens versus patch-only gFID.
The feature target raises the dependence and lowers gFID for SigLIP, DINOv2, and DINOv3. Both axes are log-scaled.
(c) With the feature target fixed, global-augmented generation improves gFID for all four encoders,
achieving better generation than pixel-target baselines.
}
\label{fig:target_generator_ablation}
\end{figure*}

We verify \method{}, feature-loss tokenization, against pixel-loss tokenization
using four pretrained vision encoders: SigLIP~\citep{siglip},
SigLIP2~\citep{siglip2}, DINOv2~\citep{dinov2}, and DINOv3~\citep{dinov3}.
We compare the generation performance of a generative model trained
on ImageNet-1k~\citep{imagenet} images at $256\!\times\!256$ resolution using  different tokenizers.
Details of ablation setups are given in App.~\ref{app:ablations}.
Additionally, App.~\ref{app:mae} shows that
tokenizers built on MAE~\citep{he2022mae}, whose features are learned by pixel reconstruction,
generate poorly under every supervision.\looseness-1

\textbf{Feature targets work better than pixel-only or mixed feature--pixel targets.}
Feature supervision improves the generation over pixel supervision for three
of the four encoders, as shown in Fig.~\ref{fig:target_patchonly_bars}.
With patch-only generation, replacing pixel targets with feature targets
lowers gFID for SigLIP, DINOv2, and DINOv3, while combining pixel and feature losses, \emph{i.e.},
letting gradients from both decoders update the discrete tokens,
gives intermediate results.
Fig.~\ref{fig:patchonly_target_mi_gfid} relates these gFID changes to the spatial dependence
between adjacent discrete patch tokens.
The feature target raises this dependence and lowers gFID for SigLIP, DINOv2, and DINOv3,
showing that the feature target improves the token structure for generation.
SigLIP2 is the exception, possibly because, unlike the other three encoders, it is not trained
only in feature space. Its training adds a text decoder that generates captions of whole images
and of specific regions~\citep{siglip2}.
App.~\ref{app:token_structure} details the estimator and presents other token-structure
measures.\looseness-1

\Needspace*{22\baselineskip}
\begin{wrapfigure}{r}{0.42\textwidth}
\vspace{\dimexpr-\intextsep-1pt\relax}
\centering
\includegraphics[width=0.93698\linewidth]{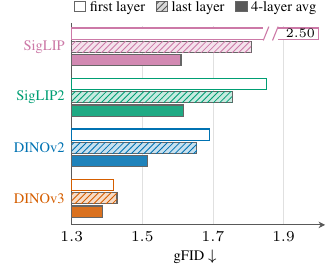}
\caption{\textbf{Feature-target depth.}
First-layer, last-layer and four-layer-average targets with global-augmented generation.
The SigLIP first-layer bar has a break to accommodate its gFID of $2.50$.}
\label{fig:feature_target_depth}
\end{wrapfigure}

\textbf{Global-augmented generation facilitates generative modelling.}
Holding the tokenizer fixed and adding the global conditioning to generative modeling
lowers gFID from $1.70$ to $1.61$ for SigLIP, $1.80$ to $1.62$ for SigLIP2,
$1.63$ to $1.51$ for DINOv2, and $1.48$ to $1.39$ for DINOv3,
as shown in Fig.~\ref{fig:global_augmented_bars}.
With the global token, \method{} with SigLIP2 overtakes
the counterpart trained with pixel reconstruction losses.

\textbf{Leveraging features from multiple layers improves generation.}
The four-layer average gives the best measured gFID for each encoder in
Fig.~\ref{fig:feature_target_depth}, with global-augmented generation throughout.
Compared with the first and last layers, respectively, averaging yields
the best generation performance consistently across all encoders.
Tab.~\ref{tab:target_head_ablation} in App.~\ref{app:ablations} reports
the numbers presented in this section in detail,
and with the accompanying reconstruction results for each ablation.\looseness-1

\WFclear

\section{State-of-the-art image generation with discrete tokens}
\label{sec:experiments}

\begin{table}[t]
\centering
\caption{Comparison with discrete baselines on ImageNet-1k $256\!\times\!256$.
\method{} outperforms all discrete baselines with and without guidance.
Size denotes the generator parameter count.
TFLOPs reports the sampling cost per image without guidance.
Baselines use their released sampling settings, and App.~\ref{app:gfid_prdc} reports
\method{} at gFID-selected settings.
$^{\ddagger}$VAR's unguided generator cost is estimated as half its guided cost.
\textbf{Bold} and \underline{underline} mark the best and second-best values
across all models in each metric column.}
\label{tab:fdr6}
\fdrtablefont
\setlength{\tabcolsep}{0.5pt}
\fdrsetwidth
\begin{tabular}{@{}l@{\hspace{2.7pt}}c r@{\hspace{3pt}}G FFFFFF G FFFFFF@{}}
\toprule
& & & \multicolumn{7}{c}{without guidance} & \multicolumn{7}{c}{with guidance} \\
\cmidrule(lr){4-10}\cmidrule(lr){11-17}
method & size & TFLOPs & \fdrheadg{gFDr6} & \fdrhead{incep.} & \fdrhead{cvnx2} & \fdrhead{dino2} & \fdrhead{mae} & \fdrhead{siglip2} & \fdrhead{clip} & \fdrheadg{gFDr6} & \fdrhead{incep.} & \fdrhead{cvnx2} & \fdrhead{dino2} & \fdrhead{mae} & \fdrhead{siglip2} & \fdrhead{clip} \\
\midrule
\multicolumn{17}{l}{\emph{Pixel-based}} \\
VAR-d30 \citep{var2024} & 2.0B & $^{\ddagger}$2.29 & 12.14 & 3.09 & 5.09 & 11.92 & 10.21 & 22.60 & 19.91 & 6.75 & 1.16 & 1.64 & 5.30 & 7.25 & 11.95 & 13.22 \\
LlamaGen-3B \citep{llamagen2024} & 3.1B & 4.18 & 13.70 & 5.62 & 8.70 & 13.33 & 9.34 & 23.32 & 21.91 & 5.97 & 1.26 & 1.40 & 4.75 & 5.35 & 10.46 & 12.59 \\
TiTok-S-128 \citep{yu2024image} & 0.3B & 5.08 & 21.45 & 5.11 & 10.51 & 22.62 & 14.22 & 41.96 & 34.27 & 15.32 & 1.19 & 1.98 & 14.04 & 13.53 & 30.23 & 30.96 \\
Open-MAGVIT2-XL \citep{luo2024open} & 1.5B & 1.16 & 18.70 & 6.85 & 12.83 & 19.89 & 10.29 & 34.18 & 28.13 & 7.72 & 1.39 & 2.24 & 7.15 & 5.53 & 14.98 & 15.03 \\
MaskBit-305M \citep{weber2024maskbit} & 0.3B & 41.79 & 14.65 & 2.36 & 6.84 & 13.23 & 10.45 & 27.15 & 27.84 & 11.68 & 0.93 & 3.06 & 9.03 & 9.71 & 21.43 & 25.94 \\
RAR-XXL \citep{yu2025randomized} & 1.5B & 0.96 & 19.31 & 2.50 & 7.07 & 16.95 & 12.50 & 32.90 & 43.93 & 15.07 & 0.88 & 2.71 & 10.23 & 11.39 & 24.73 & 40.48 \\
\midrule
\multicolumn{17}{l}{\emph{Mixed}} \\
Gigatok-XXL \citep{xiong2025gigatok} & 1.4B & 1.20 & 4.82 & 1.18 & 1.24 & 3.78 & 4.84 & 8.78 & 9.10 & 4.24 & 1.72 & 1.44 & 2.96 & 4.30 & 7.27 & 7.76 \\
VFMTok2-XXL \citep{vfmtok2026} & 1.4B & 1.09 & 4.38 & 1.22 & 0.99 & 3.25 & 4.82 & 7.73 & 8.29 & 4.37 & 2.32 & 1.79 & 2.98 & 4.67 & 7.01 & 7.44 \\
BAR-B \citep{yu2026bar} & 0.4B & 0.67 & 5.15 & 1.01 & 1.70 & 5.21 & 5.82 & 9.40 & 7.73 & 4.89 & \underline{0.69} & 0.95 & 4.63 & 5.89 & 9.32 & 7.86 \\
BAR-L \citep{yu2026bar} & 1.1B & 1.14 & 4.57 & \textbf{0.83} & 1.38 & 4.59 & 5.22 & 8.14 & 7.26 & 4.11 & \textbf{0.61} & 0.79 & 3.72 & 5.02 & 7.49 & 7.06 \\
\midrule
\multicolumn{17}{l}{\emph{Feature-based}} \\
\textbf{\method{}-B-SigLIP} & 0.4B & 0.71 & 4.46 & 1.58 & 0.89 & 4.49 & 5.86 & 8.00 & 5.94 & 4.14 & 1.28 & 0.75 & 3.96 & 5.51 & 7.71 & 5.63 \\
\textbf{\method{}-B-SigLIP2} & 0.4B & 0.71 & 4.00 & 1.42 & 0.98 & 3.79 & 4.87 & 7.52 & 5.41 & 3.46 & 1.30 & 0.86 & 3.09 & 4.14 & 6.29 & 5.05 \\
\textbf{\method{}-B-DINOv2} & 0.4B & 0.71 & 4.40 & 0.92 & 1.36 & 4.12 & 4.93 & 8.31 & 6.74 & 3.56 & 1.15 & 0.89 & 3.00 & 4.21 & 6.39 & 5.71 \\
\textbf{\method{}-B-DINOv3} & 0.4B & 0.71 & 3.58 & 1.09 & 0.82 & 3.30 & 4.62 & 6.50 & 5.15 & 3.39 & 1.06 & \textbf{0.70} & 3.00 & 4.43 & 6.28 & \underline{4.91} \\
\midrule
\textbf{\method{}-L-SigLIP} & 1.1B & 1.27 & 4.00 & 1.49 & \underline{0.80} & 3.94 & 5.09 & 7.12 & 5.56 & 3.92 & 1.52 & 0.85 & 3.80 & 4.82 & 7.01 & 5.54 \\
\textbf{\method{}-L-SigLIP2} & 1.1B & 1.27 & \underline{3.45} & 1.11 & 0.82 & 3.37 & \underline{4.04} & \underline{6.29} & \underline{5.11} & 3.27 & 1.12 & 0.78 & 2.98 & \underline{3.75} & 6.07 & 4.94 \\
\textbf{\method{}-L-DINOv2} & 1.1B & 1.27 & 3.47 & \underline{0.87} & 0.83 & \underline{3.11} & 4.13 & 6.33 & 5.52 & \underline{3.20} & 0.95 & \underline{0.74} & \textbf{2.69} & 3.77 & \underline{5.79} & 5.28 \\
\textbf{\method{}-L-DINOv3} & 1.1B & 1.27 & \textbf{3.13} & 0.96 & \textbf{0.68} & \textbf{2.95} & \textbf{4.00} & \textbf{5.57} & \textbf{4.62} & \textbf{3.08} & 1.09 & 0.82 & \underline{2.93} & \textbf{3.73} & \textbf{5.28} & \textbf{4.65} \\
\bottomrule
\end{tabular}
\end{table}

To demonstrate the effectiveness of our approach, we evaluate it on ImageNet-1k~\citep{imagenet}
at $256\!\times\!256$ and $512\!\times\!512$ resolution, as well as text-to-image (T2I) generation.
Our models achieve state-of-the-art performance among methods using discrete tokenizers
and outperform recent continuous generative models at a matched inference cost on ImageNet-1k $256\!\times\!256$.
\method{} achieves better performance on ImageNet-1k $512\!\times\!512$ than both
continuous and discrete baselines. On T2I, \method{} outperforms RAEv2~\citep{Singh2026raev2} after
a short fine-tuning on three T2I benchmarks,
including GenEval~\citep{geneval}, DPG-Bench~\citep{ella}, and GenAI-Bench~\citep{genaibench}.\looseness-1

\subsection{ImageNet-1k generation at $256\!\times\!256$}
\label{sec:exp_main}

We train our $256\!\times\!256$ \method{} tokenizers on ImageNet-1k using SigLIP~\citep{siglip},
SigLIP2~\citep{siglip2}, DINOv2~\citep{dinov2}, and DINOv3~\citep{dinov3} encoders,
with a vocabulary size of $65,536$.
For each tokenizer, we train the same autoregressive generator as BAR~\citep{yu2026bar}, the previous
state-of-the-art discrete model for ImageNet-1k, with our global-augmented modelling.
App.~\ref{app:tokenizer_details} specifies the implementation details including  vision encoder versions, tokenizer and generator instantiations, \emph{etc.}\looseness-1

Fr\'echet distance in the Inception space can misrank modern generative models, and
measuring it in stronger feature spaces such as DINOv2 agrees better with human
judgments~\citep{dinov2FD}.
We therefore report gFDr6, the mean Fr\'echet-distance ratio across
Inception-v3~\citep{inceptionv3}, ConvNeXt~V2~\citep{convnextv2},
DINOv2~\citep{dinov2}, MAE~\citep{he2022mae}, SigLIP2~\citep{siglip2},
and CLIP~\citep{clip} feature spaces, following
\citet{fdloss2026}.
In App.~\ref{app:sampling} we detail the sampling configurations for each model.
In App.~\ref{app:gfid_prdc} we report gFID~\citep{FID} and precision, recall~\citep{improvedprecisionrecall},
density and coverage~\citep{densitycoverage} in Inception space as complementary diagnostics. 
Unless stated otherwise, baselines are evaluated by us using their released setups.\looseness-1

\textbf{\method{} outperforms discrete models. \ }
In Tab.~\ref{tab:fdr6} we compare \method{} using different vision encoders
and two model sizes to existing discrete image generative models in terms of gFDr6
and its per-feature metrics.
All four FeaTok-L variants outperform the strongest prior discrete baseline in gFDr6, showing consistent gains across four pretrained visual feature encoders satisfying Def.~\ref{def:features}. 
\method{}-DINOv3 is the strongest discrete model at both generator scales,
with and without guidance.
The largest improvements appear without guidance, where we reduce the prior
baseline gFDr6 of $4.38$ from VFMTok2-XXL~\citep{vfmtok2026} to $3.13$.
With guidance, we reduce the $4.11$ gFDr6 of BAR-L to $3.08$.
Increasing the generator from base to large consistently improves gFDr6; for DINOv3, it decreases from $3.58$ to $3.13$ without guidance and from $3.39$ to $3.08$ with guidance.
Importantly, these generation gains do not coincide with improved reconstruction: rFID remains nearly constant across \method{} tokenizers ($0.92$–$1.01$; App.~\ref{sec:exp_recon}), reinforcing that reconstruction fidelity alone does not explain generative performance.
Figs.~\ref{fig:qual-a}--\ref{fig:qual-d} in App.~\ref{app:qualitative} show
qualitative samples.\looseness-1

\begin{table}[t]
\centering
\caption{Comparison with baselines using continuous features on ImageNet-1k at $256\times256$.
Without guidance, \method{} achieves lower gFDr6 than all listed baselines at a lower sampling cost.
Size denotes the generator parameter count, and TFLOPs reports the sampling cost per image without guidance.
\textbf{Bold} and \underline{underline} mark the best and second-best values in each metric column.
Guided continuous-baseline results are transcribed from~\citet{Singh2026raev2}.}
\label{tab:fdr6_continuous}
\fdrtablefont
\setlength{\tabcolsep}{0.5pt}
\fdrsetwidth
\begin{tabular}{@{}l@{\hspace{2.7pt}}c r@{\hspace{3pt}}G FFFFFF G FFFFFF@{}}
\toprule
& & & \multicolumn{7}{c}{without guidance} & \multicolumn{7}{c}{with guidance} \\
\cmidrule(lr){4-10}\cmidrule(lr){11-17}
method & size & TFLOPs & \fdrheadg{gFDr6} & \fdrhead{incep.} & \fdrhead{cvnx2} & \fdrhead{dino2} & \fdrhead{mae} & \fdrhead{siglip2} & \fdrhead{clip} & \fdrheadg{gFDr6} & \fdrhead{incep.} & \fdrhead{cvnx2} & \fdrhead{dino2} & \fdrhead{mae} & \fdrhead{siglip2} & \fdrhead{clip} \\
\midrule
\multicolumn{17}{l}{\emph{Pixel-based}} \\
SiT-XL/2 \citep{Ma2024} & 675M & 57.82 & 16.05 & 5.24 & 8.34 & 16.77 & 9.40 & 30.05 & 26.52 & 8.44 & 1.26 & 2.02 & 7.89 & 5.62 & 16.14 & 17.69 \\
\midrule
\multicolumn{17}{l}{\emph{Mixed}} \\
SiT-XL/2-REPA \citep{Yu2025repa} & 675M & 12.69 & 11.39 & 3.57 & 5.10 & 10.81 & 7.34 & 20.70 & 20.80 & 5.45 & 0.85 & 1.22 & 4.27 & 3.85 & 9.87 & 12.65 \\
LightningDiT-XL/1 \citep{Yao2025} & 675M & 57.45 & 6.22 & 1.33 & 2.42 & 5.77 & 4.08 & 11.77 & 11.95 & 4.57 & 0.85 & 1.09 & 3.76 & 3.02 & 8.47 & 10.21 \\
DDT-XL/2 \citep{Wang2025ddt} & 675M & 64.25 & 12.72 & 3.86 & 5.52 & 11.89 & 8.20 & 23.06 & 23.78 & 5.70 & 0.75 & 1.02 & 4.26 & 4.11 & 10.16 & 13.86 \\
REPA-E-XL/1 \citep{repae2025} & 675M & 59.57 & 4.32 & 1.03 & 2.35 & 3.81 & \textbf{3.48} & 7.38 & 7.84 & \underline{3.04} & 0.70 & 1.28 & 2.44 & \textbf{2.52} & 5.04 & 6.28 \\
REG-XL/2 \citep{reg2025} & 675M & 61.22 & 6.38 & 1.05 & 1.74 & 4.98 & 4.96 & 11.72 & 13.81 & 4.64 & 0.92 & 1.14 & 3.45 & 3.02 & 8.42 & 10.86 \\
\midrule
\multicolumn{17}{l}{\emph{Feature-based}} \\
RAE-XL \citep{Zheng2025rae} & 839M & 15.59 & 4.32 & \textbf{0.91} & 2.23 & 3.01 & 4.79 & \underline{5.66} & 9.36 & 3.26 & \underline{0.69} & 1.79 & \underline{2.11} & 3.30 & \underline{3.79} & 7.87 \\
RAEv2-XL \citep{Singh2026raev2} & 875M & 47.72 & \underline{3.95} & 1.00 & \underline{1.41} & \textbf{2.39} & 5.83 & \underline{5.66} & \underline{7.41} & \textbf{2.17} & \textbf{0.64} & \textbf{0.77} & \textbf{1.15} & \underline{2.67} & \textbf{2.54} & \underline{5.21} \\ \midrule
\textbf{\method{}-L-DINOv3} & 1.1B & 1.27 & \textbf{3.13} & \underline{0.96} & \textbf{0.68} & \underline{2.95} & \underline{4.00} & \textbf{5.57} & \textbf{4.62} & 3.08 & 1.09 & \underline{0.82} & 2.93 & 3.73 & 5.28 & \textbf{4.65} \\
\bottomrule
\end{tabular}
\end{table}

\textbf{\method{} is competitive with continuous models at lower sampling cost.}
Tab.~\ref{tab:fdr6_continuous} compares against continuous baselines at their
standard sampling settings.
Without guidance, \method{}-L-DINOv3 achieves gFDr6 $3.13$, outperforming
RAEv2 ($3.95$) while using $1.27$ rather than $47.72$ TFLOPs per image.
With guidance, our $3.08$ is competitive with
RAE-XL ($3.26$)~\citep{Zheng2025rae} and REPA-E ($3.04$)~\citep{repae2025},
although RAEv2 retains the best gFDr6 ($2.17$).\looseness-1
 
To compare at similar sampling costs, Tab.~\ref{tab:fdr6_shortened_raev2}
evaluates RAEv2 with fewer sampling steps, tuning its internal guidance at each
step count for the best gFDr6 (App.~\ref{app:sampling}).
With guidance, \method{} achieves gFDr6 $3.08$ at $2.31$ TFLOPs per image,
compared with $3.43$ at $3.07$ TFLOPs for six-step RAEv2.
RAEv2 is not trained to perform few-step sampling.
We therefore also compare with Drift~\citep{driftmodel},
iMF~\citep{imfimprovedmeanflow}, and pMF~\citep{pmfpixelmeanflow},
which are explicitly trained for one-step generation.
Without guidance, their gFDr6 values range from $10.51$ to $20.34$, and even the best
guided value is $6.87$, whereas unguided \method{} achieves $3.13$.
App.~\ref{app:cost} provides detailed cost measurements, including FLOPs, wall-clock
throughput, and peak memory.\looseness-1

\begin{table}[t]
\centering
\caption{Generation with matched inference cost on ImageNet $256\times256$.
TFLOPs are computed without guidance per image.
Guided one-step results are from~\citet{fdloss2026}.
Drift models share the same setup w and w/o guidance.
\textbf{Bold} and \underline{underline} mark the best and second-best values in each metric column.}
\label{tab:fdr6_shortened_raev2}
\fdrtablefont
\setlength{\tabcolsep}{0.5pt}
\fdrsetwidth
\begin{tabular}{@{}l@{\hspace{2.7pt}}c@{\hspace{3pt}}r@{\hspace{3pt}}G FFFFFF G FFFFFF@{}}
\toprule
& & & \multicolumn{7}{c}{without guidance} & \multicolumn{7}{c}{with guidance} \\
\cmidrule(lr){4-10}\cmidrule(lr){11-17}
method & steps & TFLOPs & \fdrheadg{gFDr6} & \fdrhead{incep.} & \fdrhead{cvnx2} & \fdrhead{dino2} & \fdrhead{mae} & \fdrhead{siglip2} & \fdrhead{clip} & \fdrheadg{gFDr6} & \fdrhead{incep.} & \fdrhead{cvnx2} & \fdrhead{dino2} & \fdrhead{mae} & \fdrhead{siglip2} & \fdrhead{clip} \\
\midrule
\multicolumn{17}{l}{\emph{Baselines with continuous features}} \\
RAEv2-XL~\citep{Singh2026raev2} & 3 & 1.65 & 38.90 & 14.12 & 11.69 & 16.68 & 69.14 & 78.74 & 43.06 & 8.95 & 2.98 & 3.15 & 6.41 & 10.49 & 16.15 & 14.50 \\
RAEv2-XL~\citep{Singh2026raev2} & 6 & 3.07 & \underline{8.49} & 2.52 & 2.90 & \underline{5.32} & 12.75 & \underline{14.45} & \underline{13.01} & \underline{3.43} & 1.27 & 1.29 & \textbf{2.11} & \underline{4.54} & \textbf{4.92} & \underline{6.42} \\
\midrule
\multicolumn{17}{l}{\emph{Baselines of one-step flow models}} \\
Drift-L (latent)~\citep{driftmodel} & 1 & 0.79 & 10.92 & \underline{0.91} & 2.03 & 10.35 & \underline{6.51} & 24.12 & 21.59 & 10.92 & \underline{0.91} & 2.03 & 10.35 & 6.51 & 24.12 & 21.59 \\
Drift-L (pixel)~\citep{driftmodel} & 1 & 0.17 & 10.51 & \textbf{0.85} & \underline{0.73} & 6.33 & \underline{6.51} & 22.13 & 26.49 & 10.51 & \textbf{0.85} & \textbf{0.73} & 6.33 & 6.51 & 22.13 & 26.49 \\
iMF-XL/2~\citep{imfimprovedmeanflow} & 1 & 0.97 & 20.34 & 6.33 & 9.26 & 19.10 & 13.28 & 41.90 & 32.15 & 8.39 & 1.09 & 1.72 & 7.30 & 6.09 & 17.02 & 17.14 \\
pMF-L/16~\citep{pmfpixelmeanflow} & 1 & 0.23 & 11.97 & 1.86 & 1.68 & 8.76 & 13.12 & 27.75 & 18.67 & 9.09 & 1.62 & 1.36 & 6.70 & 9.72 & 20.34 & 14.81 \\
pMF-H/16~\citep{pmfpixelmeanflow} & 1 & 0.54 & 11.79 & 2.37 & 2.38 & 9.58 & 10.28 & 27.79 & 18.34 & 6.87 & 1.37 & 1.15 & 5.43 & 6.25 & 15.33 & 11.68 \\
\midrule
\textbf{\method{}-L-DINOv3} & 256 & 1.27 & \textbf{3.13} & 0.96 & \textbf{0.68} & \textbf{2.95} & \textbf{4.00} & \textbf{5.57} & \textbf{4.62} & \textbf{3.08} & 1.09 & \underline{0.82} & \underline{2.93} & \textbf{3.73} & \underline{5.28} & \textbf{4.65} \\
\bottomrule
\end{tabular}
\end{table}

\begin{figure}[t]
\centering
\includegraphics[width=\textwidth]{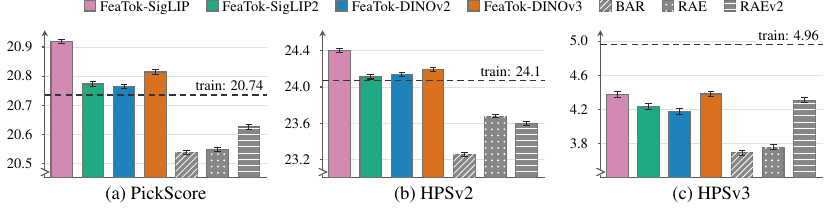}
\caption{Human-preference rewards on $50$K generated ImageNet images
with $95\%$ confidence intervals. \method{} achieves the highest rewards
compared to baselines, and also closer to the rewards measured on the training set.
Dashed lines represent the mean of the rewards over training-data.
HPSv2 scores are displayed multiplied by $100$.
\method{} uses large generators sampled with guidance.\looseness-1
}
\label{fig:reward}
\end{figure}

\textbf{\method{} reaches higher human-preference rewards, closer to those of the training data.}
Fig.~\ref{fig:reward} compares generated and real ImageNet-1k data using PickScore~\citep{Kirstain2023Pick},
HPSv2~\citep{Wu2023HPSv2}, and HPSv3~\citep{hpsv3}, all learned from human preference comparisons.
These scores
are automatic proxies for human evaluations, and 
complement the distributional measurements in Tabs.~\ref{tab:fdr6} and~\ref{tab:fdr6_continuous}.
\method{} performs better than baselines both in terms of the raw value of the rewards and the closeness to the training data.
On PickScore and HPSv2, the SigLIP2, DINOv2, and DINOv3 variants are closer to the real-data reference than every
baseline.
On the more discriminative HPSv3,
\method{}-DINOv3 and \method{}-SigLIP have the
reward scores closest to those of the training set,
followed by RAEv2, all substantially closer than BAR-L and RAE.\looseness-1

\subsection{\method{} outperforms BAR-L and RAE at $512\!\times\!512$}
\label{sec:exp_512}
Following~\citet{yu2026bar},
we train a tokenizer for ImageNet-1k generation at $512\!\times\!512$ with the
DINOv3 vision encoder, using  vocabulary size $1,024$ per token. Implementation details are in Apps.~\ref{app:tokenizer_details} and \ref{app:generator}.
We show the gFDr6 results for
$512$ resolution in Tab.~\ref{tab:res512}.
The benefit of \method{} over baseline is larger at higher resolution. 
\method{}-L-DINOv3 reduces the unguided gFDr6 of BAR-L from
$3.82$ to $2.17$ and the guided result from $2.94$ to $1.80$.
These are reductions of $43\%$ and $39\%$, compared with $32\%$ and $25\%$ for the
corresponding large-scale comparison at 256 resolution.
Our method outperforms BAR-L in all feature spaces without guidance,
and in all feature spaces except Inception with guidance.
\method{} also outperforms RAE in gFDr6, with $2.17$ against $3.03$ without guidance and $1.80$
against $2.13$ with guidance, although RAE is better in the Inception and SigLIP2 spaces in both
settings.
RAEv2 does not provide a model at $512$ resolution.\looseness-1

\begin{table}[t]
\centering
\caption{Generation on ImageNet-1k $512\times512$.
\method{} outperforms BAR-L and RAE with and without guidance.
\textbf{Bold} and \underline{underline} mark the best and second-best values in each metric column.
$^{\dagger}$RAE uses autoguidance instead of classifier-free guidance, and its unguided row uses
the released configuration with autoguidance disabled.}
\label{tab:res512}
\fontsize{7.4}{8.4}\selectfont
\setlength{\tabcolsep}{0.5pt}
\fdrsetwidth
\begin{tabular*}{\textwidth}{@{\extracolsep{\fill}}l G FFFFFF G FFFFFF@{}}
\toprule
& \multicolumn{7}{c}{without guidance} & \multicolumn{7}{c}{with guidance} \\
\cmidrule(lr){2-8}\cmidrule(lr){9-15}
method & \fdrheadg{gFDr6} & \fdrhead{incep.} & \fdrhead{cvnx2} & \fdrhead{dino2} & \fdrhead{mae} & \fdrhead{siglip2} & \fdrhead{clip} & \fdrheadg{gFDr6} & \fdrhead{incep.} & \fdrhead{cvnx2} & \fdrhead{dino2} & \fdrhead{mae} & \fdrhead{siglip2} & \fdrhead{clip} \\
\midrule
\multicolumn{15}{l}{\emph{Baselines}} \\
RAE-XL$^{\dagger}$~\citep{Zheng2025rae} & \underline{3.03} & \textbf{0.97} & \underline{1.42} & \underline{2.07} & 3.11 & \textbf{4.40} & 6.19 & \underline{2.13} & \textbf{0.75} & 1.10 & \underline{1.34} & \underline{2.02} & \textbf{2.74} & \underline{4.84} \\
BAR-L \citep{yu2026bar} & 3.82 & 1.19 & 1.77 & 2.86 & \underline{2.80} & 8.98 & \underline{5.30} & 2.94 & \underline{0.78} & \underline{1.02} & 2.24 & 2.54 & 5.81 & 5.22 \\
\midrule
\textbf{\method{}-L-DINOv3} & \textbf{2.17} & \underline{1.16} & \textbf{0.74} & \textbf{1.41} & \textbf{1.69} & \underline{4.63} & \textbf{3.38} & \textbf{1.80} & 0.94 & \textbf{0.65} & \textbf{1.26} & \textbf{1.30} & \underline{3.51} & \textbf{3.17} \\
\bottomrule
\end{tabular*}
\end{table}

\begin{table}[t]
\centering
\caption{Text-to-image results after pre-training and fine-tuning.
\method{} is broadly comparable to RAEv2, and surpasses all baselines
after 500 fine-tuning updates.
Continuous-latent baseline results are taken from~\citet{Singh2026raev2}.
}
\label{tab:t2i}
\setlength{\tabcolsep}{3pt}
\fontsize{8}{8}\selectfont
\begin{tabular*}{\textwidth}{@{\extracolsep{\fill}}lcccccc@{}}
\toprule
& \multicolumn{3}{c}{after pre-training} & \multicolumn{3}{c}{after fine-tuning} \\
\cmidrule(lr){2-4}\cmidrule(lr){5-7}
model & GenEval~$\uparrow$ & GenAI~$\uparrow$ & DPG~$\uparrow$ & GenEval~$\uparrow$ & GenAI~$\uparrow$ & DPG~$\uparrow$ \\
\midrule
\multicolumn{7}{l}{\emph{continuous models}} \\
Flux-VAE \citep{flux2024} & 0.417 & 0.573 & 77.6 & 0.783 & 0.639 & 79.2 \\
RAE \citep{Zheng2025rae} & 0.584 & 0.632 & \underline{80.1} & 0.815 & 0.672 & 80.6 \\
RAEv2 \citep{Singh2026raev2} & \textbf{0.624} & 0.638 & \textbf{81.7} & \underline{0.827} & 0.680 & \underline{82.3} \\
\midrule
\multicolumn{7}{l}{\emph{discrete models}} \\
BAR \citep{yu2026bar} & 0.564 & \underline{0.664} & 79.8 & 0.817 & \underline{0.688} & 81.7 \\
\textbf{\method{}} & \underline{0.611} & \textbf{0.668} & 79.8 & \textbf{0.837} & \textbf{0.698} & \textbf{83.4} \\
\bottomrule
\end{tabular*}
\end{table}

\subsection{Extension to text-to-image generation}
\label{sec:exp_t2i}
Finally, we demonstrate the usefulness of \method{} by applying it to text-to-image (T2I) generation.
Following RAEv2~\citep{Singh2026raev2},
we use BLIP-3o datasets~\citep{blip3o} (Long-Caption, Short-Caption, and JourneyDB) for pre-training
and BLIP-3o-60K for fine-tuning.
We replace the class conditioning in \method{} with
a sequence of text embeddings provided by Qwen3-0.6B~\citep{qwen3} to enable T2I generation,
re-using our \method{}-SigLIP2 tokenizer trained on ImageNet-1k.
We pre-train 75k updates with batch size $2048$, while
baselines use 150k updates with batch size $1024$, giving the same total number
of training examples processed.
Then we further fine-tune the pre-trained model for $500$ updates,
while baselines are fine-tuned for approximately 3k updates.\looseness-1

\begin{figure}[t]
\centering
\includegraphics[width=\textwidth]{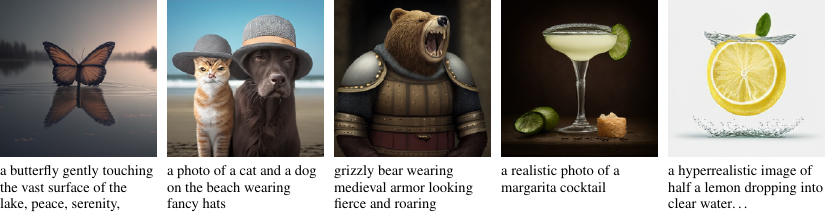}
\caption{Text-to-image samples with abbreviated prompts.
Full prompts are available in Tab.~\ref{tab:mjhq-prompts}.}

\label{fig:t2i-main-qual}
\end{figure}

After pre-training, \method{} is broadly comparable to RAEv2. It is slightly lower
on GenEval~\citep{geneval} ($0.611$ versus $0.624$)
and DPG-Bench~\citep{ella} ($79.8$ versus $81.7$),
but higher on GenAI-Bench~\citep{genaibench}
($0.668$ versus $0.638$).
After only $500$ additional updates, \method{} exceeds RAEv2
on all three benchmarks: $0.837$ versus
$0.827$ on GenEval, $0.698$ versus $0.680$ on GenAI-Bench, and $83.4$ versus $82.3$ on
DPG-Bench.
The comparison with BAR~\citep{yu2026bar} isolates the effect of the tokenizer directly, whose tokenizer is trained with pixel losses and a feature-alignment loss using SigLIP2.
After pre-training, \method{} is ahead of BAR on GenEval, slightly ahead on
GenAI-Bench, and tied on DPG-Bench.
After fine-tuning, all three benchmarks rank \method{} above BAR, with $0.837$ versus $0.817$ on
GenEval, $0.698$ versus $0.688$ on GenAI-Bench, and $83.4$ versus $81.7$ on DPG-Bench.
The advantage of feature-loss tokens over mixed tokens thus carries over from class-conditional
to text-to-image generation.
We provide several T2I qualitative examples in Fig.~\ref{fig:t2i-main-qual}.
More qualitative samples are presented in App.~\ref{sec:qual-t2i} (Fig.~\ref{fig:mjhq-qual}).

\section{Related work}
\label{sec:related_work}

\textbf{Pretrained representations for continuous generation.}
Continuous generators show that the pretraining objective of the features matters.
RAE~\citep{Zheng2025rae} and RAEv2~\citep{Singh2026raev2} train diffusion models in the feature
spaces of pretrained vision encoders, paired with learned image decoders, and RAE finds that MAE
features~\citep{he2022mae} reconstruct best but generate worse than DINOv2 features.
REPA~\citep{Yu2025repa} aligns intermediate SiT~\citep{Ma2024} features with pretrained
representations, where MAE is the weakest of the tested targets, and DDT~\citep{Wang2025ddt}
adopts this alignment.
REG~\citep{reg2025} denoises a pretrained class token together with the image latents,
VA-VAE~\citep{Yao2025} aligns VAE latents with a pretrained vision encoder, and
REPA-E~\citep{repae2025} tunes the VAE and the diffusion model end to end with the REPA loss.

\textbf{Discrete tokenization with pretrained representations.}
VQ-VAE~\citep{vqvae} learns discrete image tokens through pixel reconstruction, and later
tokenizers simplify the quantizer~\citep{magvit2,fsq2024}.
\citet{vqkdwang2024vqunderstanding} train tokenizers to reconstruct the features
from image understanding encoders, and in autoregressive generation
all but the MAE teacher outperform pixel-trained VQGAN tokens.
Recent tokenizers keep pixel reconstruction as the central objective and add pretrained encoder
features through feature alignment~\citep{yu2026bar}, semantic regularization~\citep{xiong2025gigatok},
self-distillation~\citep{vqrae2025}, or feature reconstruction from several encoder
depths~\citep{dinotok2025,chen2026ideal,vfmtok,vfmtok2026}.
TA-Tok~\citep{tatok} trains feature-reconstructed tokens for unified multimodal models, but decodes
images through pixel-trained VQ tokens or latents.
In \method{}, only pretrained visual features (Def.~\ref{def:features}) shape the tokens, which
reconstruct a multi-layer target and include a quantized global token.

\section{Conclusion}
\label{sec:conclusion}
We presented \method{}, a discrete image tokenizer whose tokens are shaped only by feature
reconstruction, while a pixel decoder trained on detached tokens renders them into images.
We also discretize the encoder's global representation and model it together with the patch tokens.
Across four pretrained encoders, \method{} generates better than tokenizers trained with pixel
or combined losses.
\method{} achieves the best gFDr6 among the evaluated discrete generators on ImageNet at $256\!\times\!256$ and
$512\!\times\!512$, outperforms continuous generators at a matched sampling cost, achieves higher
human-preference reward-model scores, and transfers to text-to-image generation after pretraining
and only $500$ fine-tuning updates.
We believe these results are a step toward effective discrete representations for unified
multimodal modeling, and that the low sampling cost of \method{} makes it attractive for scaling
multimodal training.\looseness-1

\clearpage
\appendix
\section{Tokenizer implementation details}
\label{app:tokenizer_details}

This section specifies the final \method{} tokenizer used in Sec.~\ref{sec:experiments}.
We follow the notation presented in Sec.~\ref{sec:method}.
We follow BAR~\citep{yu2026bar}, the previous state-of-the-art method, in our tokenizer
and generator training.
BAR uses bit representations for its tokens and an autoregressive model with a masked
modeling head to predict the bits within each token.
Ablation protocols and variants are described in App.~\ref{app:ablations}.

\textbf{Which pretrained encoders we use.}
\method{} uses SigLIP-L, SigLIP2-SO400M, DINOv2-L with registers, or DINOv3-L.
Tab.~\ref{tab:encoders} identifies their pretrained checkpoints.
Note that DINOv2-L uses a patch size of $14$, whereas the other encoders use a patch size of $16$.
For DINOv2-L, we resize the image to $224\times224$ so that all encoders
encode each image into $16\times16$ patches.

\begin{table}[h]
\centering
\caption{Pretrained source encoders and checkpoints, shared by the frozen target
encoder $\hat E$ and the initialization of the trainable encoder $E$.}
\label{tab:encoders}
\scriptsize
\setlength{\tabcolsep}{5pt}
\begin{tabular}{@{}l l l r r@{}}
\toprule
name in main text & variant & checkpoint & hidden dim & patch size \\
\midrule
SigLIP & ViT-L/16 & \texttt{google/siglip-large-patch16-256} & 1024 & 16 \\
SigLIP2 & SO400M/16, NaFlex & \texttt{google/siglip2-so400m-patch16-naflex} & 1152 & 16 \\
DINOv2 & ViT-L/14, registers & \texttt{facebook/dinov2-with-registers-large} & 1024 & 14 \\
DINOv3 & ViT-L/16         & \texttt{facebook/dinov3-vitl16-pretrain-lvd1689m} & 1024 & 16 \\
\bottomrule
\end{tabular}
\end{table}

\textbf{Quantizer implementation.}
We follow BAR and use finite scalar quantization (FSQ)~\citep{fsq2024}
with $16$ $2$-level slots
for tokenizers trained on ImageNet-1k at $256\times256$.
A trainable input projector
$P_{{\rm in},\theta}:\mathbb{R}^{D}\rightarrow\mathbb{R}^{d}$ maps the adapted backbone
features $h$ to $d=16$ scalar latents, $u_i=P_{{\rm in},\theta}(E^{(L)}(x)_i)$, where
$L$ denotes the final encoder layer and $i$ indexes the $N$ image patches. These latents are
quantized coordinatewise to two levels. For coordinate $\ell$ with output levels
$a_\ell^-<a_\ell^+$ and threshold $b_\ell$,
\begin{equation}
    B_{i\ell}=\mathbf{1}[u_{i\ell}\geq b_\ell],
    \qquad
    q_{i\ell}=a_\ell^-+(a_\ell^+-a_\ell^-)B_{i\ell}.
    \label{eq:two_level_fsq}
\end{equation}
The bit vector $B_i\in\{0,1\}^{d}$ indexes one of $2^{16}$ patch tokens. We use the
straight-through variable $q_i^{\rm st}=u_i+\operatorname{sg}(q_i-u_i)$, where
$\operatorname{sg}$ is stop-gradient.
The feature decoder reads the straight-through codes through its own input projections, which
carry the feature-loss gradient to $P_{{\rm in},\theta}$ and the encoder.
A separate trainable output projector $P_{{\rm out},\theta}$ maps the codes to the pixel-decoder
width and belongs to the pixel branch only.
Its input is detached, so the two decoders share only the discrete codes.
Both projections are learned; only the coordinatewise rounding rule is fixed.
Other differentiable
quantizers can replace this module without changing the \method{} framework.

\textbf{Discrete global token.}
In addition to its $N$ patch features, $E$ emits a global feature.
A separate global quantizer $Q_{\rm{global}}$ maps this feature to a $16$-bit global code
$z_{\rm global}$, which is prepended to the patch codes $z_{1:N}$. Given the external class label $c$,
the generator models $p(z_{\rm global},z_{1:N}\mid c)$ and samples $z_{\rm global}$ first, so it
supplies global context to the patch tokens.
The feature decoder reconstructs both global and patch feature targets,
so the global code receives direct reconstruction supervision.
The patch feature target is the average of representations from four layers,
whereas the global target is the last-layer representation.
The pixel decoder consumes only the sampled patch tokens, allowing a fair comparison
between feature-loss and pixel-loss targets.

\textbf{Architecture and training.}
The trainable encoder $E$ and the frozen encoder $\hat E$ are initialized from the same
pretrained checkpoint (Tab.~\ref{tab:encoders}).
The feature decoder $D_{\rm feat}$ is a ViT matched to the encoder, and the pixel decoder
$D_{\rm pix}$ is the ViT-L decoder of BAR~\citep{yu2026bar}, shared across all encoders.
Tab.~\ref{tab:tokenizer_hparams} lists the architectures and training hyperparameters.
All tokenizers are trained on NVIDIA H200 GPUs.

\textbf{Losses.}
$\mathcal{L}_{\rm feat}$ sums $\ell_1$, $\ell_2$, and cosine distances and averages the patch
and global terms. $\mathcal{L}_{\rm reg}$ is the mean squared error between the last-layer
features of $E$ and $\hat E$.
Following BAR, $\mathcal{L}_{\rm pix}$ combines pixel, perceptual, Gram, and adversarial losses
with an alignment loss that regresses the frozen last-layer features from the third
pixel-decoder block through an MLP.
All pixel-branch terms act on stop-gradient tokens, so they update only $D_{\rm pix}$ and
the output projector $P_{{\rm out},\theta}$, which no other module uses.
The encoder, the input projector, and the feature decoder receive gradients only from
$\mathcal{L}_{\rm feat}$ and $\mathcal{L}_{\rm reg}$.
Because FSQ rounds each coordinate to fixed levels without a learned codebook, it needs no
commitment or codebook loss.
Tab.~\ref{tab:tokenizer_hparams} gives all loss weights.\looseness-1

\begin{table}[t]
\centering
\caption{\method{} tokenizer architecture and training hyperparameters at $256\times256$,
shared across encoders unless noted. We follow the tokenizer setup of BAR~\citep{yu2026bar} and change
only the parts specific to \method{}: the feature decoder and feature losses, the encoder
regularizer, the discrete global token, and the stop-gradient that restricts the pixel losses
to the pixel decoder. Epochs are over ImageNet-1k, with optimizer steps in parentheses.}
\label{tab:tokenizer_hparams}
\scriptsize
\setlength{\tabcolsep}{6pt}
\begin{tabular}{@{}l l l@{}}
\toprule
category & parameter & value \\
\midrule
\multirow{4}{*}{architecture}
 & feature decoder $D_{\rm feat}$ & ViT, $24$ layers, width $1024$ (ViT-L encoders), \\
 & & \quad $27$ layers, width $1152$ (SigLIP2) \\
 & pixel decoder $D_{\rm pix}$ & ViT-L, $24$ layers, width $1024$, patch size $16$ \\
 & quantizer & FSQ, $16$ bits per patch token and $16$ bits for the global token \\
\midrule
\multirow{9}{*}{optimization}
 & optimizer & AdamW, $\beta=(0.9, 0.999)$ \\
 & learning rate & $10^{-4}$ to $10^{-5}$, cosine decay \\
 & encoder learning rate & $0.1\times$ the learning rate \\
 & weight decay & $0$ \\
 & warmup & $2$ epochs ($10$K steps) \\
 & training length & $40$ epochs ($200$K steps) \\
 & batch size & $256$ \\
 & precision, gradient clipping & bf16, max grad norm $1.0$ \\
 & weight averaging & EMA, decay $0.999$ \\
\midrule
\multirow{2}{*}{feature losses}
 & $\mathcal{L}_{\rm feat}$ ($\ell_1$, $\ell_2$, cosine) & $1.0$, $1.0$, $1.0$ \\
 & $\lambda_{\rm reg}$ & $0.25$ (SigLIP), $0.25$ (SigLIP2), $0.5$ (DINOv2), $0.25$ (DINOv3) \\
\midrule
\multirow{6}{*}{pixel losses}
 & $\ell_2$, $\ell_1$ reconstruction & $1.0$, $0.1$ \\
 & perceptual & $1.1$ (LPIPS and ConvNeXt-S output loss, $1.0:0.1$) \\
 & Gram & $10$ \\
 & adversarial & $0.05$, discriminator on a frozen DINO ViT-S/8 backbone, from $20$ epochs \\
 & decoder feature alignment & $0.7$ \\
 & $\lambda_{\rm pix}$ & $1.0$ \\
\midrule
data & augmentation & resize to $256$, random $256\times256$ crop, random flip \\
\bottomrule
\end{tabular}
\end{table}

\textbf{Tokenizer at $512\times512$.}
For ImageNet-1k at $512\times512$ (Sec.~\ref{sec:exp_512}), we train a separate tokenizer from
DINOv3-L rather than fine-tuning the $256\times256$ one.
The encoder processes the full $512\times512$ image, giving a $32\times32$ grid of $1024$ patch
tokens.
Following BAR~\citep{yu2026bar}, each patch token and the global token carry $10$ bits, so the
quantizer has $2^{10}$ possible tokens.
The feature target, losses, and optimization follow the $256\times256$ recipe.
Tab.~\ref{tab:tokenizer_hparams_512} lists all settings and marks those that differ from
Tab.~\ref{tab:tokenizer_hparams}.

\begin{table}[t]
\centering
\caption{\method{} tokenizer architecture and training hyperparameters at $512\times512$, used
for \method{}-L-DINOv3 in Tab.~\ref{tab:res512}. $^{\ast}$Entries that differ from the
$256\times256$ recipe in Tab.~\ref{tab:tokenizer_hparams}. Epochs are over ImageNet-1k, with
optimizer steps in parentheses.}
\label{tab:tokenizer_hparams_512}
\scriptsize
\setlength{\tabcolsep}{6pt}
\begin{tabular}{@{}l l l@{}}
\toprule
category & parameter & value \\
\midrule
\multirow{5}{*}{architecture}
 & encoder input$^{\ast}$ & DINOv3-L on the full $512\times512$ image, $32\times32$ grid of $1024$ patch tokens \\
 & feature decoder $D_{\rm feat}$ & ViT, $24$ layers, width $1024$ \\
 & pixel decoder $D_{\rm pix}$ & ViT-L, $24$ layers, width $1024$, patch size $16$ \\
 & quantizer$^{\ast}$ & FSQ, $10$ bits per patch token and $10$ bits for the global token \\
 & memory$^{\ast}$ & gradient checkpointing \\
\midrule
\multirow{9}{*}{optimization}
 & optimizer & AdamW, $\beta=(0.9, 0.999)$ \\
 & learning rate & $10^{-4}$ to $10^{-5}$, cosine decay \\
 & encoder learning rate & $0.1\times$ the learning rate \\
 & weight decay & $0$ \\
 & warmup & $2$ epochs ($10$K steps) \\
 & training length & $40$ epochs ($200$K steps) \\
 & batch size & $256$, from $4$ gradient-accumulation steps$^{\ast}$ \\
 & precision, gradient clipping & bf16, max grad norm $1.0$ \\
 & weight averaging & EMA, decay $0.999$ \\
\midrule
\multirow{2}{*}{feature losses}
 & $\mathcal{L}_{\rm feat}$ ($\ell_1$, $\ell_2$, cosine) & $1.0$, $1.0$, $1.0$ \\
 & $\lambda_{\rm reg}$ & $0.25$ \\
\midrule
\multirow{6}{*}{pixel losses}
 & $\ell_2$, $\ell_1$ reconstruction & $1.0$, $0.1$ \\
 & perceptual & $1.1$ (LPIPS and ConvNeXt-S output loss, $1.0:0.1$) \\
 & Gram & $10$ \\
 & adversarial & $0.05$, frozen DINO ViT-S/8 discriminator at $448\times448$ input$^{\ast}$, from $20$ epochs \\
 & decoder feature alignment & $0.7$ \\
 & $\lambda_{\rm pix}$ & $1.0$ \\
\midrule
data & augmentation$^{\ast}$ & resize to $512$, random $512\times512$ crop, random flip \\
\bottomrule
\end{tabular}
\end{table}

\section{Generator implementation details}
\label{app:generator}

We follow BAR~\citep{yu2026bar} to implement the generative model,
with our global-augmented generative modelling.

\textbf{BAR generative model.}
Recall that BAR learns tokens with FSQ in a bit representation and models their distribution
with an adapted autoregressive model.
Instead of a single categorical prediction over the entire codebook, BAR uses a
lightweight masked bit modeling head.
During training, some bits of the target token are masked, and the head learns
to recover them with a bit-wise cross-entropy objective.
During sampling, it starts with all bits masked and progressively fills them in,
conditioning later predictions on the bits already revealed.
The autoregressive context stays fixed during these within-token steps.
Once the token is complete, it becomes part of the context for the next token.
This avoids an output distribution with one logit per codebook entry while
allowing dependencies among a token's bits.

The token-level training objective described in Sec.~\ref{sec:method_tokenizer} is
\begin{equation}
    \mathcal{L}_{\rm gen}
    =\mathbb{E}_{(z,c)}\!\left[
        \lambda_{\rm global}\,\ell_{\rm CE}(z_{\rm global}\mid c)
        +\lambda_{\rm patch}\,\frac{1}{N}\sum_{i=1}^{N}
        \ell_{\rm CE}(z_i\mid z_{\rm global},z_{<i},c)
    \right],
    \label{eq:generative_objective}
\end{equation}
where $\ell_{\rm CE}$ is the conditional cross-entropy prediction loss.

\textbf{Our implementation.}
We instantiate the conditional cross-entropy in Eq.~\ref{eq:generative_objective}
with BAR's masked bit modeling objective~\citep{yu2026bar}.
Let $B_j=(B_{j1},\ldots,B_{jd})\in\{0,1\}^{d}$ be the clean bit representation of token $z_j$,
with $d=16$, where $\ell\in\{1,\ldots,d\}$ indexes the bits.
For each token, we sample a masking ratio $r_j$ and mask $n_j=\mathrm{round}(d\,r_j)$ of its
bits, chosen uniformly at random, giving a mask $m_j\in\{0,1\}^{d}$ with $m_{j\ell}=1$ if bit $\ell$
is masked.
Following BAR, patch tokens draw $r_j=\max(1-t,\,1/d)$ with
$t=\sigma(\epsilon)$, $\epsilon\sim\mathcal N(0,0.8^2)$, and $\sigma$ the sigmoid.
The global token, which BAR does not have, draws $t\sim\mathcal U[0,1]$ instead.
The lower bound $r_j\geq 1/d$ ensures $n_j\geq 1$, so 
every token has at least one masked bit to
predict.
Replacing each masked bit $B_{j\ell}$ by a mask symbol gives $\widetilde B_j$.
The prediction head estimates the likelihood of the clean bits conditioned on
$\widetilde B_j$, the masking ratio $r_j$, and the autoregressive context $C_j$.
The mask positions are already encoded by the mask symbols in $\widetilde B_j$.
During teacher forcing, $C_{\rm global}=c$ and
$C_i=(z_{\rm global},z_{<i},c)$ for patch $i$.

Let $\mathcal S$ be a set of token positions, either the global token
$\mathcal S_{\rm global}=\{\mathrm{global}\}$ or the patch tokens
$\mathcal S_{\rm patch}=\{1,\ldots,N\}$.
For a training batch of $M$ sequences, indexed by $b$, the masked-bit loss over $\mathcal S$ is
the weighted cross-entropy over all bits of the tokens in $\mathcal S$:
\begin{equation}
    \ell_{\rm MBM}(\mathcal S)
    =-\frac{\displaystyle\sum_{b=1}^{M}\sum_{j\in\mathcal S}\sum_{\ell=1}^{d}
        w^{(b)}_{j\ell}\log p_{\phi,\ell}\!\left(B^{(b)}_{j\ell}\mid\widetilde B^{(b)}_j,r^{(b)}_j,C^{(b)}_j\right)}
        {\displaystyle\sum_{b=1}^{M}\sum_{j\in\mathcal S}\sum_{\ell=1}^{d}w^{(b)}_{j\ell}},
    \label{eq:masked_bit_loss}
\end{equation}
where $p_{\phi,\ell}$ is the head's two-class prediction at bit position $\ell$ and
$w_{j\ell}=m_{j\ell}+\eta(1-m_{j\ell})$ weights masked bits by $1$ and visible bits by
$\eta=0.1$.
The complete objective combines the global and patch losses:
\begin{equation}
    \mathcal L_{\rm gen}^{\rm MBM}
    =\lambda_{\rm global}\,\ell_{\rm MBM}(\mathcal S_{\rm global})
    +\lambda_{\rm patch}\,\ell_{\rm MBM}(\mathcal S_{\rm patch}),
    \label{eq:generative_objective_mbm}
\end{equation}
with $\lambda_{\rm patch}=1-\lambda_{\rm global}$, averaged over training batches and over the
sampled masks.
Each term $-\log p_{\phi,\ell}(B_{j\ell}\mid\cdot)$ is the cross-entropy between the observed
bit and the head's two-class prediction, so Eq.~\ref{eq:generative_objective_mbm}
specifies the token prediction loss $\ell_{\rm CE}$ in Eq.~\ref{eq:generative_objective}
for our masked bit modeling head.
Because the normalization pools the bits of all tokens in $\mathcal S$, a token with more
masked bits contributes more, so the patch loss is a bit-weighted average over patch
positions rather than exactly the uniform average $\frac{1}{N}\sum_{i=1}^{N}$ of
Eq.~\ref{eq:generative_objective}.
The bit prediction task additionally conditions on the visible bits within the
current token, making it a denoising cross-entropy objective rather than an exact
evaluation of the complete token's autoregressive negative log-likelihood.\looseness-1

\method{} retains the BAR generator backbone and masked-bit prediction mechanism while
changing the tokenizer that supplies its training codes.
For global-augmented generation, we additionally prepend a sampled global token
to the patch token sequence.
This global token is image-dependent and thus
distinct from the external class label or text condition $c$.
BAR-B and BAR-L denote the base and large generator configurations.
In our experiments for ImageNet-1k at $256\times256$, each patch token contains $16$ bits,
corresponding to $2^{16}$ possible tokens.
For ImageNet-1k at $512\times512$, we use $10$ bits per token following BAR~\citep{yu2026bar}.
We use allocation to specify the unmasking process for each token during the sampling,
\emph{e.g.}, $[4,4,4,4]$ specifies four within-token sampling steps, each
revealing four bits. These within token sampling steps
are distinct from the autoregressive steps over patch positions.
App.~\ref{app:ablation_protocol} gives the fixed ablation sampling protocol,
and App.~\ref{app:sampling} lists the sampling settings used for the main
comparisons.\looseness-1

\textbf{Architecture and training.}
Tab.~\ref{tab:generator_hparams} lists the generator architectures and training
hyperparameters, which follow BAR~\citep{yu2026bar}.
The class label enters as repeated condition tokens, and for global-augmented generation the
global token is likewise repeated after them.
For classifier-free guidance, a fixed fraction of training samples drop their conditions.
Most of these drop both the class label and the global token, and the rest drop only one of
them, so the generator also learns each condition on its own.
Following BAR, training starts with randomly permuted token orders, linearly anneals the
probability of a random order to zero, and ends with raster order.
The global-token loss weight $\lambda_{\rm global}$ of Eq.~\ref{eq:generative_objective_mbm}
depends on the encoder and, for SigLIP, on the generator size, with $\lambda_{\rm patch}=1-\lambda_{\rm global}$.
The value $1/257$ weights the global token roughly like each of the $256$ patch tokens, and the
ablation generators use $\lambda_{\rm global}=0.5$.
Generators train on tokens precomputed with the EMA weights of their tokenizers, on NVIDIA
H200 GPUs.\looseness-1

\begin{table}[t]
\centering
\caption{Generator architecture and training hyperparameters at $256\times256$. We follow the generator setup
of BAR~\citep{yu2026bar} and add only the global token: its conditioning, its joint condition
dropout, and the loss weight $\lambda_{\rm global}$. Epochs are over ImageNet-1k, with
optimizer steps in parentheses.}
\label{tab:generator_hparams}
\scriptsize
\setlength{\tabcolsep}{6pt}
\begin{tabular}{@{}l l >{\centering\arraybackslash}p{2.9cm} >{\centering\arraybackslash}p{2.9cm}@{}}
\toprule
category & parameter & BAR-B & BAR-L \\
\midrule
\multirow{4}{*}{architecture}
 & layers & $24$ & $32$ \\
 & width & $768$ & $1280$ \\
 & attention heads & $12$ & $16$ \\
 & masked bit modeling head & \multicolumn{2}{c}{$3$ layers, width $2048$} \\
\midrule
\multirow{4}{*}{conditioning}
 & class condition tokens & \multicolumn{2}{c}{$32$} \\
 & global token repeats & \multicolumn{2}{c}{$32$} \\
 & condition dropout & \multicolumn{2}{c}{$10\%$ of samples} \\
 & dropped conditions & \multicolumn{2}{c}{$60\%$ both, $20\%$ class only, $20\%$ global only} \\
\midrule
\multirow{10}{*}{optimization}
 & optimizer & \multicolumn{2}{c}{AdamW, $\beta=(0.9, 0.96)$} \\
 & learning rate & \multicolumn{2}{c}{$4\times10^{-4}$ to $10^{-5}$, cosine decay} \\
 & weight decay & \multicolumn{2}{c}{$0.03$} \\
 & warmup & \multicolumn{2}{c}{$100$ epochs ($62.5$K steps)} \\
 & random-order annealing & \multicolumn{2}{c}{$200$ to $300$ epochs ($125$K to $187.5$K steps)} \\
 & training length & \multicolumn{2}{c}{$400$ epochs ($250$K steps)} \\
 & batch size & \multicolumn{2}{c}{$2048$} \\
 & precision, gradient clipping & \multicolumn{2}{c}{bf16, max grad norm $1.0$} \\
 & dropout, attention dropout & $0.0$, $0.0$ & $0.2$, $0.2$ \\
 & weight averaging & \multicolumn{2}{c}{none} \\
\midrule
\multirow{7}{*}{loss}
 & visible-bit weight $\eta$ & \multicolumn{2}{c}{$0.1$} \\
 & patch mask ratio & \multicolumn{2}{c}{$\max(1-t,\,1/16)$, $t=\sigma(\epsilon)$, $\epsilon\sim\mathcal N(0,0.8^2)$} \\
 & global mask ratio & \multicolumn{2}{c}{$\max(1-t,\,1/16)$, $t\sim\mathcal U[0,1]$} \\
 & $\lambda_{\rm global}$, SigLIP & $0.5$ & $1/257$ \\
 & $\lambda_{\rm global}$, SigLIP2 & $1/257$ & $1/257$ \\
 & $\lambda_{\rm global}$, DINOv2 & $0.2$ & $0.2$ \\
 & $\lambda_{\rm global}$, DINOv3 & $0.2$ & $0.2$ \\
\bottomrule
\end{tabular}
\end{table}

\textbf{Generator at $512\times512$.}
At $512\times512$, we train a single BAR-L generator on the tokens of the $512\times512$
tokenizer (Tab.~\ref{tab:tokenizer_hparams_512}).
It models a sequence of one global token followed by $1024$ patch tokens of $10$ bits each.
Following the $512\times512$ recipe of BAR~\citep{yu2026bar}, it trains for $200$ epochs, half
the $256\times256$ schedule, with warmup and random-order annealing scaled accordingly.
We set $\lambda_{\rm global}=1/1025$, which weights the global token like each of the $1024$
patch tokens.
Tab.~\ref{tab:generator_hparams_512} lists all settings and marks those that differ from the
BAR-L column of Tab.~\ref{tab:generator_hparams}.\looseness-1

\begin{table}[t]
\centering
\caption{Generator architecture and training hyperparameters at $512\times512$, used for
\method{}-L-DINOv3 in Tab.~\ref{tab:res512}. $^{\ast}$Entries that differ from the BAR-L column
of the $256\times256$ recipe in Tab.~\ref{tab:generator_hparams}. Epochs are over ImageNet-1k,
with optimizer steps in parentheses.}
\label{tab:generator_hparams_512}
\scriptsize
\setlength{\tabcolsep}{6pt}
\begin{tabular}{@{}l l l@{}}
\toprule
category & parameter & BAR-L \\
\midrule
\multirow{5}{*}{architecture}
 & layers & $32$ \\
 & width & $1280$ \\
 & attention heads & $16$ \\
 & masked bit modeling head & $3$ layers, width $2048$ \\
 & sequence$^{\ast}$ & $1$ global token and $1024$ patch tokens, $10$ bits each \\
\midrule
\multirow{4}{*}{conditioning}
 & class condition tokens & $32$ \\
 & global token repeats & $32$ \\
 & condition dropout & $10\%$ of samples \\
 & dropped conditions & $60\%$ both, $20\%$ class only, $20\%$ global only \\
\midrule
\multirow{10}{*}{optimization}
 & optimizer & AdamW, $\beta=(0.9, 0.96)$ \\
 & learning rate & $4\times10^{-4}$ to $10^{-5}$, cosine decay \\
 & weight decay & $0.03$ \\
 & warmup$^{\ast}$ & $50$ epochs ($31.25$K steps) \\
 & random-order annealing$^{\ast}$ & $100$ to $150$ epochs ($62.5$K to $93.75$K steps) \\
 & training length$^{\ast}$ & $200$ epochs ($125$K steps) \\
 & batch size & $2048$ \\
 & precision, gradient clipping & bf16, max grad norm $1.0$ \\
 & dropout, attention dropout & $0.2$, $0.2$ \\
 & weight averaging & none \\
\midrule
\multirow{4}{*}{loss}
 & visible-bit weight $\eta$ & $0.1$ \\
 & patch mask ratio$^{\ast}$ & $\max(1-t,\,1/10)$, $t=\sigma(\epsilon)$, $\epsilon\sim\mathcal N(0,0.8^2)$ \\
 & global mask ratio$^{\ast}$ & $\max(1-t,\,1/10)$, $t\sim\mathcal U[0,1]$ \\
 & $\lambda_{\rm global}$$^{\ast}$ & $1/1025$ \\
\bottomrule
\end{tabular}
\end{table}

\section{Ablation details}
\label{app:ablations}

\begin{table}[t]
\centering
\caption{\textbf{Ablations for \method{}.}
Feature-only loss outperforms pixel and combined pixel--feature losses for three pretrained encoders.
Jointly modeling global and patch tokens with feature-only supervision achieves the best gFID for all four encoders.
We report rFID on the ImageNet validation set and gFID on $50$K ImageNet samples without guidance, with lower values indicating better performance.
The tokenizer target specifies which losses train the discrete tokens, and the generator type specifies whether global-augmented generative modeling is used.
Feature denotes the normalized average of four encoder layers, whereas first layer and last layer use a single encoder layer as the target.
rFID depends only on the tokenizer, so rows that share a tokenizer report the same value.
\textbf{Bold} and \underline{underline} indicate the best and second-best
result in each metric column.}
\label{tab:target_head_ablation}
\small
\setlength{\tabcolsep}{3pt}
\begin{tabular*}{\linewidth}{@{\extracolsep{\fill}}ll cc cc cc cc@{}}
\toprule
 & & \multicolumn{2}{c}{SigLIP} & \multicolumn{2}{c}{SigLIP2} & \multicolumn{2}{c}{DINOv2} & \multicolumn{2}{c}{DINOv3} \\
\cmidrule(lr){3-4}\cmidrule(lr){5-6}\cmidrule(lr){7-8}\cmidrule(l){9-10}
tokenizer target & generator type & rFID & gFID & rFID & gFID & rFID & gFID & rFID & gFID \\
\midrule
pixel & patch-only & 1.40 & 1.88 & 1.16 & 1.68 & \underline{0.85} & 1.76 & \textbf{0.72} & 2.92 \\
pixel + feature & patch-only & 1.26 & 1.74 & 1.08 & 1.74 & \textbf{0.83} & 1.72 & \underline{0.73} & 1.98 \\
feature & patch-only & \underline{1.01} & \underline{1.70} & \textbf{0.92} & 1.80 & 0.93 & 1.63 & 0.99 & 1.48 \\
pixel + feature & global-augmented & 1.26 & 1.77 & 1.08 & \underline{1.65} & \textbf{0.83} & \underline{1.60} & \underline{0.73} & 1.67 \\
first layer & global-augmented & \textbf{0.93} & 2.50 & \underline{1.03} & 1.85 & 1.26 & 1.69 & 1.03 & \underline{1.42} \\
last layer & global-augmented & 1.56 & 1.81 & 1.57 & 1.75 & 1.65 & 1.65 & 1.26 & 1.43 \\
feature (\method{}) & global-augmented & \underline{1.01} & \textbf{1.61} & \textbf{0.92} & \textbf{1.62} & 0.93 & \textbf{1.51} & 0.99 & \textbf{1.39} \\
\bottomrule
\end{tabular*}
\end{table}

\subsection{Ablation protocol and sampling without guidance}
\label{app:ablation_protocol}

The controlled ablations in Sec.~\ref{sec:ablations} compare tokenizer
design choices using BAR generators~\citep{yu2026bar}
specified in App.~\ref{app:generator} at $256\times256$ resolution.
We evaluate $50$K generated images under a fixed sampling configuration, rather
than choosing a separate best sampling configuration for each ablation.
We use guidance scale $1.0$, \emph{i.e.}, no guidance, sampling temperature $2.0$,
and bit-group allocation $[4,4,4,4]$.
For more details about the sampling hyperparameters, check App.~\ref{app:sampling}.
For global-augmented models, the global token is sampled first
with the same sampling hyperparameters, and then conditions patch generation.
Patch-only models omit the global token, but retain class conditioning.
We use the unguided protocol for ablation because it avoids the additional
unconditional prediction used for classifier-free guidance and removes guidance
strength as an additional sampling hyperparameter.
The full results presented in Sec.~\ref{sec:ablations} are detailed
in Tab.~\ref{tab:target_head_ablation}.\looseness-1

\subsection{Tokenizers from an MAE encoder}
\label{app:mae}

We also trained tokenizers from MAE-L/16~\citep{he2022mae}, whose features are learned by pixel
reconstruction and therefore fall outside Def.~\ref{def:features}.
These runs use the same tokenizer and BAR-B generator recipes, with $250$K generator steps and
$50$K samples without guidance at patch temperature $2.0$ and allocation $[4,4,4,4]$.
They differ in three respects: they omit source-preservation regularization, use the last MAE
layer as the feature target, and use MAE's auxiliary class token as the global token, sampled at
temperature $2.5$ and allocation $[2,2,5,7]$.
MAE-based tokenizers generate poorly under every supervision.
With patch-only generation, gFID is $5.74$ for the pixel target, $5.94$ for combined pixel and
feature targets, and $6.40$ for the feature target, and global-augmented generation reaches
$5.55$ and $5.75$ with the combined and feature targets.
Feature supervision thus does not improve MAE-based tokens, and every MAE configuration stays far
above the gFID range of Tab.~\ref{tab:target_head_ablation}, so we do not pursue MAE further.

\subsection{Reconstruction results}
\label{sec:exp_recon}

Tab.~\ref{tab:recon_main} reports reconstruction for four source encoders under
pixel and feature supervision, using regularized tokenizers throughout.
With pixel supervision, the pixel loss trains the encoder and quantizer.
With feature supervision, the pixel gradient is cut at the discrete representation,
and the feature decoder reconstructs the four-layer-average target.\looseness-1

Across the four encoders, rFID ranges from $0.72$ to $1.40$.
The six-space view complements rFID: feature targets improve rFDr6 over pixel
targets for SigLIP ($2.68$ to $2.45$), SigLIP2 ($2.09$ to $2.04$), and
DINOv3 ($2.24$ to $2.08$), but not DINOv2 ($1.82$ to $2.06$).
These reconstruction scores alone do not explain the generation differences
in Sec.~\ref{sec:exp_main}.\looseness-1

\textbf{Pixel fidelity versus perceptual quality.}
Feature supervision lowers PSNR and SSIM for all four encoders, \emph{e.g.}, from $20.61$ to $17.45$
PSNR for DINOv3, so the tokens no longer reproduce each input pixel by pixel.
For image generation, however, the goal is images that look real to humans, not pixel-exact
copies of a given input.
PSNR and SSIM compare each reconstruction with its input pixel by pixel, so they penalize small
shifts of texture and detail that humans barely notice, and they correlate poorly with human
perceptual judgments~\citep{lpips}.
Distribution-level metrics instead ask whether the images look like real images, and they favor
feature supervision.
Feature targets improve rFDr6 for three of the four encoders, and the same feature-supervised
tokenizers give lower gFID in the controlled ablations (Tab.~\ref{tab:target_head_ablation}),
the best gFDr6 among discrete generators (Tab.~\ref{tab:fdr6}), and higher human-preference
reward-model scores (Fig.~\ref{fig:reward}).
The lower paired pixel fidelity is therefore a trade-off rather than a free improvement.
Feature-supervised tokens discard some input detail, and not all of it need be perceptually
irrelevant, but for generation the trade-off favors feature supervision in the distribution-level
and reward-model evaluations above.\looseness-1

\begin{table}[t]
\centering
\caption{Reconstruction of regularized tokenizers on ImageNet at $256\times256$.
Feature denotes the four-layer-average target.
\textbf{Bold} marks the better result within each source block.}
\label{tab:recon_main}
\small
\setlength{\tabcolsep}{5pt}
\begin{tabular}{@{}ll cccc@{}}
\toprule
source & supervision & rFID~$\downarrow$ & rFDr6~$\downarrow$ & PSNR~$\uparrow$ & SSIM~$\uparrow$ \\
\midrule
\multirow{2}{*}{SigLIP}
 & pixel & 1.40 & 2.68 & \textbf{17.89} & \textbf{0.426} \\
 & feature & \textbf{1.01} & \textbf{2.45} & 17.27 & 0.418 \\
\midrule
\multirow{2}{*}{SigLIP2}
 & pixel & 1.16 & 2.09 & \textbf{18.81} & \textbf{0.466} \\
 & feature & \textbf{0.92} & \textbf{2.04} & 17.67 & 0.433 \\
\midrule
\multirow{2}{*}{DINOv2}
 & pixel & \textbf{0.85} & \textbf{1.82} & \textbf{20.04} & \textbf{0.519} \\
 & feature & 0.93 & 2.06 & 17.50 & 0.433 \\
\midrule
\multirow{2}{*}{DINOv3}
 & pixel & \textbf{0.72} & 2.24 & \textbf{20.61} & \textbf{0.545} \\
 & feature & 0.99 & \textbf{2.08} & 17.45 & 0.424 \\
\bottomrule
\end{tabular}
\end{table}

\subsection{Token Structure Measurement}
\label{app:token_structure}

\begin{figure}[t]
\centering
\includegraphics[width=0.45\linewidth]{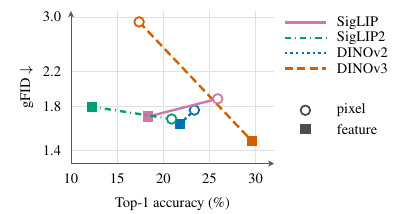}
\caption{\textbf{Linear separability and generation.}
Discrete-token probe accuracy on $25$K test images versus patch-only gFID on $50$K samples
without guidance.
The gFID axis is log-scaled.}
\label{fig:target_discrete_probe_gfid}
\end{figure}

\textbf{Spatial-dependence diagnostic.}
The spatial dependence in Fig.~\ref{fig:patchonly_target_mi_gfid} measures how informative one
patch token is about an adjacent patch token in the same image.
Rather than estimating MI between complete categorical token IDs, we use a
component-wise proxy based on the tokenizer's binary code representation.
Each patch token is represented by $16$ bits.
For each pair of patch positions at a given offset, we estimate the MI, in bits, between pairs
of these bits from their joint occurrence frequencies across $25$K class-balanced ImageNet
training images.
We average these estimates over the $256$ bit pairs, over all position pairs of the offset, and
over the four cardinal directions.
Higher values indicate stronger statistical dependence between the bits of spatially separated
tokens.
The same estimator is applied to all eight tokenizers.
Each point in
Fig.~\ref{fig:patchonly_target_mi_gfid} pairs the estimate for adjacent patches with the gFID of
the patch-only generator from Tab.~\ref{tab:target_head_ablation}.
Replacing the pixel target by the feature target raises the dependence for
SigLIP, DINOv2, and DINOv3, but it lowers the dependence for SigLIP2.
This diagnostic measures component-wise dependence in the tokenizer's code
parameterization, not MI between full token IDs or the prediction loss of a learned
generator.\looseness-1

\textbf{Discrete-token linear probes. \ \ }
We fit linear classifiers on mean-pooled patch bits using $100$K training
images, select the learning rate and weight decay on $10$K validation images,
and evaluate on $25$K test images held out from probe fitting,
averaging accuracy over three training seeds.
As in Fig.~\ref{fig:patchonly_target_mi_gfid},
Fig.~\ref{fig:target_discrete_probe_gfid} pairs each
tokenizer with the gFID of its patch-only generator from Tab.~\ref{tab:target_head_ablation}.
Replacing the pixel target by the feature target lowers probe accuracy but improves gFID for
SigLIP and DINOv2.
For SigLIP2, probe accuracy and gFID both worsen, and for DINOv3, both improve.
Thus, this measure of linear separability does not consistently track class-conditional
generation quality, unlike the spatial dependence in
Fig.~\ref{fig:patchonly_target_mi_gfid}.\looseness-1

\section{Evaluation protocol}
\label{app:eval}

This section collects the experimental setup and measurement conventions used
in Sec.~\ref{sec:experiments}.\looseness-1

\textbf{Experimental setup.}
We evaluate class-conditioned image generation on ImageNet-1k~\citep{imagenet} at $256\times256$
and $512\times512$, and text-conditioned generation on GenEval~\citep{geneval}, GenAI-Bench~\citep{genaibench},
and DPG-Bench~\citep{ella}.

\textbf{Reconstruction and generation metrics. \ \ }
For a fixed feature extractor, Fr\'echet distance compares the empirical means and
covariances of real and reconstructed or generated image features.
rFID is the tokenizer's reconstruction measurement that compares the reconstructed
images and original images in the Inception space using the ImageNet-1k validation
set.
gFID compares generated images with the original training images using
Fr\'echet distance in Inception space.
We re-use the released reference statistics from~\citet{jit}.
We also report gFDr6 introduced by \citet{fdloss2026},
which reports generative Fr\'echet distance
in six different feature spaces.
The six-space evaluation uses Inception-v3~\citep{inceptionv3},
ConvNeXt~V2~\citep{convnextv2}, DINOv2~\citep{dinov2}, MAE~\citep{he2022mae},
SigLIP2~\citep{siglip2}, and CLIP~\citep{clip}, with the checkpoints of \citet{fdloss2026}:
ConvNeXt~V2-B, DINOv2 ViT-L/14, MAE ViT-L/16, SigLIP2 So400m/16, and CLIP ViT-L/14.
In each space, the Fr\'echet-distance ratio divides the measured distance by that
space's fixed reference normalization constant, \emph{i.e.}, the distance computed
between the training and validation sets. gFDr6 averages these six ratios
for generated images, and rFDr6 averages them for reconstructions.
We re-use the released reference statistics from \citet{fdloss2026}.
All generation distribution metrics use $50$K generated samples,
and PRDC compares them with $10$K real reference images.

\textbf{Measured here versus transcribed.}
For the ImageNet comparisons in Tabs.~\ref{tab:fdr6},
\ref{tab:fdr6_continuous}, \ref{tab:fdr6_shortened_raev2}, and~\ref{tab:res512},
all per-representation gFDr
and gFDr6 values are evaluated by us using 50k samples, except the guided continuous-baseline
results in Tab.~\ref{tab:fdr6_continuous}, transcribed from
Tab.~7 of \citet{Singh2026raev2}, and the Drift results and guided iMF and pMF results
in Tab.~\ref{tab:fdr6_shortened_raev2}, transcribed from \citet{fdloss2026} for the
original models without FD-loss post-training.
We evaluate the original iMF and pMF checkpoints without guidance ourselves, at guidance
scale $1$ with their released guidance intervals.
Drift's released models already sample without guidance, so its transcribed results fill
both column groups of Tab.~\ref{tab:fdr6_shortened_raev2}.
Fig.~\ref{fig:teaser} uses the same unguided one-step values.
For text-to-image generation, the continuous-baseline results in
Tab.~\ref{tab:t2i} are taken from \citet{Singh2026raev2}, whereas the
discrete-baseline (BAR-L) results use our own implementations and evaluations.
All \method{} results, reward scores in Fig.~\ref{fig:reward}, and appendix
ablations are evaluated by us.
For its final models, BAR~\citep{yu2026bar} fine-tunes the pixel decoder for $40$ additional
epochs.
Our primary BAR comparison excludes this decoder-only post-training for both methods, which
separates the effect of the proposed tokenization and generation recipe from an additional
decoder-refinement stage.
BAR's original and refined decoders consume the same discrete tokens.
The original decoder is used in Tabs.~\ref{tab:fdr6} and~\ref{tab:mainall_full_k3},
Figs.~\ref{fig:teaser} and~\ref{fig:reward}, and the qualitative samples in
App.~\ref{app:qualitative}.
Tab.~\ref{tab:fdr6_gfidsel} additionally reports BAR with the refined decoder at the same
sampling settings.
Baselines use their released sampling settings when available, and otherwise settings selected by
a sweep (App.~\ref{app:sampling}), and App.~\ref{app:gfid_prdc} additionally
reports gFDr6 for \method{} at gFID-selected configurations (Tab.~\ref{tab:fdr6_gfidsel}).\looseness-1

\begin{table}[t]
\centering
\caption{Sampling configurations of \method{} for the $256\times256$ results in Tab.~\ref{tab:fdr6}.}
\label{tab:sampparams}
\scriptsize
\setlength{\tabcolsep}{4pt}
\begin{tabular}{@{}ll ccc ccc@{}}
\toprule
& & \multicolumn{3}{c}{patch branch} & \multicolumn{3}{c}{global branch} \\
\cmidrule(lr){3-5}\cmidrule(lr){6-8}
model & guidance & alloc. & gs & temp & alloc. & gs & temp \\
\midrule
\multicolumn{8}{l}{\emph{base generator ($0.4$B)}} \\
\multirow{2}{*}{\textbf{\method{}-SigLIP}}
 & without & $[4,4,4,4]$ & 1.0 & 1.8 & $[4,4,4,4]$ & 1.0 & 1.4 \\
 & with & $[4,4,4,4]$ & 4.0 & 1.6 & $[4,4,4,4]$ & 1.0 & 2.0 \\
\midrule
\multirow{2}{*}{\textbf{\method{}-SigLIP2}}
 & without & $[4,4,4,4]$ & 1.0 & 1.6  & $[3,3,5,5]$ & 1.0 & 2.0 \\
 & with & $[4,4,4,4]$ & 4.0 & 1.6  & $[4,4,4,4]$ & 1.0 & 2.0 \\
\midrule
\multirow{2}{*}{\textbf{\method{}-DINOv2}}
 & without & $[4,4,4,4]$ & 1.0 & 2.0  & $[4,4,4,4]$ & 1.0 & 2.0 \\
 & with & $[4,4,4,4]$ & 3.0 & 2.0  & $[4,4,4,4]$ & 1.5 & 2.0 \\
\midrule
\multirow{2}{*}{\textbf{\method{}-DINOv3}}
 & without & $[4,4,4,4]$ & 1.0 & 1.75 & $[3,3,5,5]$ & 1.0 & 2.0 \\
 & with & $[4,4,4,4]$ & 3.0 & 1.6  & $[4,4,4,4]$ & 1.0 & 2.0 \\
\midrule
\multicolumn{8}{l}{\emph{large generator ($1.1$B)}} \\
\multirow{2}{*}{\textbf{\method{}-SigLIP}}
 & without & $[4,4,4,4]$ & 1.0 & 1.8 & $[2,2,5,7]$ & 1.0 & 2.0 \\
 & with & $[4,4,4,4]$ & 2.0 & 1.8 & $[4,4,4,4]$ & 1.5 & 2.4 \\
\midrule
\multirow{2}{*}{\textbf{\method{}-SigLIP2}}
 & without & $[4,4,4,4]$ & 1.0 & 1.75 & $[4,4,4,4]$ & 1.0 & 2.0 \\
 & with & $[4,4,4,4]$ & 2.5 & 1.6  & $[4,4,4,4]$ & 1.0 & 2.0 \\
\midrule
\multirow{2}{*}{\textbf{\method{}-DINOv2}}
 & without & $[4,4,4,4]$ & 1.0 & 2.0  & $[4,4,4,4]$ & 1.0 & 2.0 \\
 & with & $[2,2,5,7]$ & 3.0 & 2.5  & $[4,4,4,4]$ & 1.5 & 2.5 \\
\midrule
\multirow{2}{*}{\textbf{\method{}-DINOv3}}
 & without & $[3,3,5,5]$ & 1.0 & 2.0  & $[3,3,5,5]$ & 1.0 & 2.0 \\
 & with & $[4,4,4,4]$ & 2.0 & 1.9  & $[4,4,4,4]$ & 1.5 & 2.0 \\
\bottomrule
\end{tabular}
\end{table}

\section{Sampling hyperparameters}
\label{app:sampling}

We report the sampling hyperparameters for all the main results we presented in the paper.

\textbf{The knobs.}
\method{} samples two branches, and they are configured independently: the $256$ patch positions
and the single global position each carry their own guidance scale, temperature and bit-group
allocation. We write an allocation as its list of group sizes, so $[2,2,5,7]$ unmasks a
token's $16$ bits in four groups of $2$, $2$, $5$ and $7$ bits, and $[b^{k}]$ abbreviates
$k$ groups of $b$ bits. The list sums to the bits per token ($16$ at $256^2$, $10$ at
$512^2$) and its \emph{length} sets the sampling cost, since a $k$-group schedule issues
$k{+}1$ forwards per position. A guidance scale of $1.0$ is
no-guidance, so every row without guidance carries patch guidance $1.0$.
The global branch is unguided in every configuration without guidance and in half of the guided ones.

\textbf{Reduced-step RAEv2. \ \ }
The released RAEv2 sampling configuration for ImageNet $256\times256$
(\url{https://github.com/nanovisionx/RAEv2/blob/8a0d238/configs/stage2/sampling/imagenet-dinov3l-k7.yaml})
uses $100$ ODE steps and an internal-guidance scale of $1.78$, which we use for all other RAEv2
results.
For the $3$- and $6$-step RAEv2 rows of Tab.~\ref{tab:fdr6_shortened_raev2}, we select
the internal-guidance scale with the lowest gFDr6 among the evaluated settings, rather
than keeping the released scale of $1.78$.
The selected scales are $4.0$ and $2.6$, respectively.\looseness-1

\textbf{Sampling hyperparameters for resolution $512\times512$. \ \ }
At $512$ a token carries $10$ bits, so the allocations sum to $10$ rather than $16$.
Without guidance, \method{}-L-DINOv3 in Tab.~\ref{tab:res512} samples the patch branch
with allocation $[2,2,3,3]$, scale $1.0$, and temperature $2.4$, and the global branch
with allocation $[2,2,2,4]$, scale $1.0$, and temperature $2.0$.
With guidance, the patch branch uses allocation $[2,2,2,4]$, scale $3.5$, and
temperature $2.6$, and the global branch uses allocation $[2,2,2,4]$, scale $1.0$,
and temperature $2.5$.
Guided BAR-L~\citep{yu2026bar} and RAE~\citep{Zheng2025rae} use the sampling settings
released with those models.
BAR does not release an unguided $512$ configuration, so we did a hyperparameter sweep and
set its unguided sampling configuration as
allocation $[2,2,3,3]$, scale $1.0$, and temperature $2.4$.

\begin{table}[t]
\centering
\caption{ImageNet-1k generation at $256\times256$, without and with guidance,
evaluated by gFID and PRDC.
\textbf{Bold} and \underline{underline} mark the highest and second-highest
precision, recall, density, and coverage within each discrete generator-scale block,
with ties at the displayed precision sharing a rank.}
\label{tab:mainall_full_k3}
\scriptsize
\setlength{\tabcolsep}{2.5pt}
\begin{tabular}{@{}l ccccc ccccc@{}}
\toprule
& \multicolumn{5}{c}{generation@256 w/o guidance} & \multicolumn{5}{c}{generation@256 w/ guidance} \\
\cmidrule(lr){2-6}\cmidrule(lr){7-11}
method & gFID~$\downarrow$ & Prec.~$\uparrow$ & Rec.~$\uparrow$ & Dens.~$\uparrow$ & Cov.~$\uparrow$ & gFID~$\downarrow$ & Prec.~$\uparrow$ & Rec.~$\uparrow$ & Dens.~$\uparrow$ & Cov.~$\uparrow$ \\
\midrule
\multicolumn{11}{l}{\emph{base generator}} \\
BAR-B~\citep{yu2026bar} & 1.64 & 0.800 & 0.614 & \textbf{1.153} & \textbf{0.997} & 1.13 & 0.775 & \textbf{0.660} & 1.060 & 0.994 \\
\textbf{\method{}-B-SigLIP} & 1.61 & \textbf{0.807} & 0.621 & \underline{1.146} & \underline{0.996} & 1.27 & 0.778 & 0.654 & 1.068 & 0.994 \\
\textbf{\method{}-B-SigLIP2} & 1.53 & \underline{0.802} & 0.621 & 1.142 & 0.995 & 1.12 & 0.782 & 0.655 & 1.067 & \textbf{0.996} \\
\textbf{\method{}-B-DINOv2} & 1.47 & 0.782 & \underline{0.645} & 1.085 & \underline{0.996} & 1.10 & \underline{0.784} & \underline{0.659} & \underline{1.069} & \underline{0.995} \\
\textbf{\method{}-B-DINOv3} & 1.34 & 0.783 & \textbf{0.647} & 1.082 & \underline{0.996} & 1.07 & \textbf{0.789} & 0.652 & \textbf{1.088} & \textbf{0.996} \\
\midrule
\multicolumn{11}{l}{\emph{large generator}} \\
BAR-L~\citep{yu2026bar} & 1.36 & 0.798 & 0.631 & 1.121 & \underline{0.995} & 1.02 & 0.780 & \textbf{0.667} & 1.059 & \textbf{0.996} \\
\textbf{\method{}-L-SigLIP} & 1.37 & \underline{0.800} & 0.633 & \textbf{1.134} & \underline{0.995} & 1.26 & \textbf{0.801} & 0.636 & \textbf{1.128} & \textbf{0.996} \\
\textbf{\method{}-L-SigLIP2} & 1.33 & \textbf{0.801} & 0.631 & \underline{1.127} & \textbf{0.996} & 1.07 & \underline{0.785} & 0.656 & \underline{1.070} & \underline{0.994} \\
\textbf{\method{}-L-DINOv2} & 1.26 & 0.787 & \underline{0.643} & 1.091 & \textbf{0.996} & 1.15 & 0.784 & 0.652 & \underline{1.070} & \underline{0.994} \\
\textbf{\method{}-L-DINOv3} & 1.15 & 0.791 & \textbf{0.651} & 1.092 & \underline{0.995} & 1.05 & 0.781 & \underline{0.665} & 1.053 & \textbf{0.996} \\
\midrule
\multicolumn{11}{l}{\emph{continuous latents}} \\
RAEv2~\citep{Singh2026raev2} & 1.61 & 0.790 & 0.634 & 1.116 & 0.995 & 1.04 & 0.792 & 0.652 & 1.099 & 0.996 \\
\bottomrule
\end{tabular}
\end{table}

\section{Comparison using conventional gFID and PRDC metrics.}
\label{app:gfid_prdc}
In the main paper, we report results in gFDr6~\citep{fdloss2026}.
We compare our method with baselines using gFID~\citep{FID}
and Precision, Recall~\citep{improvedprecisionrecall}, Density,
and Coverage~\citep{densitycoverage} in Tab.~\ref{tab:mainall_full_k3}.
Per conventions, we use different hyperparameters
during sampling for gFID evaluation than gFDr6.

gFID measures the Fr\'echet distance in the feature space of an Inception-v3
network~\citep{inceptionv3} trained for ImageNet classification.
Its sensitivity to ImageNet class information means that it can be lowered substantially without
improving the images~\citep{kynkaanniemi2022role}, and Inception features represent the content
of modern generators poorly, so gFID often disagrees with human raters~\citep{dinov2FD,jayasumana2024cmmd}.
These problems are most severe at the low values reached by current ImageNet generators.
ImageNet validation images score a gFID of $1.68$ against the training set~\citep{fdloss2026}, so
generators below $2.0$, and especially below $1.68$, appear as close to the training data as real
images, although their samples remain clearly distinguishable from real ones~\citep{fdloss2026}.
In this regime, gFID can also be optimized directly: post-training a generator on Inception features
lowers its gFID from $3.31$ to $0.81$ without producing the best samples, whereas post-training on
modern representations improves object structure while gFID becomes worse~\citep{fdloss2026}.
gFDr6 addresses these issues by averaging Fr\'echet distances over six representation spaces,
each normalized by the distance between the training and validation sets, so no single feature
space dominates and a value of $1$ means as close to the training data as real validation images.\looseness-1

Tab.~\ref{tab:mainall_full_k3} provides additional evidence.
At these low-gFID operating points, improvements in Inception Fr\'echet distance
do not consistently agree with the fidelity diagnostics.
Across the ten BAR and \method{} models, the guided configuration has lower gFID
and higher recall in every case, but lower precision in seven and lower density
in nine.
For example, BAR-B's gFID improves from $1.64$ to $1.13$, while precision falls
from $0.800$ to $0.775$ and density from $1.153$ to $1.060$. Recall rises from
$0.614$ to $0.660$.
Thus, rather than demonstrating a uniform fidelity gain, these selected
configurations often show a fidelity--recall trade-off in the opposite direction
to the usual motivation for guidance.
Coverage remains near saturation ($0.994$--$0.997$).
The results caution against interpreting a small gFID improvement as an
unambiguous improvement in generation quality, motivating evaluation across
multiple representations and complementary reward measures.\looseness-1

Without guidance, every \method{}-B model has a lower gFID than BAR-B ($1.34$--$1.61$ against
$1.64$), three of the four \method{}-L models improve on BAR-L ($1.15$--$1.33$ against $1.36$, with
\method{}-L-SigLIP at $1.37$), and \method{}-L-DINOv3 reaches $1.15$ against $1.61$ for RAEv2.
With guidance, three of the four \method{}-B models improve over BAR-B, while BAR-L reaches the
lowest large-model gFID, $1.02$, against $1.04$ for RAEv2 and $1.05$ for \method{}-L-DINOv3.
These guided values are all well below the $1.68$ of real validation images, in the regime where
gFID no longer tracks sample quality.
Under gFDr6, which accounts for all six representation spaces, \method{}-L-DINOv3 improves over
BAR-L by $25\%$ with guidance (Tab.~\ref{tab:fdr6}).\looseness-1

The baselines in Tab.~\ref{tab:fdr6} use their released sampling settings, which were selected
for gFID, whereas the \method{} settings there are selected for gFDr6.
Tab.~\ref{tab:fdr6_gfidsel} therefore also reports gFDr6 for every \method{} model at its
gFID-selected configuration, the one used in Tab.~\ref{tab:mainall_full_k3}, so that both sides
follow the same selection criterion.
Under this criterion, all four \method{}-B models and three of the four \method{}-L models still
improve on the original-decoder BAR model of the same size, with and without guidance.
\method{}-L-DINOv3 reduces gFDr6 from $4.57$ to $3.35$ without guidance and from $4.11$ to $3.45$
with guidance, while \method{}-L-SigLIP is the exception, at $4.75$ and $4.42$.
\method{} does not use the additional $40$ epochs of decoder-only fine-tuning, yet its DINOv3
tokenizers still improve on BAR with the refined decoder in all four settings, \emph{e.g.},
\method{}-L-DINOv3 reduces gFDr6 from $4.10$ to $3.35$ without guidance and from $3.67$ to $3.45$
with guidance, and its DINOv2 tokenizers do so in three of the four settings.\looseness-1

\begin{table}[t]
\centering
\caption{gFDr6 on ImageNet-1k $256\times256$ at gFID-selected sampling configurations.
BAR uses its released settings, and each \method{} model uses the configuration of
Tab.~\ref{tab:mainall_full_k3}.
Refined-decoder rows replace BAR's pixel decoder by the one fine-tuned for $40$ additional
epochs, with the same tokens and sampling settings.
Columns follow Tab.~\ref{tab:fdr6}.
\textbf{Bold} and \underline{underline} mark the best and second-best values within each
generator size in each metric column.}
\label{tab:fdr6_gfidsel}
\fdrtablefont
\setlength{\tabcolsep}{0.5pt}
\fdrsetwidth
\begin{tabular}{@{}l G FFFFFF G FFFFFF@{}}
\toprule
& \multicolumn{7}{c}{without guidance} & \multicolumn{7}{c}{with guidance} \\
\cmidrule(lr){2-8}\cmidrule(l){9-15}
method & \fdrheadg{gFDr6} & \fdrhead{incep.} & \fdrhead{cvnx2} & \fdrhead{dino2} & \fdrhead{mae} & \fdrhead{siglip2} & \fdrhead{clip} & \fdrheadg{gFDr6} & \fdrhead{incep.} & \fdrhead{cvnx2} & \fdrhead{dino2} & \fdrhead{mae} & \fdrhead{siglip2} & \fdrhead{clip} \\
\midrule
BAR-B~\citep{yu2026bar} & 5.15 & 1.01 & 1.70 & 5.21 & 5.82 & 9.40 & 7.73 & 4.89 & 0.69 & 0.95 & 4.63 & 5.89 & 9.32 & 7.86 \\
BAR-B (refined decoder) & 4.62 & 0.99 & 1.65 & 4.71 & 5.23 & 8.39 & 6.76 & 4.36 & \textbf{0.67} & 0.90 & 4.08 & 5.35 & 8.32 & 6.81 \\
\textbf{\method{}-B-SigLIP} & 5.08 & 1.01 & 1.31 & 5.63 & 6.11 & 9.28 & 7.15 & 4.63 & 0.79 & 1.03 & 4.77 & 5.77 & 8.91 & 6.51 \\
\textbf{\method{}-B-SigLIP2} & 4.66 & 0.96 & \underline{1.29} & 4.97 & 5.14 & 8.85 & \underline{6.73} & 4.03 & 0.70 & \underline{0.85} & 3.88 & 4.71 & 7.65 & \underline{6.39} \\
\textbf{\method{}-B-DINOv2} & \underline{4.40} & \underline{0.92} & 1.36 & \textbf{4.12} & \underline{4.93} & \underline{8.31} & 6.74 & \underline{3.87} & 0.69 & \underline{0.85} & \textbf{3.37} & \textbf{4.51} & \underline{7.34} & 6.43 \\
\textbf{\method{}-B-DINOv3} & \textbf{4.21} & \textbf{0.87} & \textbf{1.24} & \underline{4.26} & \textbf{4.87} & \textbf{7.73} & \textbf{6.30} & \textbf{3.80} & \underline{0.68} & \textbf{0.75} & \underline{3.66} & \underline{4.61} & \textbf{7.01} & \textbf{6.07} \\
\midrule
BAR-L~\citep{yu2026bar} & 4.57 & 0.83 & 1.38 & 4.59 & 5.22 & 8.14 & 7.26 & 4.11 & \underline{0.61} & 0.79 & 3.72 & 5.02 & 7.49 & 7.06 \\
BAR-L (refined decoder) & 4.10 & 0.81 & 1.34 & 4.11 & 4.72 & 7.34 & 6.26 & \underline{3.67} & \textbf{0.60} & \underline{0.76} & \underline{3.29} & 4.56 & \underline{6.77} & \underline{6.00} \\
\textbf{\method{}-L-SigLIP} & 4.75 & 0.85 & 1.09 & 5.23 & 5.59 & 8.69 & 7.08 & 4.42 & 0.78 & 0.92 & 4.73 & 5.14 & 8.19 & 6.74 \\
\textbf{\method{}-L-SigLIP2} & 4.02 & 0.84 & 1.13 & 4.30 & 4.33 & 7.47 & 6.05 & 3.91 & 0.66 & 0.89 & 3.83 & \underline{4.17} & 7.52 & 6.41 \\
\textbf{\method{}-L-DINOv2} & \underline{3.57} & \underline{0.78} & \underline{0.88} & \textbf{3.25} & \underline{4.21} & \underline{6.58} & \underline{5.72} & 3.91 & 0.71 & 0.86 & \textbf{3.17} & 4.43 & 7.38 & 6.90 \\
\textbf{\method{}-L-DINOv3} & \textbf{3.35} & \textbf{0.74} & \textbf{0.75} & \underline{3.36} & \textbf{4.03} & \textbf{5.98} & \textbf{5.25} & \textbf{3.45} & 0.65 & \textbf{0.68} & 3.39 & \textbf{4.00} & \textbf{6.34} & \textbf{5.63} \\
\bottomrule
\end{tabular}
\end{table}

\section{Sampling-cost measurement}
\label{app:cost}

\textbf{TFLOPs.}
TFLOPs is arithmetic per image, counted with \texttt{torch.utils.flop\_counter} in the
$2\times$MAC convention, and is therefore hardware-independent unlike wall-clock
throughput.
It covers the full sampling loop of the generator plus one decoder pass that maps the
generated tokens or latents to pixels.
Classifier-free guidance duplicates the generator batch and roughly doubles its cost
(BAR, \method{}, SiT, LightningDiT, DDT and the discrete baselines), whereas REPA-E and
REG gate the doubled forward to a guidance interval and pay $1.8\times$, RAEv2's internal
guidance reuses a single forward and costs nothing, and RAE's autoguidance adds a smaller
second network for $+30\%$.
Guidance never duplicates the decoder, since only the conditional sample is decoded.
Each decoder is built from its official release and driven through the released
sampler's decode call.
Decoder costs range from $0.16$ TFLOPs for the pixel decoder shared by BAR and \method{}
to $0.62$ for the SD-VAE decoder of the latent diffusion baselines.
They are small relative to multi-step continuous generators, but not negligible for the cheapest
samplers: the pixel decoder adds $15$--$32\%$ to the generator cost of BAR and \method{}, and the
SD-VAE decoder exceeds the generator cost of the latent one-step models.\looseness-1

\textbf{Throughput and memory. \ \ }
Tab.~\ref{tab:throughput} measures sampling on one NVIDIA L40S GPU in bf16 without
compilation.
Each run times the generation of the discrete tokens or continuous latents and the
subsequent pixel decoding separately, so we report generation-only and end-to-end
throughput from the same run.
\emph{Throughput} is the number of images produced per second, computed as batch size
divided by the generation time, or by the generation and decoding time for end-to-end
throughput.
\emph{Memory} is peak GPU memory usage during sampling and decoding at the stated batch
size, reported in GiB, not memory per image.\looseness-1

For peak throughput, we compare each model and guidance setting at its own best
end-to-end batch size, selected from $\{1,8,32,128,256,512,1024,2048\}$.
Larger batches benefit autoregressive models, whose end-to-end peaks occur at batches
$256$--$1024$.
RAEv2 instead peaks at batch $8$ with $1.89$ images/s and slows to about $1.1$ images/s
at larger batches on this GPU.
\method{}-B peaks at $31.82$ images/s without guidance and $16.31$ with guidance.
\method{}-L reaches $15.35$ and $7.75$ images/s, which is $8.1\times$ and $4.2\times$ the
corresponding peak of RAEv2.
Decoding takes $1.3$--$4.3\%$ of end-to-end time at batch $256$ for BAR and \method{} and
$0.4\%$ for RAEv2, and it never raises peak memory.
We report the corresponding batch size and memory alongside each peak.\looseness-1

\begin{table}[t]
\centering
\caption{Sampling throughput and peak GPU memory on one L40S (bf16, uncompiled).
Throughput is in generated images per second (img/s, higher is better), and memory is the peak
GPU memory in GiB (lower is better).
\emph{gen.} counts generation only, and \emph{e2e} includes pixel decoding.
We report a matched batch size of $256$, and each model/setting's best end-to-end throughput over
the batch sweep, with the batch size (bsz) at which it is reached and its memory.
Guided rows use classifier-free guidance for BAR and \method{} (scale $5.0$ for base and $5.3$
for large models) and internal guidance ($1.78$) for RAEv2, which needs no extra forward pass.
\method{} rows use the DINOv3 tokenizer, and throughput does not depend on the encoder.
RAEv2 uses $100$ ODE steps, the setting of its released sampling configuration
(\url{https://github.com/nanovisionx/RAEv2/blob/8a0d238/configs/stage2/sampling/imagenet-dinov3l-k7.yaml}).}
\label{tab:throughput}
\small
\setlength{\tabcolsep}{4.5pt}
\begin{tabular}{@{}ll ccc ccc@{}}
\toprule
& & \multicolumn{3}{c}{matched batch $256$} & \multicolumn{3}{c}{peak e2e over the batch sweep} \\
\cmidrule(lr){3-5}\cmidrule(lr){6-8}
system & guidance & gen.\ img/s~$\uparrow$ & e2e img/s~$\uparrow$ & GiB~$\downarrow$ & img/s~$\uparrow$ & bsz & GiB~$\downarrow$ \\
\midrule
\multirow{2}{*}{BAR-B~\citep{yu2026bar}}
 & without & 28.93 & 27.69 & 10.6 & 34.32 & 1024 & 32.2 \\
 & with & 17.06 & 16.59 & 17.8 & 17.49 & 512 & 32.2 \\
\multirow{2}{*}{\textbf{\method{}}-B}
 & without & 27.49 & 26.36 & 10.4 & 31.82 & 1024 & 31.3 \\
 & with & 15.97 & 15.57 & 17.4 & 16.31 & 512 & 31.4 \\
\midrule
\multirow{2}{*}{BAR-L~\citep{yu2026bar}}
 & without & 16.02 & 15.61 & 22.1 & 17.02 & 512 & 36.8 \\
 & with & 8.71 & 8.59 & 36.8 & 8.59 & 256 & 36.8 \\
\multirow{2}{*}{\textbf{\method{}}-L}
 & without & 14.69 & 14.36 & 22.0 & 15.35 & 512 & 36.7 \\
 & with & 7.85 & 7.75 & 36.7 & 7.75 & 256 & 36.7 \\
\midrule
\multirow{2}{*}{RAEv2 \citep{Singh2026raev2}}
 & without & 1.10 & 1.09 & 13.4 & 1.89 & 8 & 7.5 \\
 & with & 1.09 & 1.09 & 14.2 & 1.85 & 8 & 7.5 \\
\bottomrule
\end{tabular}
\end{table}

\section{Qualitative samples across tokenizer encoders}
\label{app:qualitative}

\begin{figure}[p]
\centering

    \includegraphics[width=0.78\linewidth,height=0.25\textheight,keepaspectratio]{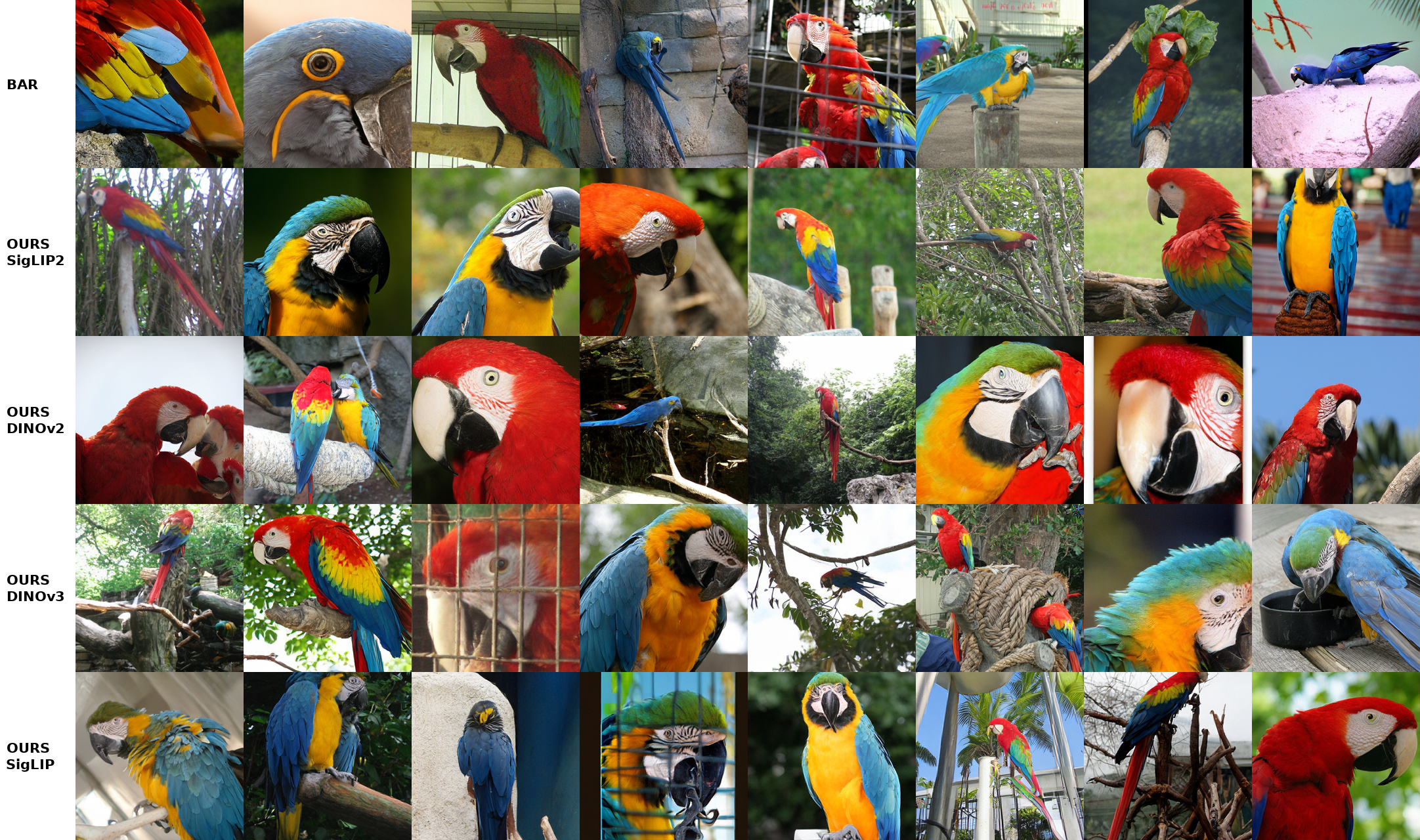}%
  \\[1pt]
  {\small macaw (88)}\\[5pt]

    \includegraphics[width=0.78\linewidth,height=0.25\textheight,keepaspectratio]{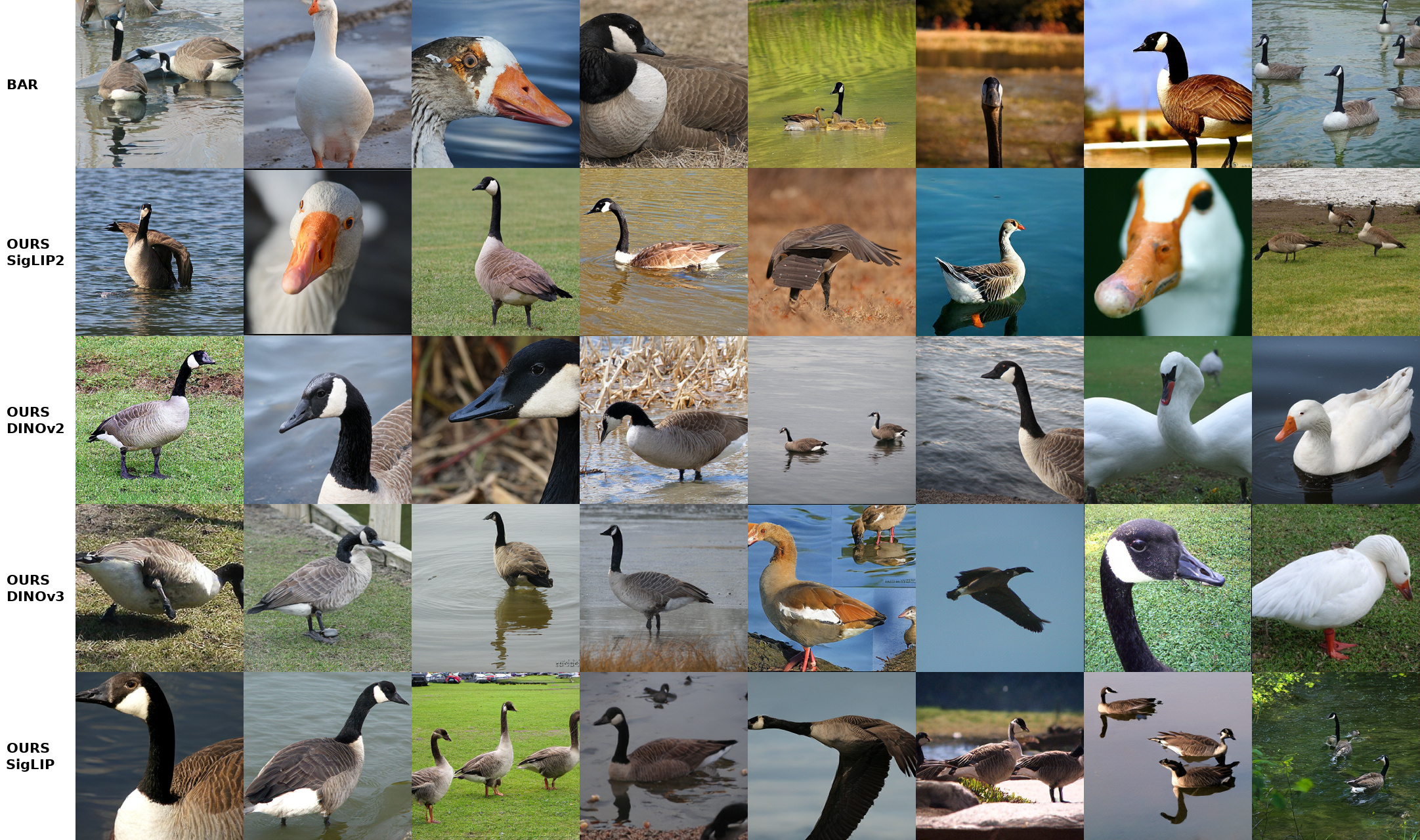}%
  \\[1pt]
  {\small goose (99)}\\[5pt]

    \includegraphics[width=0.78\linewidth,height=0.25\textheight,keepaspectratio]{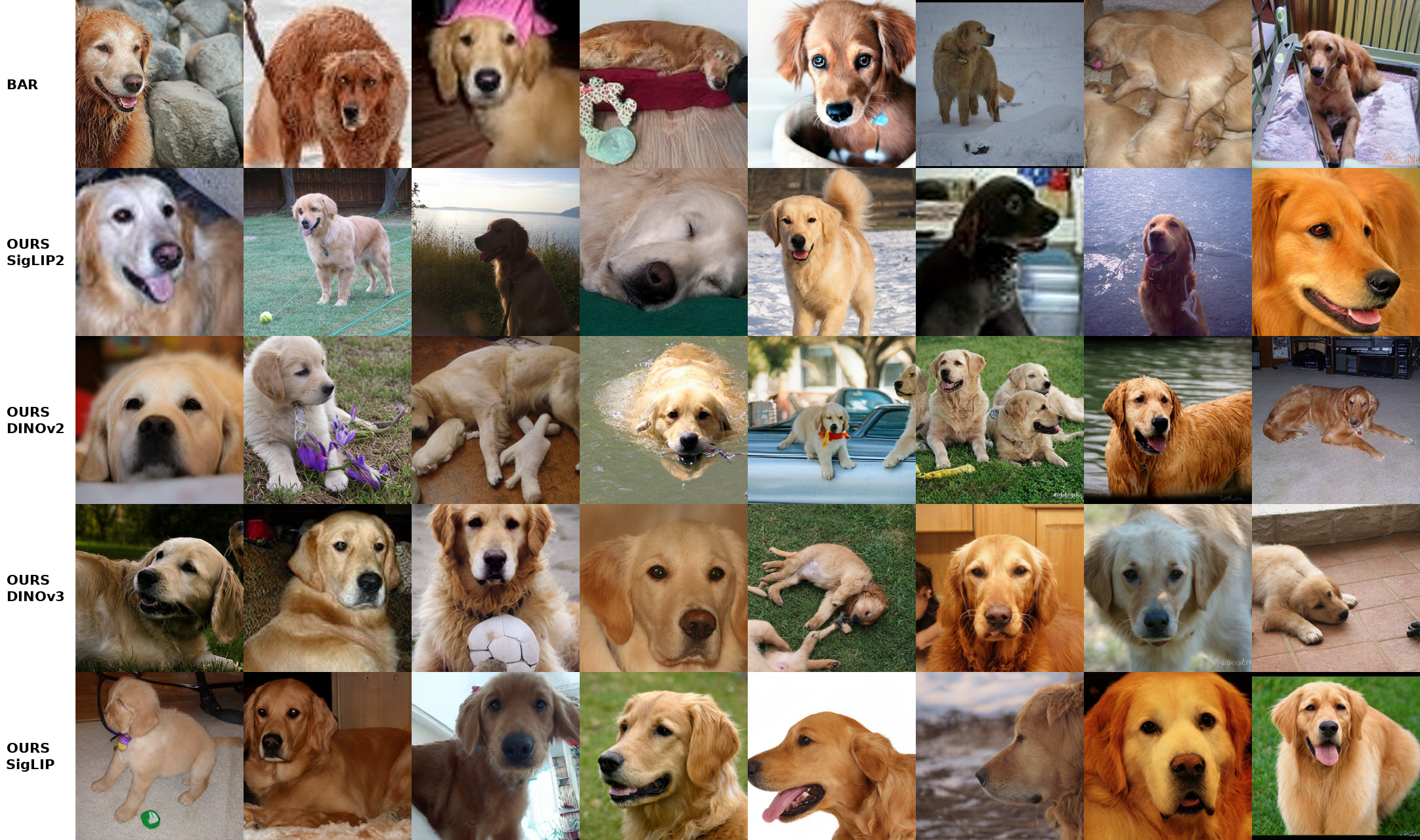}%
  \\[1pt]
  {\small golden retriever (207)}\\[5pt]
\caption{ImageNet qualitative samples (1/4). Rows: BAR-L, then \method{}-L
with SigLIP2, DINOv2, DINOv3, and SigLIP, at their respective guided settings
(App.~\ref{app:sampling}).}
\label{fig:qual-a}
\end{figure}

\begin{figure}[p]
\centering

    \includegraphics[width=0.78\linewidth,height=0.25\textheight,keepaspectratio]{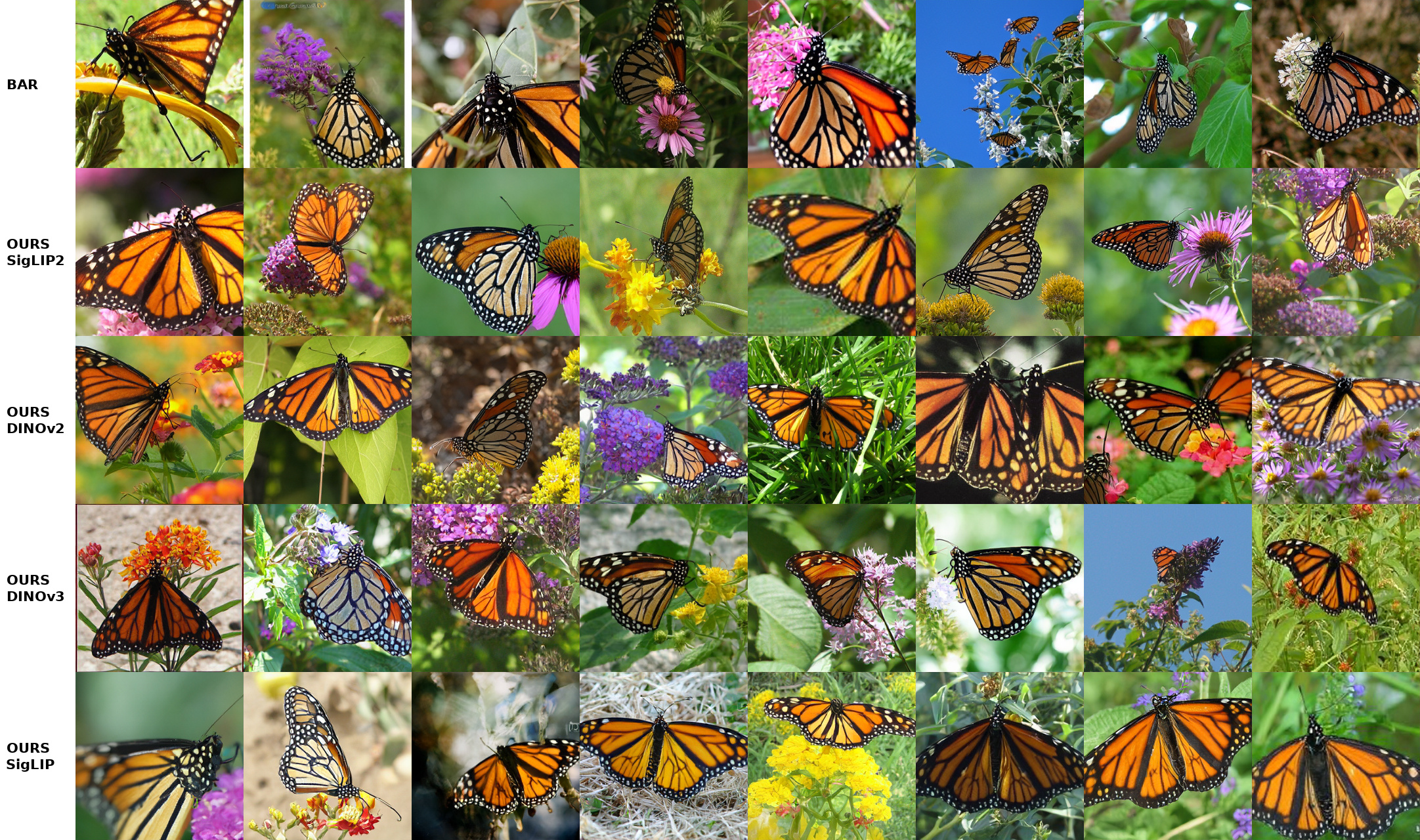}%
  \\[1pt]
  {\small monarch butterfly (323)}\\[5pt]

    \includegraphics[width=0.78\linewidth,height=0.25\textheight,keepaspectratio]{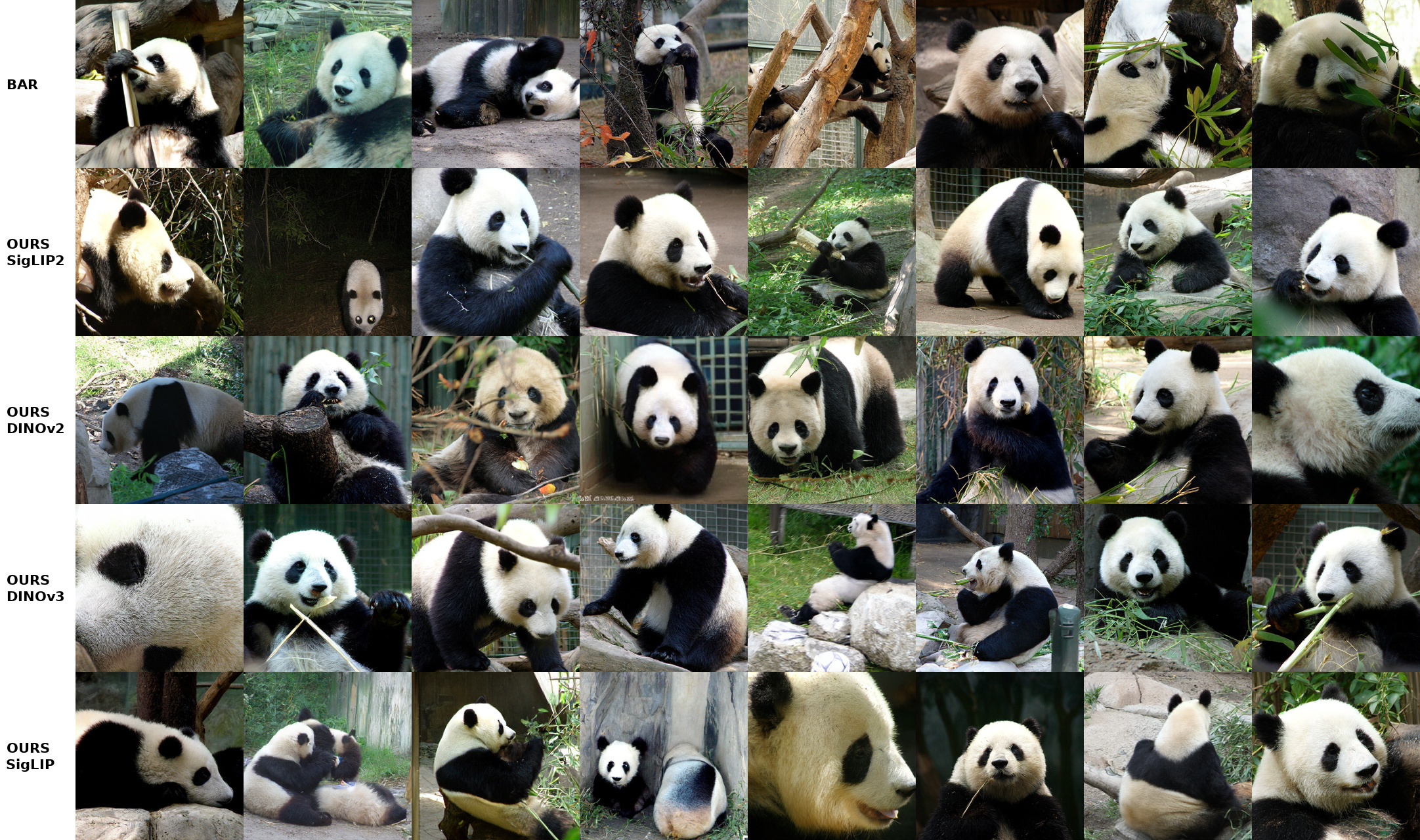}%
  \\[1pt]
  {\small giant panda (388)}\\[5pt]

    \includegraphics[width=0.78\linewidth,height=0.25\textheight,keepaspectratio]{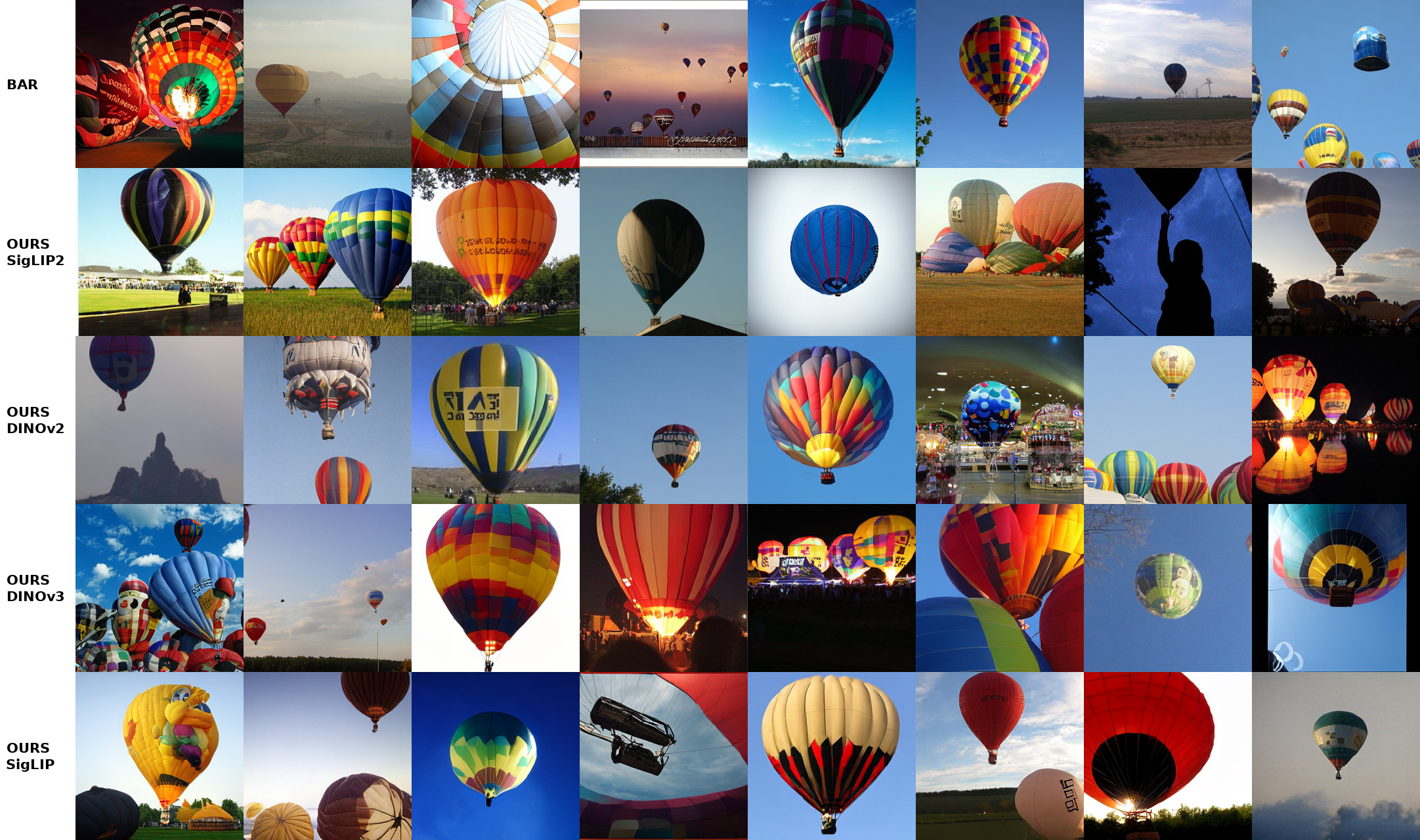}%
  \\[1pt]
  {\small balloon (417)}\\[5pt]
\caption{ImageNet qualitative samples (2/4). Rows: BAR-L, then \method{}-L
with SigLIP2, DINOv2, DINOv3, and SigLIP, at their respective guided settings
(App.~\ref{app:sampling}).}
\label{fig:qual-b}
\end{figure}

\begin{figure}[p]
\centering

    \includegraphics[width=0.78\linewidth,height=0.25\textheight,keepaspectratio]{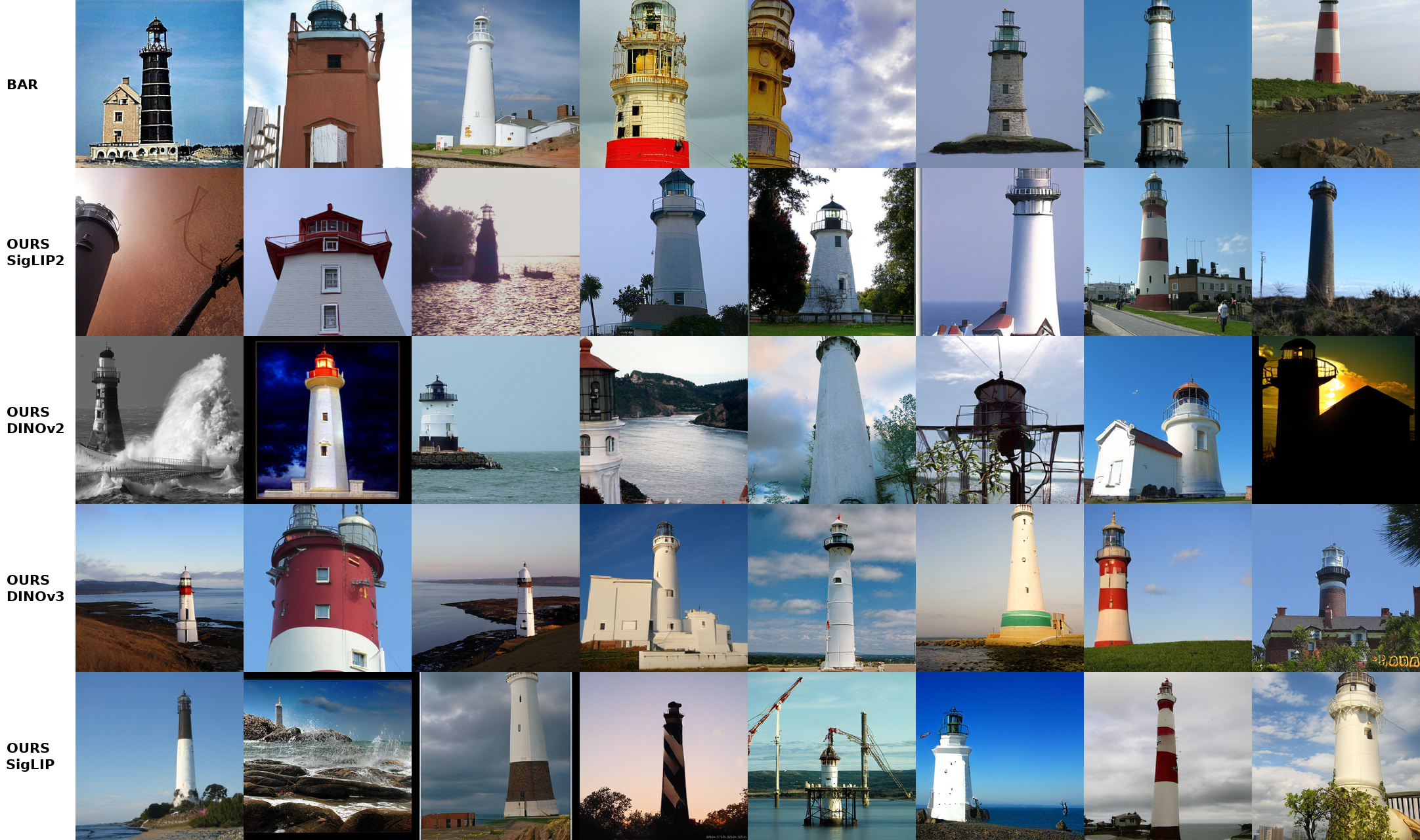}%
  \\[1pt]
  {\small lighthouse (437)}\\[5pt]

    \includegraphics[width=0.78\linewidth,height=0.25\textheight,keepaspectratio]{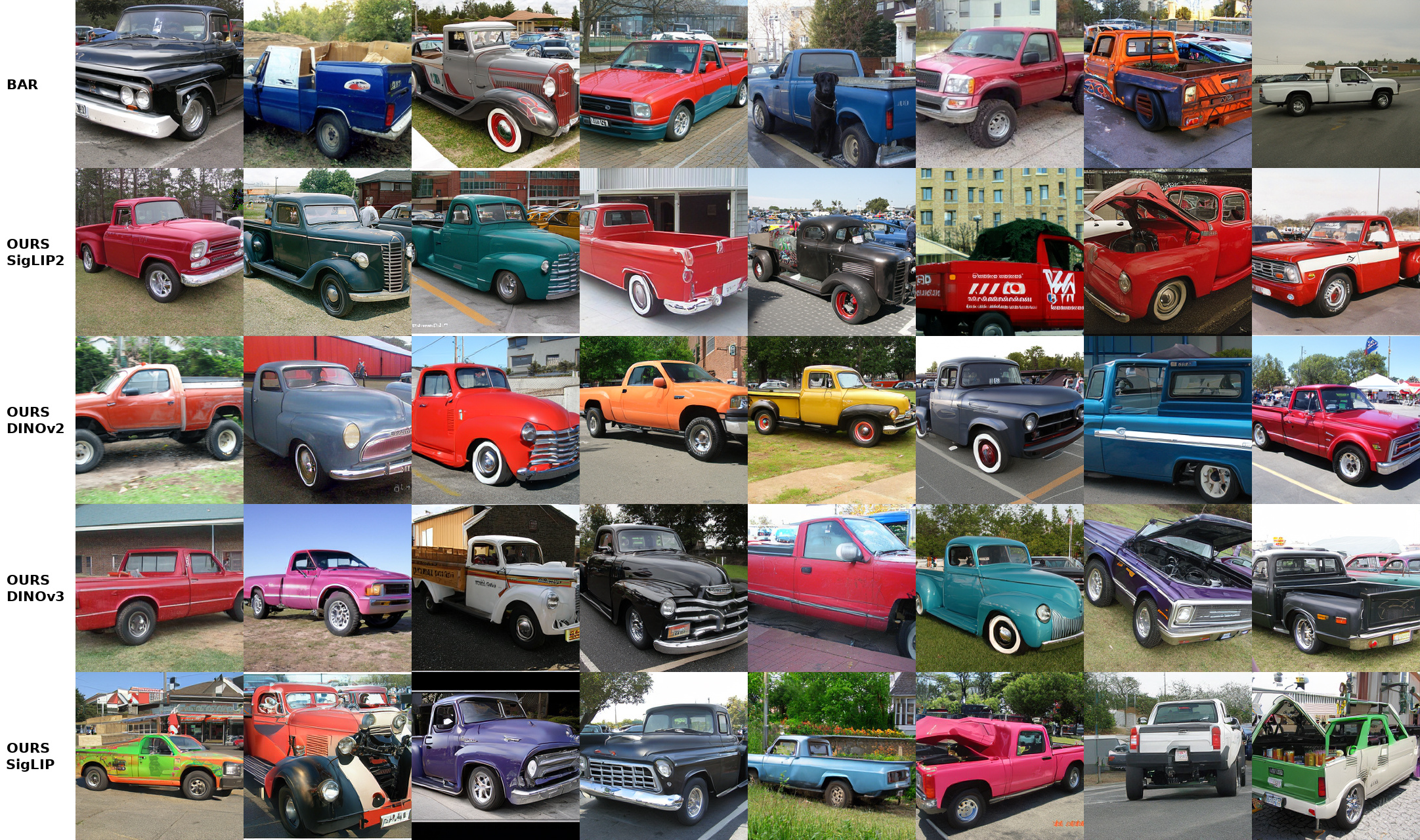}%
  \\[1pt]
  {\small pickup truck (717)}\\[5pt]

    \includegraphics[width=0.78\linewidth,height=0.25\textheight,keepaspectratio]{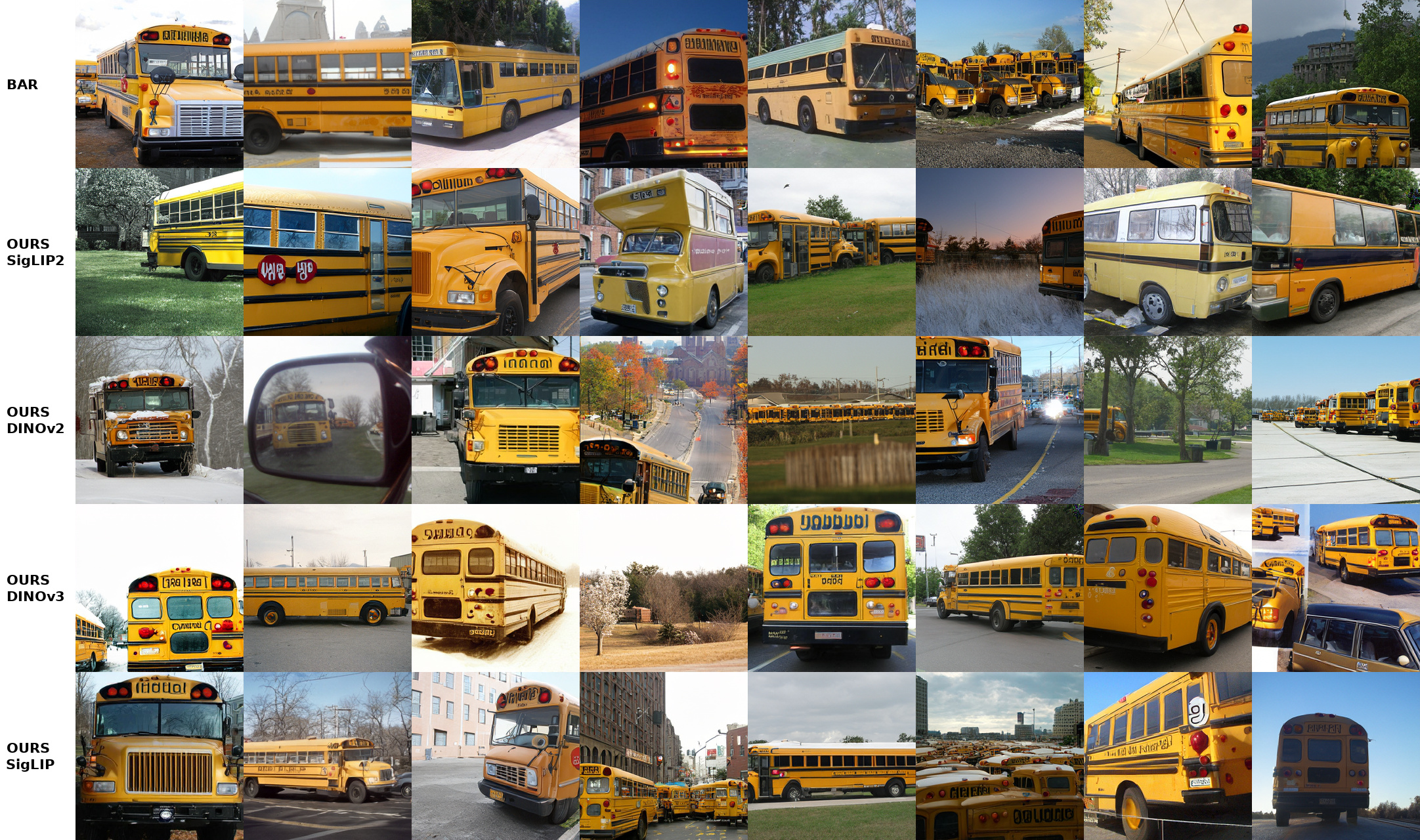}%
  \\[1pt]
  {\small school bus (779)}\\[5pt]
\caption{ImageNet qualitative samples (3/4). Rows: BAR-L, then \method{}-L
with SigLIP2, DINOv2, DINOv3, and SigLIP, at their respective guided settings
(App.~\ref{app:sampling}).}
\label{fig:qual-c}
\end{figure}

\begin{figure}[p]
\centering

    \includegraphics[width=0.78\linewidth,height=0.25\textheight,keepaspectratio]{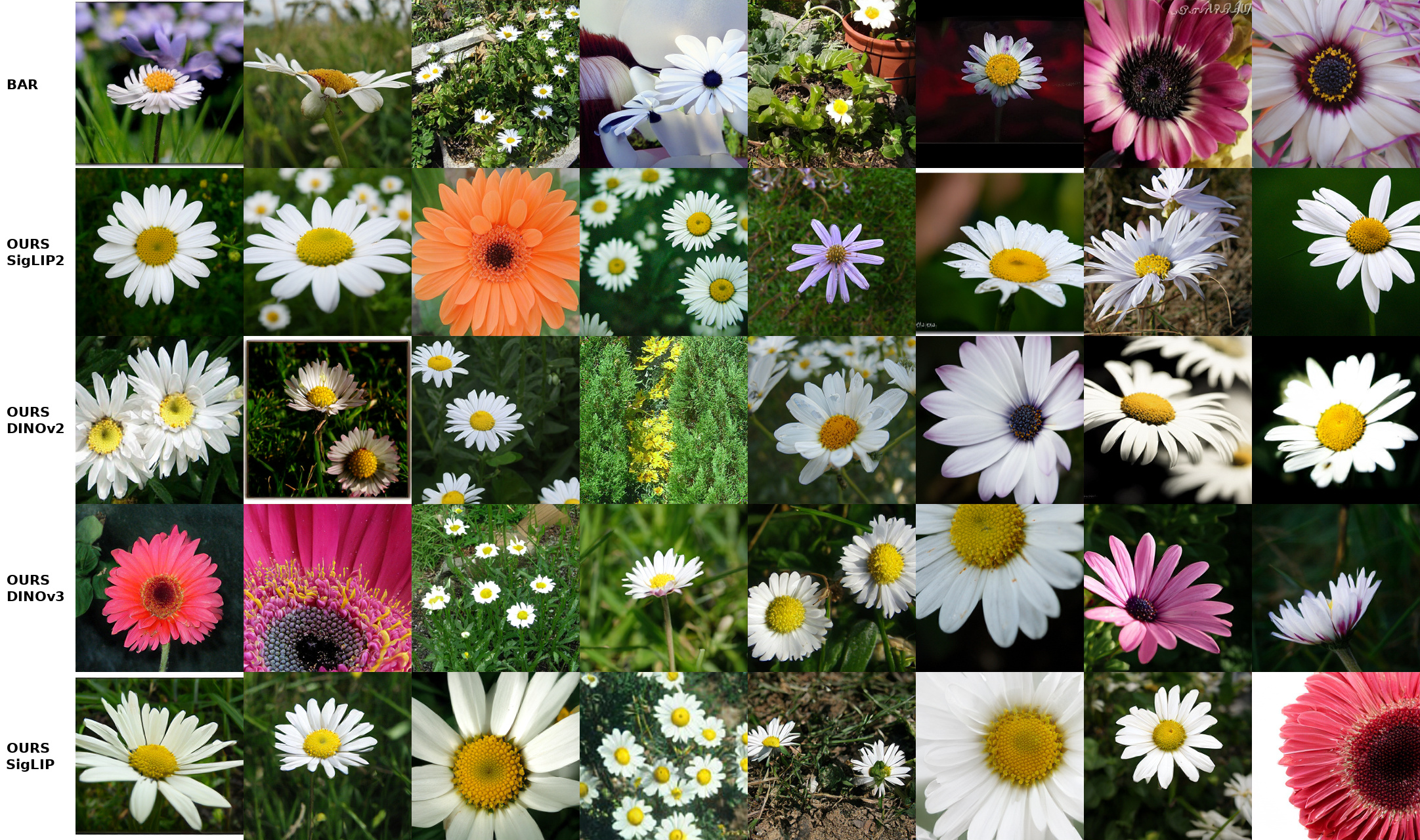}%
  \\[1pt]
  {\small daisy (985)}\\[5pt]

    \includegraphics[width=0.78\linewidth,height=0.25\textheight,keepaspectratio]{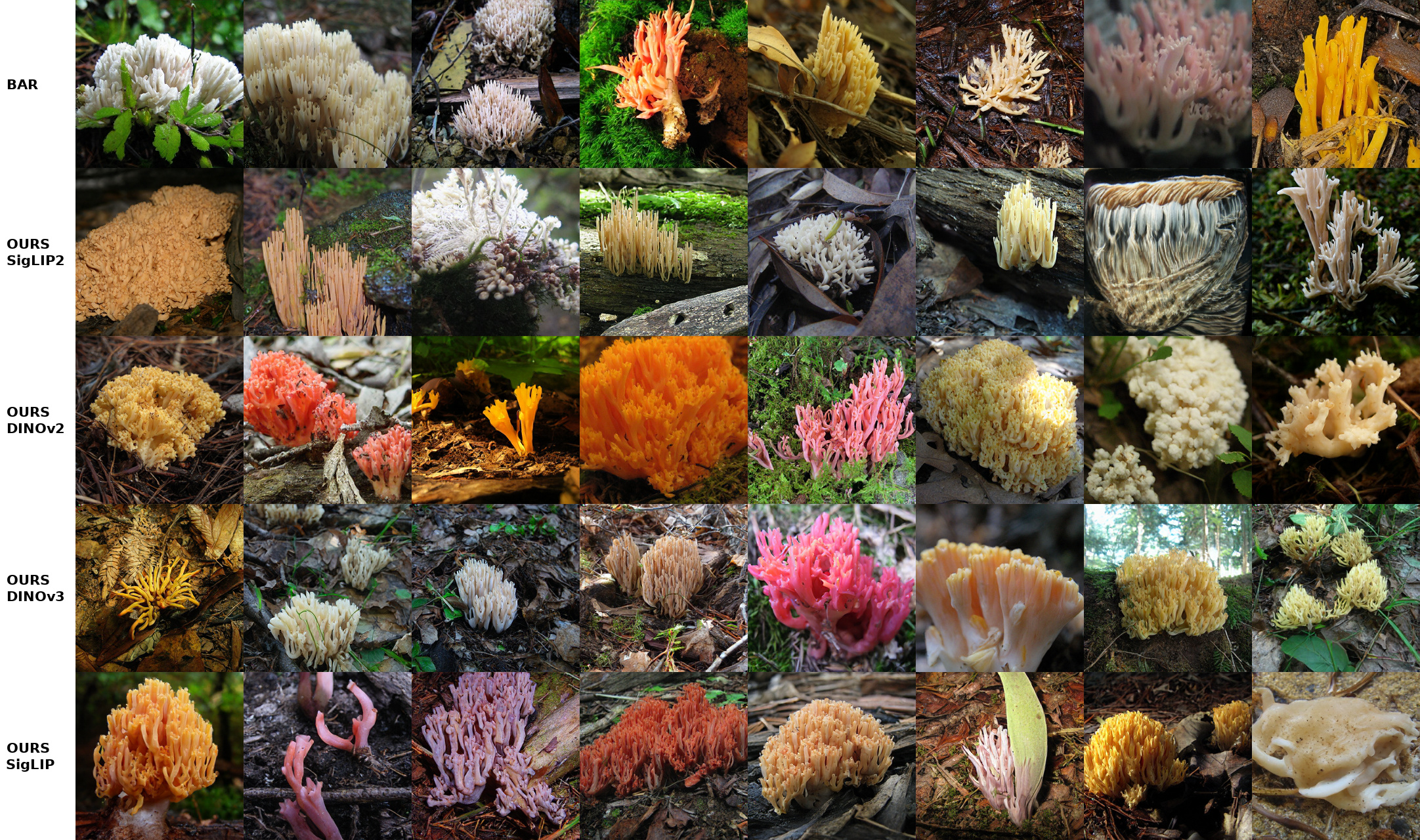}%
  \\[1pt]
  {\small coral fungus (991)}\\[5pt]

    \includegraphics[width=0.78\linewidth,height=0.25\textheight,keepaspectratio]{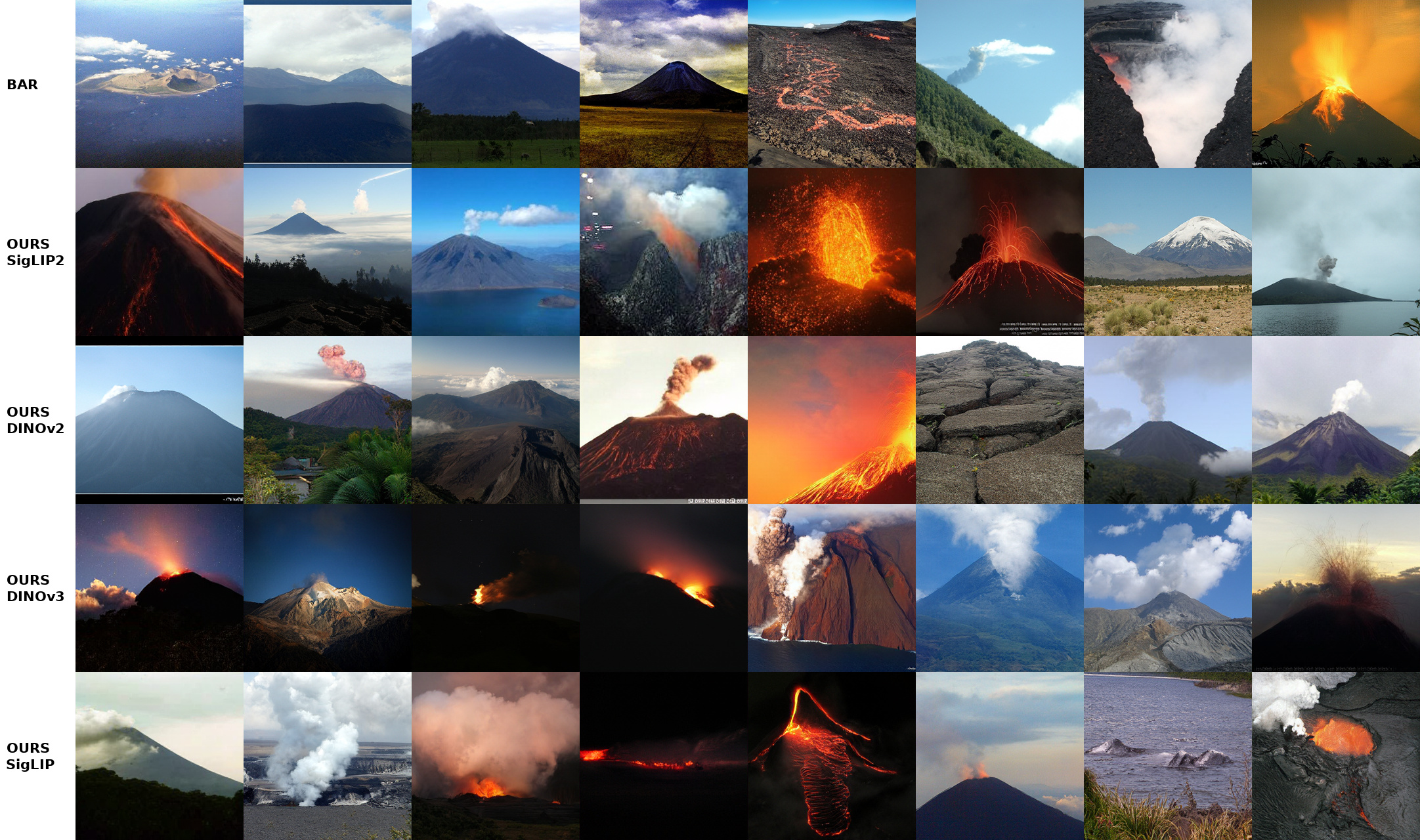}%
  \\[1pt]
  {\small volcano (980)}\\[5pt]
\caption{ImageNet qualitative samples (4/4). Rows: BAR-L, then \method{}-L
with SigLIP2, DINOv2, DINOv3, and SigLIP, at their respective guided settings
(App.~\ref{app:sampling}).}
\label{fig:qual-d}
\end{figure}

Figs.~\ref{fig:qual-a}--\ref{fig:qual-d} compare the released BAR-L model against our
\method{} models with multiple encoders. 
For each block, the five rows
are labelled on the left in the order BAR-L, then \method{}-L with SigLIP2, DINOv2, DINOv3, and SigLIP,
top to bottom.
The eight images in each row are drawn with a fixed random seed
from that class's $50$ generated samples.\looseness-1

\section{Qualitative text-to-image samples}
\label{sec:qual-t2i}

\begin{figure}[h]
\centering
  \begin{minipage}[t]{0.19\linewidth}%
    \centering
    \includegraphics[width=\linewidth]{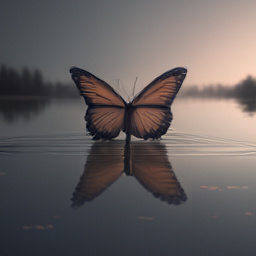}\\[1pt]
    {\tiny\stepcounter{mjhqpanel}(\themjhqpanel)~a butterfly gently touching the vast surface of the lake, peace, serenity,\par}
  \end{minipage}\hfill%
  \begin{minipage}[t]{0.19\linewidth}%
    \centering
    \includegraphics[width=\linewidth]{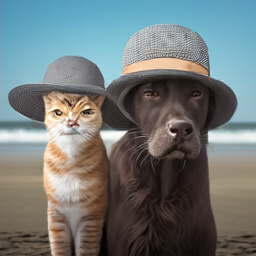}\\[1pt]
    {\tiny\stepcounter{mjhqpanel}(\themjhqpanel)~a photo of a cat and a dog on the beach wearing fancy hats\par}
  \end{minipage}\hfill%
  \begin{minipage}[t]{0.19\linewidth}%
    \centering
    \includegraphics[width=\linewidth]{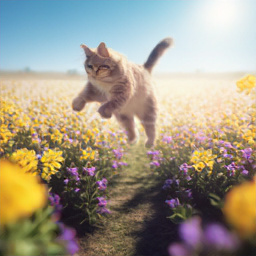}\\[1pt]
    {\tiny\stepcounter{mjhqpanel}(\themjhqpanel)~abstract photo of a cat jumping through a flower field on a bright sunny day\par}
  \end{minipage}\hfill%
  \begin{minipage}[t]{0.19\linewidth}%
    \centering
    \includegraphics[width=\linewidth]{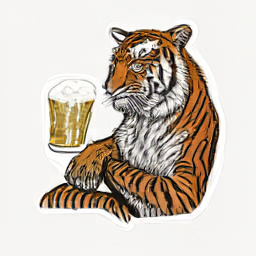}\\[1pt]
    {\tiny\stepcounter{mjhqpanel}(\themjhqpanel)~american traditional tiger drinking a beer, sticker, square\par}
  \end{minipage}\hfill%
  \begin{minipage}[t]{0.19\linewidth}%
    \centering
    \includegraphics[width=\linewidth]{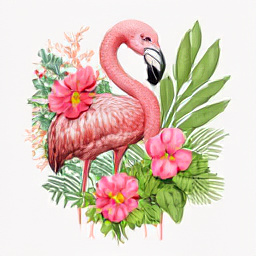}\\[1pt]
    {\tiny\stepcounter{mjhqpanel}(\themjhqpanel)~cute beautiful flamingo, floral garnish, tropical flowers, watercolor\ldots\par}
  \end{minipage}\hfill%
  \\[4pt]
  \begin{minipage}[t]{0.19\linewidth}%
    \centering
    \includegraphics[width=\linewidth]{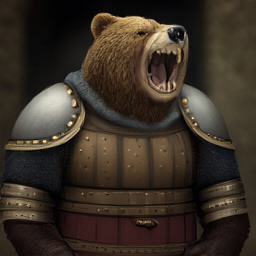}\\[1pt]
    {\tiny\stepcounter{mjhqpanel}(\themjhqpanel)~grizzly bear wearing medieval armor looking fierce and roaring\par}
  \end{minipage}\hfill%
  \begin{minipage}[t]{0.19\linewidth}%
    \centering
    \includegraphics[width=\linewidth]{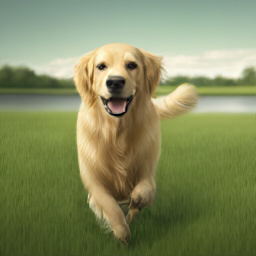}\\[1pt]
    {\tiny\stepcounter{mjhqpanel}(\themjhqpanel)~photorealistic image of a golden retriever happily running through a green\ldots\par}
  \end{minipage}\hfill%
  \begin{minipage}[t]{0.19\linewidth}%
    \centering
    \includegraphics[width=\linewidth]{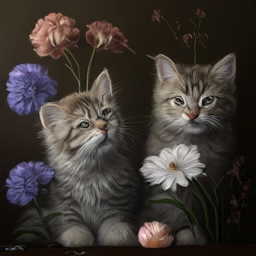}\\[1pt]
    {\tiny\stepcounter{mjhqpanel}(\themjhqpanel)~Realistic painting of flowers and kittens\par}
  \end{minipage}\hfill%
  \begin{minipage}[t]{0.19\linewidth}%
    \centering
    \includegraphics[width=\linewidth]{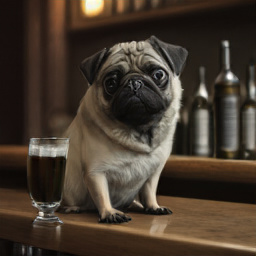}\\[1pt]
    {\tiny\stepcounter{mjhqpanel}(\themjhqpanel)~realistic photo of a pug having drinks at the bar\par}
  \end{minipage}\hfill%
  \begin{minipage}[t]{0.19\linewidth}%
    \centering
    \includegraphics[width=\linewidth]{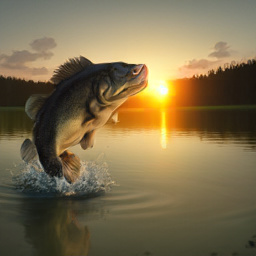}\\[1pt]
    {\tiny\stepcounter{mjhqpanel}(\themjhqpanel)~realistic photo, sunset over a lake, bass jumping out of lake, 8k\par}
  \end{minipage}\hfill%
  \\[4pt]
  \begin{minipage}[t]{0.19\linewidth}%
    \centering
    \includegraphics[width=\linewidth]{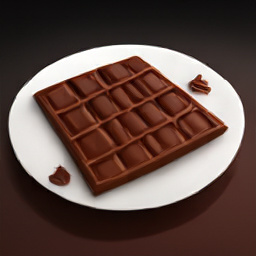}\\[1pt]
    {\tiny\stepcounter{mjhqpanel}(\themjhqpanel)~a chocolate bar on a plate, 3d, very clean\par}
  \end{minipage}\hfill%
  \begin{minipage}[t]{0.19\linewidth}%
    \centering
    \includegraphics[width=\linewidth]{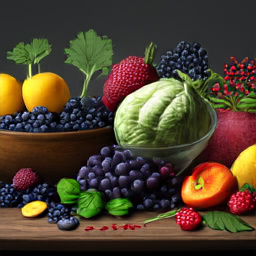}\\[1pt]
    {\tiny\stepcounter{mjhqpanel}(\themjhqpanel)~A colorful still life with a variety of vegetables, fruits and berries\ldots\par}
  \end{minipage}\hfill%
  \begin{minipage}[t]{0.19\linewidth}%
    \centering
    \includegraphics[width=\linewidth]{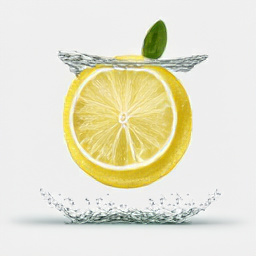}\\[1pt]
    {\tiny\stepcounter{mjhqpanel}(\themjhqpanel)~a hyperrealistic image of half a lemon dropping into clear water, but on a\ldots\par}
  \end{minipage}\hfill%
  \begin{minipage}[t]{0.19\linewidth}%
    \centering
    \includegraphics[width=\linewidth]{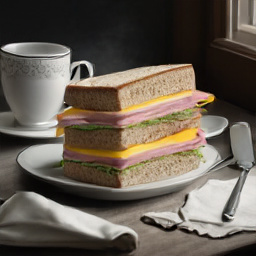}\\[1pt]
    {\tiny\stepcounter{mjhqpanel}(\themjhqpanel)~a photo of a breakfast sandwich,\par}
  \end{minipage}\hfill%
  \begin{minipage}[t]{0.19\linewidth}%
    \centering
    \includegraphics[width=\linewidth]{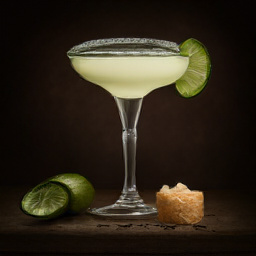}\\[1pt]
    {\tiny\stepcounter{mjhqpanel}(\themjhqpanel)~a realistic photo of a margarita cocktail\par}
  \end{minipage}\hfill%
  \\[4pt]
  \begin{minipage}[t]{0.19\linewidth}%
    \centering
    \includegraphics[width=\linewidth]{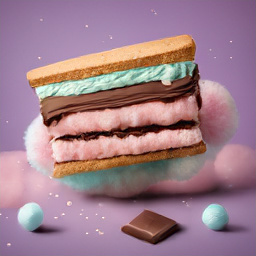}\\[1pt]
    {\tiny\stepcounter{mjhqpanel}(\themjhqpanel)~a sandwich made with cotton candy and chocolate\par}
  \end{minipage}\hfill%
  \begin{minipage}[t]{0.19\linewidth}%
    \centering
    \includegraphics[width=\linewidth]{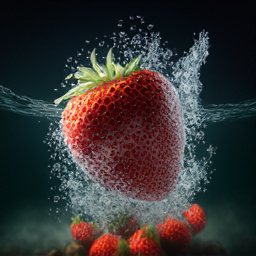}\\[1pt]
    {\tiny\stepcounter{mjhqpanel}(\themjhqpanel)~close up photograph of underwater strawberry with water splash, fine art\ldots\par}
  \end{minipage}\hfill%
  \begin{minipage}[t]{0.19\linewidth}%
    \centering
    \includegraphics[width=\linewidth]{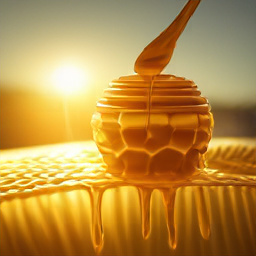}\\[1pt]
    {\tiny\stepcounter{mjhqpanel}(\themjhqpanel)~closeup photo of golden honey being poured, sunny day\par}
  \end{minipage}\hfill%
  \begin{minipage}[t]{0.19\linewidth}%
    \centering
    \includegraphics[width=\linewidth]{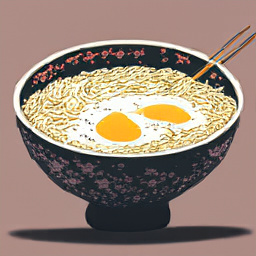}\\[1pt]
    {\tiny\stepcounter{mjhqpanel}(\themjhqpanel)~delicious eggs in ramen bowl, anime style\par}
  \end{minipage}\hfill%
  \begin{minipage}[t]{0.19\linewidth}%
    \centering
    \includegraphics[width=\linewidth]{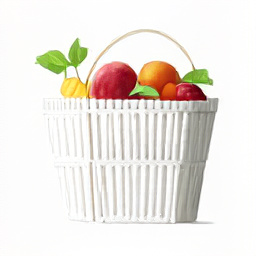}\\[1pt]
    {\tiny\stepcounter{mjhqpanel}(\themjhqpanel)~fruit basket, water colour, white background\par}
  \end{minipage}\hfill%
  \\[4pt]
  \begin{minipage}[t]{0.19\linewidth}%
    \centering
    \includegraphics[width=\linewidth]{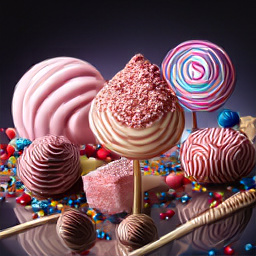}\\[1pt]
    {\tiny\stepcounter{mjhqpanel}(\themjhqpanel)~many beautiful sweets, delicate colors, mouthwatering and beautiful sweets\par}
  \end{minipage}\hfill%
  \begin{minipage}[t]{0.19\linewidth}%
    \centering
    \includegraphics[width=\linewidth]{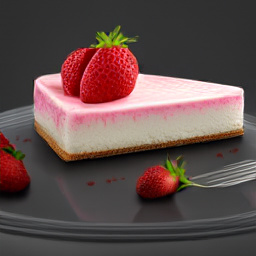}\\[1pt]
    {\tiny\stepcounter{mjhqpanel}(\themjhqpanel)~slice of strawberry cheesecake, hyper realistic,\par}
  \end{minipage}\hfill%
  \begin{minipage}[t]{0.19\linewidth}%
    \centering
    \includegraphics[width=\linewidth]{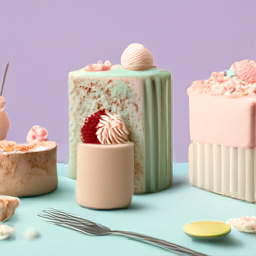}\\[1pt]
    {\tiny\stepcounter{mjhqpanel}(\themjhqpanel)~sweets cakes, commercial photoshot on a solid pastel background, realistic\par}
  \end{minipage}\hfill%
  \begin{minipage}[t]{0.19\linewidth}%
    \centering
    \includegraphics[width=\linewidth]{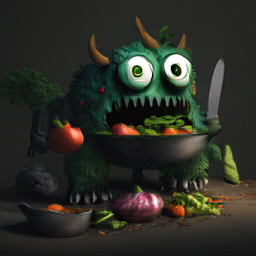}\\[1pt]
    {\tiny\stepcounter{mjhqpanel}(\themjhqpanel)~vegetable monsters being cooked realistic 4k\par}
  \end{minipage}\hfill%
  \begin{minipage}[t]{0.19\linewidth}%
    \centering
    \includegraphics[width=\linewidth]{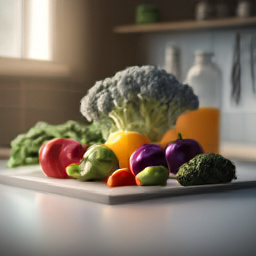}\\[1pt]
    {\tiny\stepcounter{mjhqpanel}(\themjhqpanel)~Vegetables on a countertop in a well lit kitchen, 3D photo realistic\ldots\par}
  \end{minipage}\hfill%
\caption{Text-to-image samples at $256\times256$ from \textsc{MJHQ-30K}
animal (first two rows) and food prompts. Full numbered prompts are in
Tab.~\ref{tab:mjhq-prompts}.}
\label{fig:mjhq-qual}
\end{figure}

Fig.~\ref{fig:mjhq-qual} shows samples for 
\textsc{MJHQ-30K}~\citep{mjhq}
prompts in the
\emph{animals} and \emph{food} categories. 
Prompts beneath the images are abbreviated for space.
The full prompts used for the T2I visuals are
in Tab.~\ref{tab:mjhq-prompts}.\looseness-1

\newcounter{mjhqpanel}

\begin{table}[p]
\centering
\caption{Full \textsc{MJHQ-30K}~\citep{mjhq} prompts for the panels of Fig.~\ref{fig:mjhq-qual}, numbered
in the figure's reading order (left to right, top to bottom); the number appears in front of
each panel's truncated caption. Prompts are reproduced verbatim, including the trailing
keyword lists and the original capitalisation and punctuation, since these are part of what
the model was conditioned on. Panels $1$, $2$, $6$, $15$, $13$ are the five reproduced in
Fig.~\ref{fig:t2i-main-qual}.}
\label{tab:mjhq-prompts}
\scriptsize
\setlength{\tabcolsep}{4pt}
\renewcommand{\arraystretch}{1.15}
\begin{tabular}{@{}r@{\hspace{6pt}}p{\dimexpr\linewidth-2.4em\relax}@{}}
\toprule
\# & prompt \\
\midrule
1 & a butterfly gently touching the vast surface of the lake, peace, serenity, \\
2 & a photo of a cat and a dog on the beach wearing fancy hats \\
3 & abstract photo of a cat jumping through a flower field on a bright sunny day \\
4 & american traditional tiger drinking a beer, sticker, square \\
5 & cute beautiful flamingo, floral garnish, tropical flowers, watercolor vibrant illustration, png \\
6 & grizzly bear wearing medieval armor looking fierce and roaring \\
7 & photorealistic image of a golden retriever happily running through a green field with a lake in the background with cinematic lighting, high definition, depth of field superresolution, insanely detailed \\
8 & Realistic painting of flowers and kittens \\
9 & realistic photo of a pug having drinks at the bar \\
10 & realistic photo, sunset over a lake, bass jumping out of lake, 8k \\
11 & a chocolate bar on a plate, 3d, very clean \\
12 & A colorful still life with a variety of vegetables, fruits and berries arranged on a wooden table, creating an atmosphere of a healthy and balanced diet. \\
13 & a hyperrealistic image of half a lemon dropping into clear water, but on a transparent background \\
14 & a photo of a breakfast sandwich, \\
15 & a realistic photo of a margarita cocktail \\
16 & a sandwich made with cotton candy and chocolate \\
17 & close up photograph of underwater strawberry with water splash, fine art food photography, low angle perspective, dramatic studio lighting, color graded, strong contrast, sharp focus, hyper detailed, realistic, highquality retouching, food, awardwinning photography, editorial photography, \\
18 & closeup photo of golden honey being poured, sunny day \\
19 & delicious eggs in ramen bowl, anime style \\
20 & fruit basket, water colour, white background \\
21 & many beautiful sweets, delicate colors, mouthwatering and beautiful sweets \\
22 & slice of strawberry cheesecake, hyper realistic, \\
23 & sweets cakes, commercial photoshot on a solid pastel background, realistic \\
24 & vegetable monsters being cooked realistic 4k \\
25 & Vegetables on a countertop in a well lit kitchen, 3D photo realistic, cinematic lighting \\
\bottomrule
\end{tabular}
\end{table}


\clearpage

\begin{thebibliography}{71}
\providecommand{\natexlab}[1]{#1}
\providecommand{\url}[1]{\texttt{#1}}
\expandafter\ifx\csname urlstyle\endcsname\relax
  \providecommand{\doi}[1]{doi: #1}\else
  \providecommand{\doi}{doi: \begingroup \urlstyle{rm}\Url}\fi

\bibitem[Assran et~al.(2023)Assran, Duval, Misra, Bojanowski, Vincent, Rabbat, LeCun, and Ballas]{assran2023self}
Mahmoud Assran, Quentin Duval, Ishan Misra, Piotr Bojanowski, Pascal Vincent, Michael Rabbat, Yann LeCun, and Nicolas Ballas.
\newblock Self-supervised learning from images with a joint-embedding predictive architecture.
\newblock In \emph{2023 IEEE/CVF Conference on Computer Vision and Pattern Recognition (CVPR)}, pages 15619--15629. IEEE, 2023.

\bibitem[Ba et~al.(2016)Ba, Kiros, and Hinton]{layernorm}
Jimmy~Lei Ba, Jamie~Ryan Kiros, and Geoffrey~E Hinton.
\newblock Layer normalization.
\newblock \emph{arXiv preprint arXiv:1607.06450}, 2016.

\bibitem[Bachmann et~al.(2025)Bachmann, Allardice, Mizrahi, Fini, Kar, Amirloo, El-Nouby, Zamir, and Dehghan]{flextok}
Roman Bachmann, Jesse Allardice, David Mizrahi, Enrico Fini, O{\u{g}}uzhan~Fatih Kar, Elmira Amirloo, Alaaeldin El-Nouby, Amir Zamir, and Afshin Dehghan.
\newblock Flextok: Resampling images into 1d token sequences of flexible length.
\newblock In \emph{Forty-second International Conference on Machine Learning}, 2025.
\newblock \url{https://openreview.net/forum?id=DgdOkUUBzf}.

\bibitem[Balestriero and Lecun(2024)]{balestriero2024learning}
Randall Balestriero and Yann Lecun.
\newblock How learning by reconstruction produces uninformative features for perception.
\newblock In Ruslan Salakhutdinov, Zico Kolter, Katherine Heller, Adrian Weller, Nuria Oliver, Jonathan Scarlett, and Felix Berkenkamp, editors, \emph{Proceedings of the 41st International Conference on Machine Learning}, volume 235 of \emph{Proceedings of Machine Learning Research}, pages 2566--2585. PMLR, 21--27 Jul 2024.
\newblock \url{https://proceedings.mlr.press/v235/balestriero24b.html}.

\bibitem[{Black Forest Labs}(2024)]{flux2024}
{Black Forest Labs}.
\newblock Flux.
\newblock \url{https://github.com/black-forest-labs/flux}, 2024.

\bibitem[Caron et~al.(2021)Caron, Touvron, Misra, Jegou, Mairal, Bojanowski, and Joulin]{dino2021}
Mathilde Caron, Hugo Touvron, Ishan Misra, Hervé Jegou, Julien Mairal, Piotr Bojanowski, and Armand Joulin.
\newblock Emerging properties in self-supervised vision transformers.
\newblock In \emph{2021 IEEE/CVF International Conference on Computer Vision (ICCV)}, pages 9630--9640, 2021.
\newblock \doi{10.1109/ICCV48922.2021.00951}.

\bibitem[Chen et~al.(2026{\natexlab{a}})Chen, Bi, Tan, Zhang, Zhang, Li, Xiong, Zhang, and Zhang]{aligntok2025}
Bowei Chen, Sai Bi, Hao Tan, He~Zhang, Tianyuan Zhang, Zhengqi Li, Yuanjun Xiong, Jianming Zhang, and Kai Zhang.
\newblock Aligning visual foundation encoders to tokenizers for diffusion models.
\newblock In \emph{The Fourteenth International Conference on Learning Representations}, 2026{\natexlab{a}}.
\newblock \url{https://openreview.net/forum?id=ajnBafpqmE}.

\bibitem[Chen et~al.(2025)Chen, Xu, Pan, Hu, Qin, Goldstein, Huang, Zhou, Xie, Savarese, et~al.]{blip3o}
Jiuhai Chen, Zhiyang Xu, Xichen Pan, Yushi Hu, Can Qin, Tom Goldstein, Lifu Huang, Tianyi Zhou, Saining Xie, Silvio Savarese, et~al.
\newblock Blip3-o: A family of fully open unified multimodal models-architecture, training and dataset.
\newblock \emph{arXiv preprint arXiv:2505.09568}, 2025.

\bibitem[Chen et~al.(2026{\natexlab{b}})Chen, Diao, Wang, Kong, Ren, He, Jiang, and Wu]{chen2026ideal}
Yitong Chen, Zijie Diao, Junke Wang, Lingyu Kong, Yixuan Ren, Bo~He, Yu{-}Gang Jiang, and Zuxuan Wu.
\newblock {IDEAL:} in-depth alignment makes {A} discrete representation autoencoder.
\newblock \emph{CoRR}, abs/2606.11096, 2026{\natexlab{b}}.

\bibitem[Chu et~al.(2026)Chu, Zhang, Han, Li, Yang, Yue, and Guo]{chu2026eostok}
Wenda Chu, Bingliang Zhang, Jiaqi Han, Yizhuo Li, Linjie Yang, Yisong Yue, and Qiushan Guo.
\newblock End-to-end autoregressive image generation with 1d semantic tokenizer.
\newblock In \emph{Forty-third International Conference on Machine Learning}, 2026.
\newblock \url{https://openreview.net/forum?id=fSnxuRhWoC}.

\bibitem[Deng et~al.(2026)Deng, Li, Li, Du, and He]{driftmodel}
Mingyang Deng, He~Li, Tianhong Li, Yilun Du, and Kaiming He.
\newblock Generative modeling via drifting.
\newblock \emph{arXiv preprint arXiv:2602.04770}, 2026.

\bibitem[Du et~al.(2026)Du, Guo, Li, Cui, Xu, Luo, Wei, Gai, Wang, Wu, et~al.]{vqrae2025}
Sinan Du, Jiahao Guo, Bo~Li, Shuhao Cui, Zhengzhuo Xu, Yifu Luo, Yongxian Wei, Kun Gai, Xinggang Wang, Kai Wu, et~al.
\newblock Vqrae: Representation quantization autoencoders for multimodal understanding, generation and reconstruction.
\newblock In \emph{Proceedings of the IEEE/CVF Conference on Computer Vision and Pattern Recognition}, pages 30322--30334, 2026.

\bibitem[Geng et~al.(2026)Geng, Lu, Wu, Shechtman, Kolter, and He]{imfimprovedmeanflow}
Zhengyang Geng, Yiyang Lu, Zongze Wu, Eli Shechtman, J.~Zico Kolter, and Kaiming He.
\newblock Improved mean flows: On the challenges of fastforward generative models.
\newblock In \emph{{IEEE/CVF} Conference on Computer Vision and Pattern Recognition}, pages 30467--30476. Computer Vision Foundation, 2026.

\bibitem[Ghosh et~al.(2023)Ghosh, Hajishirzi, and Schmidt]{geneval}
Dhruba Ghosh, Hannaneh Hajishirzi, and Ludwig Schmidt.
\newblock Geneval: An object-focused framework for evaluating text-to-image alignment.
\newblock In \emph{Advances in Neural Information Processing Systems 36: Annual Conference on Neural Information Processing Systems}, 2023.

\bibitem[Goodfellow et~al.(2014)Goodfellow, Pouget-Abadie, Mirza, Xu, Warde-Farley, Ozair, Courville, and Bengio]{gan2020goodfellow}
Ian~J. Goodfellow, Jean Pouget-Abadie, Mehdi Mirza, Bing Xu, David Warde-Farley, Sherjil Ozair, Aaron Courville, and Yoshua Bengio.
\newblock Generative adversarial nets.
\newblock In \emph{Advances in Neural Information Processing Systems}, volume~27. Curran Associates, Inc., 2014.

\bibitem[Gu et~al.(2024)Gu, Wang, Ge, Shan, and Shou]{gu2024rethinking}
Yuchao Gu, Xintao Wang, Yixiao Ge, Ying Shan, and Mike~Zheng Shou.
\newblock Rethinking the objectives of vector-quantized tokenizers for image synthesis.
\newblock In \emph{{IEEE/CVF} Conference on Computer Vision and Pattern Recognition}, pages 7631--7640. {IEEE}, 2024.

\bibitem[Han et~al.(2025)Han, Chen, Zhao, Wang, Zhao, Yang, He, Yue, and Jiang]{tatok}
Jiaming Han, Hao Chen, Yang Zhao, Hanyu Wang, Qi~Zhao, Ziyan Yang, Hao He, Xiangyu Yue, and Lu~Jiang.
\newblock Vision as a dialect: Unifying visual understanding and generation via text-aligned representations.
\newblock In \emph{The Thirty-ninth Annual Conference on Neural Information Processing Systems}, 2025.
\newblock \url{https://openreview.net/forum?id=ILr4UNiZcQ}.

\bibitem[He et~al.(2022)He, Chen, Xie, Li, Doll{\'a}r, and Girshick]{he2022mae}
Kaiming He, Xinlei Chen, Saining Xie, Yanghao Li, Piotr Doll{\'a}r, and Ross Girshick.
\newblock Masked autoencoders are scalable vision learners.
\newblock In \emph{2022 IEEE/CVF conference on computer vision and pattern recognition (CVPR)}, pages 15979--15988. IEEE, 2022.

\bibitem[Heusel et~al.(2017)Heusel, Ramsauer, Unterthiner, Nessler, and Hochreiter]{FID}
Martin Heusel, Hubert Ramsauer, Thomas Unterthiner, Bernhard Nessler, and Sepp Hochreiter.
\newblock Gans trained by a two time-scale update rule converge to a local nash equilibrium.
\newblock In \emph{Advances in Neural Information Processing Systems 30: Annual Conference on Neural Information Processing Systems}, pages 6626--6637, 2017.

\bibitem[Hu et~al.(2024)Hu, Wang, Fang, Fu, Cheng, and Yu]{ella}
Xiwei Hu, Rui Wang, Yixiao Fang, Bin Fu, Pei Cheng, and Gang Yu.
\newblock Ella: Equip diffusion models with llm for enhanced semantic alignment.
\newblock \emph{arXiv preprint arXiv:2403.05135}, 2024.

\bibitem[Jayasumana et~al.(2024)Jayasumana, Ramalingam, Veit, Glasner, Chakrabarti, and Kumar]{jayasumana2024cmmd}
Sadeep Jayasumana, Srikumar Ramalingam, Andreas Veit, Daniel Glasner, Ayan Chakrabarti, and Sanjiv Kumar.
\newblock Rethinking {FID:} towards a better evaluation metric for image generation.
\newblock In \emph{{IEEE/CVF} Conference on Computer Vision and Pattern Recognition}, pages 9307--9315. {IEEE}, 2024.

\bibitem[Jia et~al.(2025)Jia, Li, Shu, Zheng, Fan, Guo, Shi, Lu, Li, Guo, et~al.]{dinotok2025}
Mingkai Jia, Mingxiao Li, Zhijian Shu, Anlin Zheng, Liaoyuan Fan, Jiaxin Guo, Tianxing Shi, Dongyue Lu, Zeming Li, Xiaoyang Guo, et~al.
\newblock Dino-tok: Adapting dino for visual tokenizers.
\newblock \emph{arXiv preprint arXiv:2511.20565}, 2025.

\bibitem[Kirstain et~al.(2023)Kirstain, Polyak, Singer, Matiana, Penna, and Levy]{Kirstain2023Pick}
Yuval Kirstain, Adam Polyak, Uriel Singer, Shahbuland Matiana, Joe Penna, and Omer Levy.
\newblock Pick-a-pic: An open dataset of user preferences for text-to-image generation.
\newblock \emph{Advances in neural information processing systems}, 36:\penalty0 36652--36663, 2023.

\bibitem[Kynk{\"{a}}{\"{a}}nniemi et~al.(2019)Kynk{\"{a}}{\"{a}}nniemi, Karras, Laine, Lehtinen, and Aila]{improvedprecisionrecall}
Tuomas Kynk{\"{a}}{\"{a}}nniemi, Tero Karras, Samuli Laine, Jaakko Lehtinen, and Timo Aila.
\newblock Improved precision and recall metric for assessing generative models.
\newblock In \emph{Advances in Neural Information Processing Systems 32: Annual Conference on Neural Information Processing Systems}, pages 3929--3938, 2019.

\bibitem[Kynk{\"a}{\"a}nniemi et~al.(2023)Kynk{\"a}{\"a}nniemi, Karras, Aittala, Aila, and Lehtinen]{kynkaanniemi2022role}
Tuomas Kynk{\"a}{\"a}nniemi, Tero Karras, Miika Aittala, Timo Aila, and Jaakko Lehtinen.
\newblock The role of imagenet classes in fr\'echet inception distance.
\newblock In \emph{The Eleventh International Conference on Learning Representations}, 2023.
\newblock \url{https://openreview.net/forum?id=4oXTQ6m_ws8}.

\bibitem[Lavoie et~al.(2025)Lavoie, Noukhovitch, and Courville]{DLClavoie}
Samuel Lavoie, Michael Noukhovitch, and Aaron Courville.
\newblock Compositional discrete latent code for high fidelity, productive diffusion models.
\newblock In \emph{The Thirty-ninth Annual Conference on Neural Information Processing Systems}, 2025.
\newblock \url{https://openreview.net/forum?id=1iSnpztjbD}.

\bibitem[Leng et~al.(2025)Leng, Singh, Hou, Xing, Xie, and Zheng]{repae2025}
Xingjian Leng, Jaskirat Singh, Yunzhong Hou, Zhenchang Xing, Saining Xie, and Liang Zheng.
\newblock Repa-e: Unlocking vae for end-to-end tuning with latent diffusion transformers.
\newblock In \emph{2025 IEEE/CVF International Conference on Computer Vision (ICCV)}, pages 18262--18272. IEEE, 2025.

\bibitem[Li et~al.(2024{\natexlab{a}})Li, Lin, Pathak, Li, Fei, Wu, Ling, Xia, Zhang, Neubig, et~al.]{genaibench}
Baiqi Li, Zhiqiu Lin, Deepak Pathak, Jiayao Li, Yixin Fei, Kewen Wu, Tiffany Ling, Xide Xia, Pengchuan Zhang, Graham Neubig, et~al.
\newblock Genai-bench: Evaluating and improving compositional text-to-visual generation.
\newblock \emph{arXiv preprint arXiv:2406.13743}, 2024{\natexlab{a}}.

\bibitem[Li et~al.(2024{\natexlab{b}})Li, Kamko, Akhgari, Sabet, Xu, and Doshi]{mjhq}
Daiqing Li, Aleks Kamko, Ehsan Akhgari, Ali Sabet, Linmiao Xu, and Suhail Doshi.
\newblock Playground v2. 5: Three insights towards enhancing aesthetic quality in text-to-image generation.
\newblock \emph{arXiv preprint arXiv:2402.17245}, 2024{\natexlab{b}}.

\bibitem[Li and He(2026)]{jit}
Tianhong Li and Kaiming He.
\newblock Back to basics: Let denoising generative models denoise.
\newblock In \emph{{IEEE/CVF} Conference on Computer Vision and Pattern Recognition}, pages 36115--36125. Computer Vision Foundation, 2026.

\bibitem[Lu et~al.(2026{\natexlab{a}})Lu, Song, Xu, Ahn, Wang, Chen, Dehghan, and Yang]{gram}
Jiasen Lu, Liangchen Song, Mingze Xu, Byeongjoo Ahn, Yanjun Wang, Chen Chen, Afshin Dehghan, and Yinfei Yang.
\newblock Atoken: A unified tokenizer for vision.
\newblock In \emph{Proceedings of the IEEE/CVF Conference on Computer Vision and Pattern Recognition}, pages 28701--28711, 2026{\natexlab{a}}.

\bibitem[Lu et~al.(2026{\natexlab{b}})Lu, Lu, Sun, Zhao, Jiang, Wang, Li, Geng, and He]{pmfpixelmeanflow}
Yiyang Lu, Susie Lu, Qiao Sun, Hanhong Zhao, Zhicheng Jiang, Xianbang Wang, Tianhong Li, Zhengyang Geng, and Kaiming He.
\newblock One-step latent-free image generation with pixel mean flows.
\newblock \emph{CoRR}, abs/2601.22158, 2026{\natexlab{b}}.

\bibitem[Luo et~al.(2024)Luo, Shi, Ge, Yang, Wang, and Shan]{luo2024open}
Zhuoyan Luo, Fengyuan Shi, Yixiao Ge, Yujiu Yang, Limin Wang, and Ying Shan.
\newblock Open-magvit2: An open-source project toward democratizing auto-regressive visual generation.
\newblock \emph{arXiv preprint arXiv:2409.04410}, 2024.

\bibitem[Ma et~al.(2024)Ma, Goldstein, Albergo, Boffi, Vanden-Eijnden, and Xie]{Ma2024}
Nanye Ma, Mark Goldstein, Michael~S Albergo, Nicholas~M Boffi, Eric Vanden-Eijnden, and Saining Xie.
\newblock Sit: Exploring flow and diffusion-based generative models with scalable interpolant transformers.
\newblock In \emph{European Conference on Computer Vision}, pages 23--40. Springer, 2024.

\bibitem[Ma et~al.(2025)Ma, Wu, Sun, and Li]{hpsv3}
Yuhang Ma, Xiaoshi Wu, Keqiang Sun, and Hongsheng Li.
\newblock {HPSv3}: Towards wide-spectrum human preference score.
\newblock In \emph{Proceedings of the IEEE/CVF International Conference on Computer Vision}, pages 15086--15095, 2025.

\bibitem[Mentzer et~al.(2024)Mentzer, Minnen, Agustsson, and Tschannen]{fsq2024}
Fabian Mentzer, David Minnen, Eirikur Agustsson, and Michael Tschannen.
\newblock Finite scalar quantization: Vq-vae made simple.
\newblock In \emph{International Conference on Learning Representations}, volume 2024, pages 51772--51783, 2024.

\bibitem[Naeem et~al.(2020)Naeem, Oh, Uh, Choi, and Yoo]{densitycoverage}
Muhammad~Ferjad Naeem, Seong~Joon Oh, Youngjung Uh, Yunjey Choi, and Jaejun Yoo.
\newblock Reliable fidelity and diversity metrics for generative models.
\newblock In \emph{Proceedings of the 37th International Conference on Machine Learning}, volume 119 of \emph{Proceedings of Machine Learning Research}, pages 7176--7185. {PMLR}, 2020.

\bibitem[Oquab et~al.(2024)Oquab, Darcet, Moutakanni, Vo, Szafraniec, Khalidov, Fernandez, HAZIZA, Massa, El-Nouby, Assran, Ballas, Galuba, Howes, Huang, Li, Misra, Rabbat, Sharma, Synnaeve, Xu, Jegou, Mairal, Labatut, Joulin, and Bojanowski]{dinov2}
Maxime Oquab, Timoth{\'e}e Darcet, Th{\'e}o Moutakanni, Huy~V. Vo, Marc Szafraniec, Vasil Khalidov, Pierre Fernandez, Daniel HAZIZA, Francisco Massa, Alaaeldin El-Nouby, Mido Assran, Nicolas Ballas, Wojciech Galuba, Russell Howes, Po-Yao Huang, Shang-Wen Li, Ishan Misra, Michael Rabbat, Vasu Sharma, Gabriel Synnaeve, Hu~Xu, Herve Jegou, Julien Mairal, Patrick Labatut, Armand Joulin, and Piotr Bojanowski.
\newblock {DINO}v2: Learning robust visual features without supervision.
\newblock \emph{Transactions on Machine Learning Research}, 2024.
\newblock ISSN 2835-8856.
\newblock \url{https://openreview.net/forum?id=a68SUt6zFt}.
\newblock Featured Certification.

\bibitem[{Qwen Team}(2025)]{qwen3}
{Qwen Team}.
\newblock Qwen3 technical report, 2025.
\newblock \url{https://arxiv.org/abs/2505.09388}.

\bibitem[Radford et~al.(2021)Radford, Kim, Hallacy, Ramesh, Goh, Agarwal, Sastry, Askell, Mishkin, Clark, et~al.]{clip}
Alec Radford, Jong~Wook Kim, Chris Hallacy, Aditya Ramesh, Gabriel Goh, Sandhini Agarwal, Girish Sastry, Amanda Askell, Pamela Mishkin, Jack Clark, et~al.
\newblock Learning transferable visual models from natural language supervision.
\newblock In \emph{International conference on machine learning}, pages 8748--8763. PmLR, 2021.

\bibitem[Ramanujan et~al.(2025)Ramanujan, Tirumala, Aghajanyan, Zettlemoyer, and Farhadi]{ramanujan2026worse}
Vivek Ramanujan, Kushal Tirumala, Armen Aghajanyan, Luke Zettlemoyer, and Ali Farhadi.
\newblock When worse is better: Navigating the compression generation trade-off in visual tokenization.
\newblock In \emph{The Thirty-ninth Annual Conference on Neural Information Processing Systems}, 2025.
\newblock \url{https://openreview.net/forum?id=o8hWyJIgAV}.

\bibitem[Russakovsky et~al.(2015)Russakovsky, Deng, Su, Krause, Satheesh, Ma, Huang, Karpathy, Khosla, Bernstein, et~al.]{imagenet}
Olga Russakovsky, Jia Deng, Hao Su, Jonathan Krause, Sanjeev Satheesh, Sean Ma, Zhiheng Huang, Andrej Karpathy, Aditya Khosla, Michael Bernstein, et~al.
\newblock Imagenet large scale visual recognition challenge.
\newblock \emph{International journal of computer vision}, 115\penalty0 (3):\penalty0 211--252, 2015.

\bibitem[Sim{\'e}oni et~al.(2026)Sim{\'e}oni, Vo, Seitzer, Baldassarre, Oquab, Jose, Khalidov, Szafraniec, Yi, Ramamonjisoa, Massa, HAZIZA, Wehrstedt, Wang, Darcet, Moutakanni, Sentana, Roberts, Vedaldi, Tolan, Brandt, Couprie, Mairal, Jegou, Labatut, and Bojanowski]{dinov3}
Oriane Sim{\'e}oni, Huy~V. Vo, Maximilian Seitzer, Federico Baldassarre, Maxime Oquab, Cijo Jose, Vasil Khalidov, Marc Szafraniec, Seung~Eun Yi, Michael Ramamonjisoa, Francisco Massa, Daniel HAZIZA, Luca Wehrstedt, Jianyuan Wang, Timoth{\'e}e Darcet, Th{\'e}o Moutakanni, Leonel Sentana, Claire Roberts, Andrea Vedaldi, Jamie Tolan, John Brandt, Camille Couprie, Julien Mairal, Herve Jegou, Patrick Labatut, and Piotr Bojanowski.
\newblock {DINO}v3.
\newblock \emph{Transactions on Machine Learning Research}, 2026.
\newblock ISSN 2835-8856.
\newblock \url{https://openreview.net/forum?id=2NlGyqNjns}.
\newblock Featured Certification.

\bibitem[Singh et~al.(2026)Singh, Zheng, Wu, Zhang, Shechtman, and Xie]{Singh2026raev2}
Jaskirat Singh, Boyang Zheng, Zongze Wu, Richard Zhang, Eli Shechtman, and Saining Xie.
\newblock Improved baselines with representation autoencoders.
\newblock \emph{arXiv preprint arXiv:2605.18324}, 2026.

\bibitem[Skorokhodov et~al.(2025)Skorokhodov, Girish, Hu, Menapace, Li, Abdal, Tulyakov, and Siarohin]{Skorokhodov2025diffusability}
Ivan Skorokhodov, Sharath Girish, Benran Hu, Willi Menapace, Yanyu Li, Rameen Abdal, Sergey Tulyakov, and Aliaksandr Siarohin.
\newblock Improving the diffusability of autoencoders.
\newblock In \emph{Forty-second International Conference on Machine Learning}, 2025.
\newblock \url{https://openreview.net/forum?id=2hEDcA7xy4}.

\bibitem[Stein et~al.(2023)Stein, Cresswell, Hosseinzadeh, Sui, Ross, Villecroze, Liu, Caterini, Taylor, and Loaiza-Ganem]{dinov2FD}
George Stein, Jesse~C. Cresswell, Rasa Hosseinzadeh, Yi~Sui, Brendan~Leigh Ross, Valentin Villecroze, Zhaoyan Liu, Anthony~L. Caterini, Eric Taylor, and Gabriel Loaiza-Ganem.
\newblock Exposing flaws of generative model evaluation metrics and their unfair treatment of diffusion models.
\newblock In \emph{Thirty-seventh Conference on Neural Information Processing Systems}, 2023.
\newblock \url{https://openreview.net/forum?id=08zf7kTOoh}.

\bibitem[Sun et~al.(2024)Sun, Jiang, Chen, Zhang, Peng, Luo, and Yuan]{llamagen2024}
Peize Sun, Yi~Jiang, Shoufa Chen, Shilong Zhang, Bingyue Peng, Ping Luo, and Zehuan Yuan.
\newblock Autoregressive model beats diffusion: Llama for scalable image generation.
\newblock \emph{arXiv preprint arXiv:2406.06525}, 2024.

\bibitem[Szegedy et~al.(2016)Szegedy, Vanhoucke, Ioffe, Shlens, and Wojna]{inceptionv3}
Christian Szegedy, Vincent Vanhoucke, Sergey Ioffe, Jonathon Shlens, and Zbigniew Wojna.
\newblock Rethinking the inception architecture for computer vision.
\newblock In \emph{2016 {IEEE} Conference on Computer Vision and Pattern Recognition}, pages 2818--2826. {IEEE} Computer Society, 2016.

\bibitem[Tian et~al.(2024)Tian, Jiang, Yuan, Peng, and Wang]{var2024}
Keyu Tian, Yi~Jiang, Zehuan Yuan, Bingyue Peng, and Liwei Wang.
\newblock Visual autoregressive modeling: Scalable image generation via next-scale prediction.
\newblock \emph{Advances in neural information processing systems}, 37:\penalty0 84839--84865, 2024.

\bibitem[Tschannen et~al.(2025)Tschannen, Gritsenko, Wang, Naeem, Alabdulmohsin, Parthasarathy, Evans, Beyer, Xia, Mustafa, et~al.]{siglip2}
Michael Tschannen, Alexey Gritsenko, Xiao Wang, Muhammad~Ferjad Naeem, Ibrahim Alabdulmohsin, Nikhil Parthasarathy, Talfan Evans, Lucas Beyer, Ye~Xia, Basil Mustafa, et~al.
\newblock Siglip 2: Multilingual vision-language encoders with improved semantic understanding, localization, and dense features.
\newblock \emph{arXiv preprint arXiv:2502.14786}, 2025.

\bibitem[Van Den~Oord et~al.(2017)Van Den~Oord, Vinyals, et~al.]{vqvae}
Aaron Van Den~Oord, Oriol Vinyals, et~al.
\newblock Neural discrete representation learning.
\newblock \emph{Advances in neural information processing systems}, 30, 2017.

\bibitem[Wang et~al.(2024)Wang, Zhao, Zhang, Feng, Liu, and Kang]{vqkdwang2024vqunderstanding}
Luting Wang, Yang Zhao, Zijian Zhang, Jiashi Feng, Si~Liu, and Bingyi Kang.
\newblock Image understanding makes for a good tokenizer for image generation.
\newblock In \emph{The Thirty-eighth Annual Conference on Neural Information Processing Systems}, 2024.
\newblock \url{https://openreview.net/forum?id=RMmgu49lwn}.

\bibitem[Wang et~al.(2026)Wang, Tian, Huang, and Wang]{Wang2025ddt}
Shuai Wang, Zhi Tian, Weilin Huang, and Limin Wang.
\newblock Ddt: Decoupled diffusion transformer.
\newblock In \emph{Proceedings of the IEEE/CVF conference on computer vision and pattern recognition}, pages 40633--40642, 2026.

\bibitem[Weber et~al.(2024)Weber, Yu, Yu, Deng, Shen, Cremers, and Chen]{weber2024maskbit}
Mark Weber, Lijun Yu, Qihang Yu, Xueqing Deng, Xiaohui Shen, Daniel Cremers, and Liang-Chieh Chen.
\newblock Maskbit: Embedding-free image generation via bit tokens.
\newblock \emph{Transactions on Machine Learning Research}, 2024.
\newblock ISSN 2835-8856.
\newblock \url{https://openreview.net/forum?id=NYe2JuN3v3}.
\newblock Featured Certification, Reproducibility Certification.

\bibitem[Woo et~al.(2023)Woo, Debnath, Hu, Chen, Liu, Kweon, and Xie]{convnextv2}
Sanghyun Woo, Shoubhik Debnath, Ronghang Hu, Xinlei Chen, Zhuang Liu, In~So Kweon, and Saining Xie.
\newblock Convnext {V2:} co-designing and scaling convnets with masked autoencoders.
\newblock In \emph{{IEEE/CVF} Conference on Computer Vision and Pattern Recognition}, pages 16133--16142. {IEEE}, 2023.

\bibitem[Wu et~al.(2025)Wu, Zhang, Shi, Gao, Chen, Wang, Chen, Gao, Tang, jian Yang, Cheng, and Li]{reg2025}
Ge~Wu, Shen Zhang, Ruijing Shi, Shanghua Gao, Zhenyuan Chen, Lei Wang, Zhaowei Chen, Hongcheng Gao, Yao Tang, jian Yang, Ming-Ming Cheng, and Xiang Li.
\newblock Representation entanglement for generation: Training diffusion transformers is much easier than you think.
\newblock In \emph{The Thirty-ninth Annual Conference on Neural Information Processing Systems}, 2025.
\newblock \url{https://openreview.net/forum?id=koEALFNBj1}.

\bibitem[Wu et~al.(2023)Wu, Hao, Sun, Chen, Zhu, Zhao, and Li]{Wu2023HPSv2}
Xiaoshi Wu, Yiming Hao, Keqiang Sun, Yixiong Chen, Feng Zhu, Rui Zhao, and Hongsheng Li.
\newblock Human preference score v2: A solid benchmark for evaluating human preferences of text-to-image synthesis.
\newblock \emph{arXiv preprint arXiv:2306.09341}, 2023.

\bibitem[Xiong et~al.(2025)Xiong, Liew, Huang, Feng, and Liu]{xiong2025gigatok}
Tianwei Xiong, Jun~Hao Liew, Zilong Huang, Jiashi Feng, and Xihui Liu.
\newblock Gigatok: Scaling visual tokenizers to 3 billion parameters for autoregressive image generation.
\newblock In \emph{{IEEE/CVF} International Conference on Computer Vision}, pages 18770--18780. {IEEE}, 2025.

\bibitem[Yang et~al.(2026)Yang, Geng, Ju, Tian, and Wang]{fdloss2026}
Jiawei Yang, Zhengyang Geng, Xuan Ju, Yonglong Tian, and Yue Wang.
\newblock Representation fr{\'e}chet loss for visual generation.
\newblock \emph{arXiv preprint arXiv:2604.28190}, 2026.
\newblock \url{https://arxiv.org/abs/2604.28190}.

\bibitem[Yao et~al.(2025)Yao, Yang, and Wang]{Yao2025}
Jingfeng Yao, Bin Yang, and Xinggang Wang.
\newblock Reconstruction vs. generation: Taming optimization dilemma in latent diffusion models.
\newblock In \emph{2025 IEEE/CVF Conference on Computer Vision and Pattern Recognition (CVPR)}, pages 15703--15712. IEEE, 2025.

\bibitem[Yu et~al.(2024{\natexlab{a}})Yu, Lezama, Gundavarapu, Versari, Sohn, Minnen, Cheng, Gupta, Gu, Hauptmann, et~al.]{magvit2}
Lijun Yu, Jos{\'e} Lezama, Nitesh~Bharadwaj Gundavarapu, Luca Versari, Kihyuk Sohn, David Minnen, Yong Cheng, Agrim Gupta, Xiuye Gu, Alexander~G Hauptmann, et~al.
\newblock Language model beats diffusion-tokenizer is key to visual generation.
\newblock In \emph{International Conference on Learning Representations}, volume 2024, pages 765--783, 2024{\natexlab{a}}.

\bibitem[Yu et~al.(2024{\natexlab{b}})Yu, Weber, Deng, Shen, Cremers, and Chen]{yu2024image}
Qihang Yu, Mark Weber, Xueqing Deng, Xiaohui Shen, Daniel Cremers, and Liang-Chieh Chen.
\newblock An image is worth 32 tokens for reconstruction and generation.
\newblock \emph{Advances in Neural Information Processing Systems}, 37:\penalty0 128940--128966, 2024{\natexlab{b}}.

\bibitem[Yu et~al.(2025{\natexlab{a}})Yu, He, Deng, Shen, and Chen]{yu2025randomized}
Qihang Yu, Ju~He, Xueqing Deng, Xiaohui Shen, and Liang-Chieh Chen.
\newblock Randomized autoregressive visual generation.
\newblock In \emph{2025 IEEE/CVF International Conference on Computer Vision (ICCV)}, pages 18431--18441. IEEE, 2025{\natexlab{a}}.

\bibitem[Yu et~al.(2026)Yu, Liu, He, Zhang, Liu, Chen, and Chen]{yu2026bar}
Qihang Yu, Qihao Liu, Ju~He, Xinyang Zhang, Yang Liu, Liang-Chieh Chen, and Xi~Chen.
\newblock Autoregressive image generation with masked bit modeling.
\newblock In \emph{Forty-third International Conference on Machine Learning}, 2026.
\newblock \url{https://openreview.net/forum?id=jALSox02QD}.

\bibitem[Yu et~al.(2025{\natexlab{b}})Yu, Kwak, Jang, Jeong, Huang, Shin, and Xie]{Yu2025repa}
Sihyun Yu, Sangkyung Kwak, Huiwon Jang, Jongheon Jeong, Jonathan Huang, Jinwoo Shin, and Saining Xie.
\newblock Representation alignment for generation: Training diffusion transformers is easier than you think.
\newblock In \emph{The Thirteenth International Conference on Learning Representations}, 2025{\natexlab{b}}.
\newblock \url{https://openreview.net/forum?id=DJSZGGZYVi}.

\bibitem[Zhai et~al.(2023)Zhai, Mustafa, Kolesnikov, and Beyer]{siglip}
Xiaohua Zhai, Basil Mustafa, Alexander Kolesnikov, and Lucas Beyer.
\newblock Sigmoid loss for language image pre-training.
\newblock In \emph{2023 IEEE/CVF International Conference on Computer Vision (ICCV)}, pages 11941--11952. IEEE, 2023.

\bibitem[Zhang et~al.(2018)Zhang, Isola, Efros, Shechtman, and Wang]{lpips}
Richard Zhang, Phillip Isola, Alexei~A. Efros, Eli Shechtman, and Oliver Wang.
\newblock The unreasonable effectiveness of deep features as a perceptual metric.
\newblock In \emph{2018 {IEEE} Conference on Computer Vision and Pattern Recognition}, pages 586--595. Computer Vision Foundation / {IEEE} Computer Society, 2018.

\bibitem[Zheng et~al.(2025{\natexlab{a}})Zheng, Wang, Zhao, Deng, Wang, Zhang, and Qi]{zheng2025hita}
Anlin Zheng, Haochen Wang, Yucheng Zhao, Weipeng Deng, Tiancai Wang, Xiangyu Zhang, and Xiaojuan Qi.
\newblock Holistic tokenizer for autoregressive image generation.
\newblock In \emph{Proceedings of the IEEE/CVF International Conference on Computer Vision}, pages 16916--16926, 2025{\natexlab{a}}.

\bibitem[Zheng et~al.(2025{\natexlab{b}})Zheng, Wen, Zhang, Ma, Wang, YU, Zhang, and QI]{vfmtok}
Anlin Zheng, Xin Wen, Xuanyang Zhang, Chuofan Ma, Tiancai Wang, Gang YU, Xiangyu Zhang, and XIAOJUAN QI.
\newblock Vision foundation models as effective visual tokenizers for autoregressive generation.
\newblock In \emph{The Thirty-ninth Annual Conference on Neural Information Processing Systems}, 2025{\natexlab{b}}.
\newblock \url{https://openreview.net/forum?id=PESrAH82Zh}.

\bibitem[Zheng et~al.(2026{\natexlab{a}})Zheng, Han, Wen, Ma, Gong, Yu, Zhang, and Qi]{vfmtok2026}
Anlin Zheng, Qi~Han, Xin Wen, Chuofan Ma, Lanxi Gong, Gang Yu, Xiangyu Zhang, and Xiaojuan Qi.
\newblock Vision foundation models as generalist tokenizers for image generation.
\newblock \emph{arXiv preprint arXiv:2605.18390}, 2026{\natexlab{a}}.

\bibitem[Zheng et~al.(2026{\natexlab{b}})Zheng, Ma, Tong, and Xie]{Zheng2025rae}
Boyang Zheng, Nanye Ma, Shengbang Tong, and Saining Xie.
\newblock Diffusion transformers with representation autoencoders.
\newblock In \emph{International Conference on Learning Representations}, volume 2026, pages 35791--35820, 2026{\natexlab{b}}.

\end{thebibliography}
\end{document}